\documentclass{article}

\usepackage[square,numbers]{natbib}
\usepackage[preprint]{neurips_2025}

\usepackage[utf8]{inputenc} 
\usepackage[T1]{fontenc}    
\usepackage{url}            
\usepackage{booktabs}       
\usepackage{amsfonts}       
\usepackage{nicefrac}       
\usepackage{microtype}      
\usepackage{xcolor}         

\usepackage[colorlinks=true, colorlinks, linkcolor=red, urlcolor=magenta, anchorcolor=blue, citecolor=jade]{hyperref}

\usepackage{graphicx}
\usepackage{float}
\usepackage{subfigure} 
\usepackage[flushleft]{threeparttable}
\usepackage{amssymb}       

\usepackage{comment}       
\usepackage{mathtools}     

\usepackage{algorithm}

\usepackage{tikz}
\usetikzlibrary{shapes, arrows.meta, positioning, fit}
\usepackage{geometry} 
\usepackage{pifont}

\usepackage{algpseudocode}  

\newcommand{\mysize}{4cm}

\usepackage{amsmath}
\usepackage{bbm}
\usepackage[notransparent,inkscapeversion=1]{svg}
\usepackage{enumerate}
\usepackage{enumitem}

\usepackage{multirow}
\usepackage{multicol}
\usepackage{tabularx}
\usepackage{bm}
\usepackage{pifont}
\usepackage{makecell}
\usepackage{siunitx}
\usepackage{array}
\usepackage[export]{adjustbox}
\usepackage{supertabular}

\usepackage{utfsym}

\usepackage{longtable}
\usepackage{xltabular}
\usepackage{booktabs}
\usepackage{tabularray}
\usepackage[font=small]{caption} 

\usepackage[most]{tcolorbox}
\usepackage{float}

\usepackage{CJKutf8}
\usepackage{setspace}  

\usepackage{enumitem} 
\usepackage{listings}

\lstdefinestyle{mypython}{
    language=Python,                
    backgroundcolor=\color{white},  
    basicstyle=\ttfamily\footnotesize, 
    numbers=none,                   
    numberstyle=\tiny\color{gray},   
    keywordstyle=\color{magenta},       
    commentstyle=\color{green},     
    stringstyle=\color{stringgreen},        
    rulecolor=\color{black},        
    breaklines=true,                
    showstringspaces=false,          
}

\usepackage{color}
\usepackage{cleveref}

\usepackage{colortbl} 
\definecolor{junglegreen}{rgb}{0.16, 0.67, 0.53}
\definecolor{mint}{rgb}{0.24, 0.71, 0.54}
\definecolor{teal}{rgb}{0.0, 0.5, 0.5}
\definecolor{persiangreen}{rgb}{0.0, 0.65, 0.58}
\definecolor{cadmiumgreen}{rgb}{0.0, 0.42, 0.24}
\definecolor{stringgreen}{rgb}{0.24, 0.56, 0.48}
\definecolor{firebrick}{rgb}{0.7, 0.13, 0.13}
\definecolor{alizarin}{rgb}{0.82, 0.1, 0.26}
\definecolor{ferrarired}{rgb}{1.0, 0.11, 0.0}
\definecolor{ruddy}{rgb}{1.0, 0.0, 0.16}

\definecolor{ceruleanblue}{rgb}{0.16, 0.32, 0.75}
\definecolor{ao}{rgb}{0.0, 0.0, 1.0}
\definecolor{blue(pigment)}{rgb}{0.2, 0.2, 0.6}
\definecolor{persianblue}{rgb}{0.11, 0.22, 0.73}
\definecolor{deepskyblue}{rgb}{0.0, 0.75, 1}
\definecolor{dodgerblue}{rgb}{0.12, 0.56, 1}
\definecolor{lightcyan}{rgb}{0.88, 1.0, 1.0}
\definecolor{paleturquoise}{rgb}{0.69, 0.93, 0.93}

\definecolor{aureolin}{rgb}{0.99, 0.93, 0.0}
\definecolor{cadmiumyellow}{rgb}{1.0, 0.96, 0.0}
\definecolor{daffodil}{rgb}{1.0, 1.0, 0.19}
\definecolor{alizarin}{rgb}{0.82, 0.1, 0.26}
\definecolor{yellow(ryb)}{rgb}{1.0, 1.0, 0.2}
\definecolor{canaryyellow}{rgb}{1.0, 0.94, 0.0}

\definecolor{jade}{rgb}{0.0, 0.66, 0.42}
\definecolor{fuchsia}{rgb}{1.0, 0.0, 1.0}
\definecolor{lightgray}{rgb}{0.83, 0.83, 0.83}

\definecolor{lightyellow}{rgb}{0.99, 0.95, 0.89}
\definecolor{lightyellow2}{rgb}{0.99, 0.92, 0.75}
\definecolor{amber(sae/ece)}{rgb}{1.0, 0.49, 0.0}

\definecolor{lightpurple}{rgb}{0.95, 0.89, 0.99}
\definecolor{guppiegreen}{rgb}{0.0, 1.0, 0.5}
\definecolor{icterine}{rgb}{0.99, 0.97, 0.37}
\definecolor{lightgreen}{rgb}{0.89, 0.99, 0.95}
\definecolor{lightgreen2}{rgb}{0.56, 0.93, 0.56}
\definecolor{green(html/cssgreen)}{rgb}{0.0, 0.5, 0.0}
\definecolor{green(pigment)}{rgb}{0.0, 0.65, 0.31}
\definecolor{officegreen}{rgb}{0.0, 0.5, 0.0}
\definecolor{green(pigment)}{rgb}{0.0, 0.65, 0.31}
\definecolor{green(ncs)}{rgb}{0.0, 0.62, 0.42}
\definecolor{green(munsell)}{rgb}{0.0, 0.66, 0.47}
\definecolor{ao(english)}{rgb}{0.0, 0.5, 0.0}
\definecolor{darkspringgreen}{rgb}{0.09, 0.45, 0.27}
\definecolor{forestgreen(web)}{rgb}{0.13, 0.55, 0.13}

\definecolor{sunset}{rgb}{0.98, 0.84, 0.65}
\definecolor{peachpuff}{rgb}{1.0, 0.85, 0.73}
\definecolor{lightgreen3}{rgb}{0.69, 0.96, 0.80}
\definecolor{lightkhaki}{rgb}{0.94, 0.9, 0.55}
\definecolor{mintgreen}{rgb}{0.6, 1.0, 0.6}
\definecolor{lemon}{rgb}{1.0, 0.97, 0.0}
\definecolor{laserlemon}{rgb}{1.0, 1.0, 0.13}
\definecolor{flavescent}{rgb}{0.97, 0.91, 0.56}
\definecolor{maize}{rgb}{0.98, 0.93, 0.37}
\definecolor{laserlemon}{rgb}{1.0, 1.0, 0.13}
\definecolor{corn}{rgb}{0.98, 0.93, 0.36}
\definecolor{stildegrainyellow}{rgb}{0.98, 0.85, 0.37}
\definecolor{schoolbusyellow}{rgb}{1.0, 0.85, 0.0}
\definecolor{pastelgray}{rgb}{0.81, 0.81, 0.77}
\definecolor{gainsboro}{rgb}{0.86, 0.86, 0.86}
\definecolor{isabelline}{rgb}{0.96, 0.94, 0.93}
\definecolor{splashedwhite}{rgb}{1.0, 0.99, 1.0}

\definecolor{anti-flashwhite}{rgb}{0.95, 0.95, 0.96}
\definecolor{aliceblue}{rgb}{0.94, 0.97, 1.0}
\definecolor{platinum}{rgb}{0.9, 0.89, 0.89}
\definecolor{snow}{rgb}{1.0, 0.98, 0.98}
\definecolor{whitesmoke}{rgb}{0.96, 0.96, 0.96}
\definecolor{gainsboro}{rgb}{0.86, 0.86, 0.86}

\title{\textcolor{ruddy}{N}eXTab-\textcolor{persianblue}{V}L-\textcolor{yellow}{Chat} 
\centering
$\vcenter{\hbox{\includegraphics[scale=0.022]{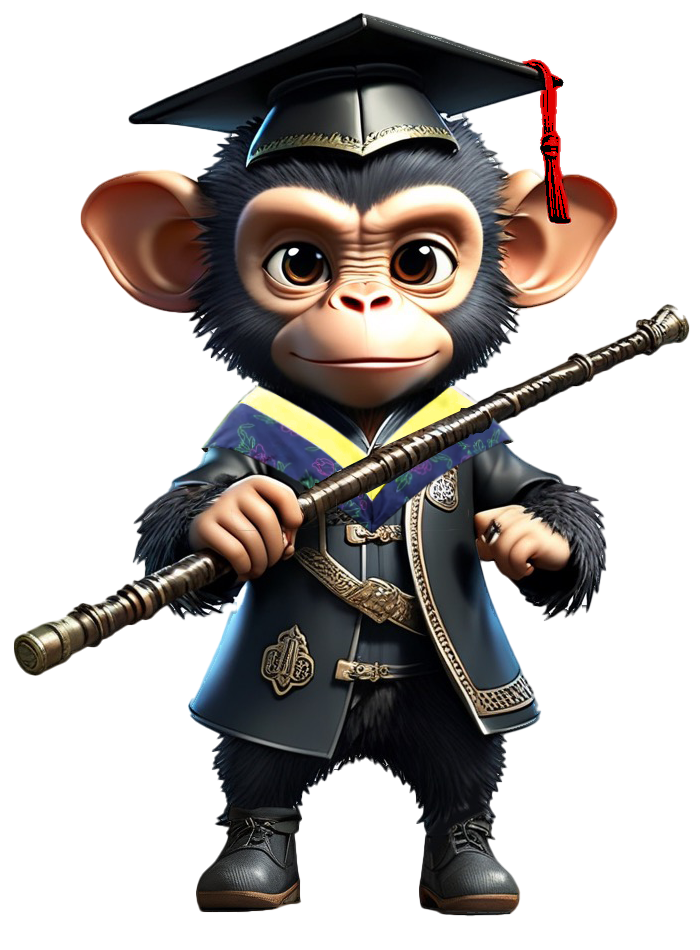}}}
$: An Advanced Large Vision-Language Dialogue Model Excels in Expert-Level Multimodal Table Reasoning}

\title{Table\textcolor{ruddy}{M}\textcolor{persianblue}{o}\textcolor{yellow}{E} 
\centering
$\vcenter{\hbox{\includegraphics[scale=0.022]{images/title/black_wukong_transparent_resized.jpg}}}
$: Mixture-of-Connector-Experts for Multimodal Table Understanding}

\title{Table\textcolor{ruddy}{M}\textcolor{persianblue}{o}\textcolor{yellow}{E} 
\centering
$\vcenter{\hbox{\includegraphics[scale=0.022]{images/title/black_wukong_transparent_resized.jpg}}}
$: Mixture-of-Connector-Experts Pretraining for Multimodal Table Understanding}

\title{Table\textcolor{ruddy}{M}\textcolor{persianblue}{o}\textcolor{yellow}{E} 
\centering
$\vcenter{\hbox{\includegraphics[scale=0.022]{images/title/black_wukong_transparent_resized.jpg}}}
$: Neuro-Symbolic Routing for Structured Expert Reasoning in Multimodal Table Understanding}

\author{
    Junwen Zhang \quad Pu Chen \quad Yin Zhang \\
  College of Computer Science and Technology\\
  Zhejiang University\\
  \texttt{junwenzhang@zju.edu.cn} \\
}

\begin{document}

\maketitle

\begin{abstract}
\label{abs_sec}
Multimodal understanding of tables in real-world contexts is challenging due to the complexity of structure, symbolic density, and visual degradation (blur, skew, watermarking, incomplete structures or fonts, multi-span or hierarchically nested layouts). Existing multimodal large language models (MLLMs) struggle with such \textit{WildStruct} conditions, resulting in limited performance and poor generalization. To address these challenges, we propose \textbf{TableMoE}, a neuro-symbolic Mixture-of-Connector-Experts (MoCE) architecture specifically designed for robust, structured reasoning over multimodal table data. TableMoE features an innovative \textbf{Neuro-Symbolic Routing} mechanism, which predicts latent semantic token roles (e.g., header, data cell, axis, formula) and dynamically routes table elements to specialized experts (Table-to-HTML, Table-to-JSON, Table-to-Code) using a confidence-aware gating strategy informed by symbolic reasoning graphs.

To facilitate effective alignment-driven pretraining, we introduce the large-scale \textbf{TableMoE-Align} dataset, consisting of 1.2M table–HTML–JSON–code quadruples across finance, science, biomedicine and industry, utilized exclusively for model pretraining. For evaluation, we curate and release four challenging \textit{WildStruct} benchmarks: \textbf{WMMFinQA}, \textbf{WMMTatQA}, \textbf{WMMTabDialog}, and \textbf{WMMFinanceMath}, designed specifically to stress-test models under real-world multimodal degradation and structural complexity. Experimental results demonstrate that TableMoE significantly surpasses existing state-of-the-art models, improving exact match and reasoning accuracy by up to 9.2\%. Extensive ablation studies validate each core component, emphasizing the critical role of Neuro-Symbolic Routing and structured expert alignment. Through qualitative analyses, we further showcase TableMoE's interpretability and enhanced robustness, underscoring the effectiveness of integrating neuro-symbolic reasoning for multimodal table understanding. The code 
and \textit{WildStruct} benchmarks are available at:
\newline \url{https://github.com/ai-agi/TableMoE}.
\newline \url{https://huggingface.co/datasets/darkme-ai/WMMTabDialog}.
\newline \url{https://huggingface.co/datasets/darkme-ai/WMMFinQA}.
\newline \url{https://huggingface.co/datasets/darkme-ai/WMMTatQA}.
\newline \url{https://huggingface.co/datasets/darkme-ai/WMMFinanceMath}.
\end{abstract}

\begin{figure}[ht]
\vskip -0.1in
\begin{center}
\includegraphics[width=0.7\columnwidth]{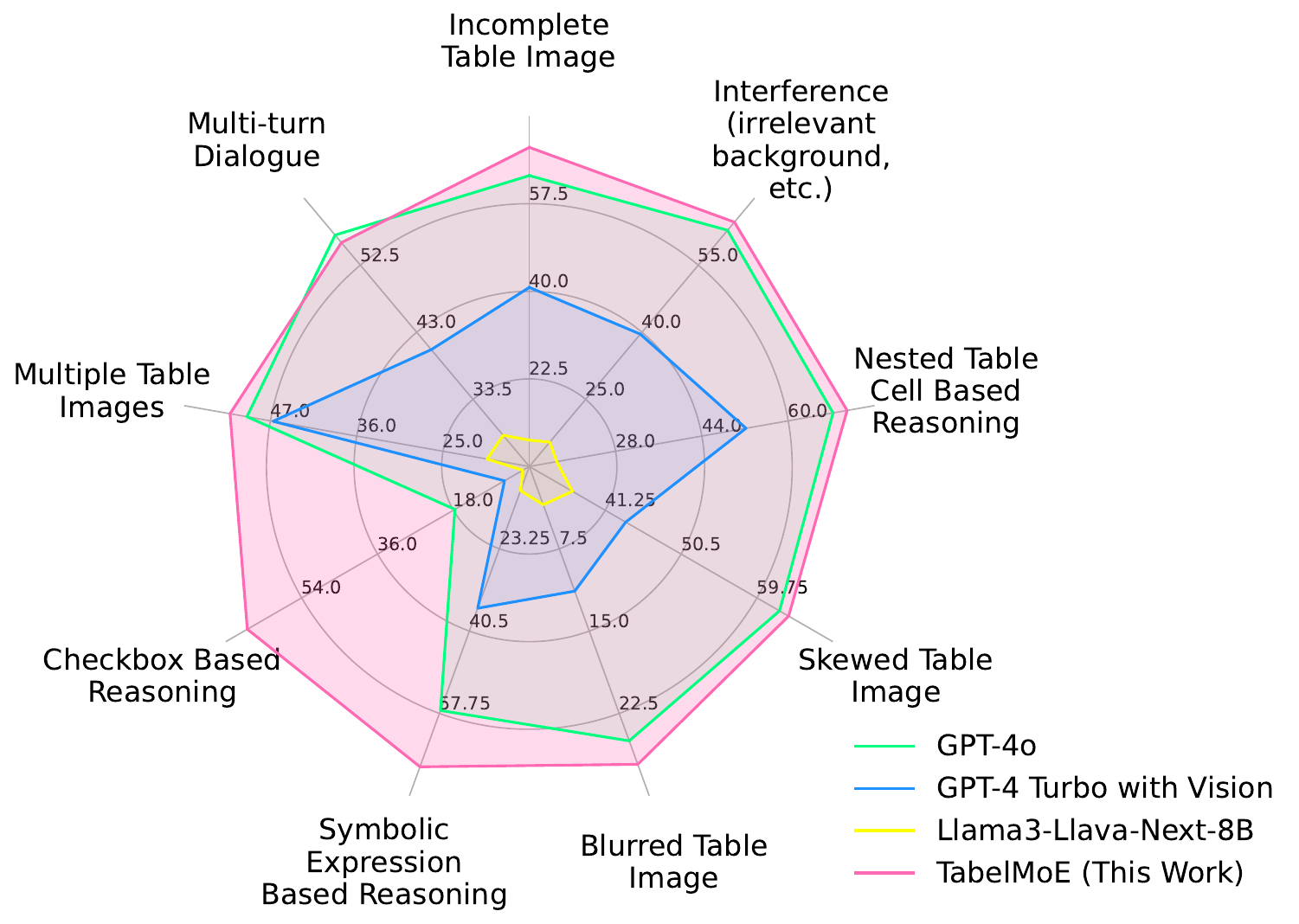}
\caption{Model comparison on WMMTabDialog-Bench for multimodal table reasoning.}
\label{fig:radar_chart_01}
\end{center}
\vskip -0.2in
\end{figure}

\section{Introduction}
Tables are a fundamental medium for communicating structured information in scientific discoveries \citep{zhong2019publaynetlargestdatasetdocument,zheng2020globaltableextractorgte,Smock_2022_CVPR}, financial reports\citep{chen-etal-2021-finqa,zhu-etal-2021-tat}, and industrial documents \citep{kweon2023openwikitabledatasetopendomain,deng2020turltableunderstandingrepresentation,li2020tablebankbenchmarkdatasettable}. They compactly encode text, numbers, and layouts, requiring both spatial perception and symbolic reasoning. Despite recent progress in multimodal large language models (MLLMs), robust understanding of real-world tables remains an open challenge. Unlike clean or synthetic benchmarks, tables in the wild often exhibit incomplete structures or fonts, symbolic clutter (e.g., formulas and equations), visual degradation (e.g., blur, skew, or watermarking), and complex nested cell hierarchies. We collectively define these phenomena as the \textit{WildStruct} challenge. This setting exposes a key limitation of existing MLLMs and vision-language systems: their architectures lack structural decomposition and fail to support interpretable reasoning over heterogeneous table elements.

To address this gap, we propose \textbf{TableMoE}, a \textit{Mixture-of-Connector-Experts} (MoCE) framework that introduces \textbf{Neuro-Symbolic Routing} for structured expert reasoning over visual tables. Unlike conventional token-level gating, our routing mechanism first predicts latent semantic roles, such as \texttt{HEADER}, \texttt{AXIS}, \texttt{DATA}, or \texttt{FORMULA}, for each token using a lightweight classifier. These roles define a symbolic reasoning graph that governs token-to-expert compatibility. Tokens are then routed to specialized connector experts through a soft neuro-symbolic compatibility function, with final routing weights calibrated using uncertainty-aware suppression. This process enables interpretable, function-aware expert activation under real-world table corruption.

Each expert is pre-trained on a distinct alignment task to capture complementary tabular semantics: (i) \textbf{Table-to-HTML} for structural layout parsing, (ii) \textbf{Table-to-JSON} for spatial triples and visual attributes, and (iii) \textbf{Table-to-Code} for executable rendering and symbolic abstraction. We initialize each expert connector via alignment-driven warm-up, following the MoE connector paradigm explored in ChartMoE \citep{xu2025chartmoemixturediverselyaligned}, Uni-Med \citep{zhu2024unimedunifiedmedicalgeneralist}, and CuMo \citep{li2024cumoscalingmultimodalllm}, but extend it with symbolic role-awareness and dynamic expert selection.

To support this framework, we release \textbf{TableMoE-Align}, a large-scale dataset comprising 1.2 million table–HTML–JSON–Code quadruples spanning finance, science, and biomedical domains. We further introduce four new \textit{WildStruct} benchmarks, \textbf{WMMFinQA}, \textbf{WMMTatQA}, \textbf{WMMTabDialog} and \textbf{WMMFinanceMath}, that evaluate model robustness under layout corruption and symbolic reasoning. On these and public datasets (e.g., FinanceMath and MMMU-Tables), TableMoE achieves new state-of-the-art results, improving exact match scores by up to \textbf{+9.2\%} over strong baselines.

Our results demonstrate that symbolic role prediction, combined with structure-aware expert routing, enables scalable, interpretable, and robust multimodal table understanding under real-world constraints.

\begin{itemize}
\item \textbf{Neuro-Symbolic Expert Routing}: We propose \textbf{TableMoE}, the first mixture-of-experts (MoE) framework tailored for table understanding that integrates symbolic structure graphs and latent token-level semantics (e.g., cell, axis, formula) into the routing process. This enables interpretable and structure-aware expert activation.

\item \textbf{Alignment-Driven Pretraining of Experts}: We introduce \textbf{TableMoE-Align}, a 1.2M-scale table-centric quadruple dataset across modalities (raw table, HTML, JSON, code). Each expert is pretrained via table-to-X alignment, equipping them with complementary spatial, structural, and symbolic reasoning capabilities.

\item \textbf{WildStruct Benchmark Protocol}: We establish the \textit{WildStruct} challenge to evaluate model robustness under realistic table degradation (e.g., highly nested layouts, symbolic clutter, missing structure or font)
, by constructing stress-test variants of existing public datasets.

\item \textbf{State-of-the-Art Performance and Interpretability}: TableMoE achieves up to \textbf{+9.2\%} absolute gains over strong baselines on FinanceMath, MMMU-Tables, and \textit{WildStruct}. Detailed ablations and expert visualizations demonstrate the effectiveness of neuro-symbolic routing and confidence-aware fusion.
\end{itemize}

\section{Related Work}

\subsection{\textit{WildStruct} Challenge}
\citet{long2021parsingtablestructureswild} introduce Cycle-CenterNet, enhancing geometric precision in table structure parsing (TSP) for images captured in uncontrolled environments. \citet{khang2025tfloptablestructurerecognition} propose TFLOP, a layout-pointer-based framework adept at parsing complex industrial tables, even amidst artifacts like watermarks. \citet{raja2021visualunderstandingcomplextable} present an object-detection model that captures long-range dependencies in tables with multi-span cells. Despite these advancements, a systematic taxonomy addressing real-world challenges in table image understanding, such as complex layouts and degraded visuals, remains underexplored. In practice, tables often exhibit partial structures, missing fonts, deep nesting, symbolic clutter, and visual noise like blur or skew. We collectively term these challenges as \textit{WildStruct}. To bridge this gap, we introduce four diagnostic benchmarks: WMMFinQA, WMMTatQA, WMMTabDialog, and WMMFinanceMath, designed to evaluate model robustness under real-world conditions. Refer to \Cref{fig:WMMFinQA_bench} and \Cref{fig:WMMTabDialog_bench} for illustrations.

\subsection{Multimodal Table Understanding and Reasoning} \label{subsec:mmtur}
Recent efforts have explored unified models for visual table understanding and reasoning. TabPedia~\citep{zhao2024tabpedia} encodes visual table tasks as conceptual tokens aligned via meditative prompts within an LLM, fostering generalizable multi-task capabilities. Table-LLaVA~\citep{zheng2024multimodaltableunderstanding} adopts a two-stage framework, structural table pretraining followed by instruction tuning, to better ground image-text pairs in QA and generation tasks. ReFocus~\citep{fu2025refocusvisualeditingchain} introduces visual chain-of-thought reasoning by editing table images to generate intermediate visual states, improving transparency and accuracy. Does Table Source Matter?~\citep{yang2025doestablesourcematter} demonstrates that domain-specific visual quality and adaptive resolution control significantly influence multimodal reasoning performance. Despite these advances, existing models rely heavily on monolithic architectures that entangle layout, semantics, and symbolic structure in a shared latent space, limiting generalization and interpretability. Mixture-of-Experts (MoE) architectures remain underutilized, with no prior work incorporating symbolic cues, such as token roles, table axes, or layout graphs, to guide structured expert routing. TableMoE addresses this gap through a neuro-symbolic MoE framework that explicitly decomposes multimodal table inputs into three alignment-driven expert pathways (\textsc{HTML}, \textsc{JSON}, \textsc{Code}). Its routing mechanism integrates role-aware semantic planning and confidence-aware fusion, enabling interpretable and robust reasoning over visually degraded, symbol-rich, and structurally incomplete tables. As the first to systematically target the \textit{WildStruct} challenge, TableMoE sets a new foundation for modular and scalable structured reasoning in real-world multimodal table understanding.

\subsection{MoE Connector}
CuMo \citep{li2024cumoscalingmultimodalllm} introduces a sparsely-gated MoE connector for multimodal vision-language models, where each expert is co-upcycled from a pre-trained MLP and selected via a Top-K router with auxiliary losses. This design improves visual-text alignment and boosts performance on instruction-following and VQA tasks without increasing inference cost.
Uni-Med \citep{zhu2024unimedunifiedmedicalgeneralist} introduces a connector mixture-of-experts (MoE) module with a a well-designed router to alleviate the tug-of-war problem for multi-task learning of MLLM in the medical field. ChartMoE \citep{xu2025chartmoemixturediverselyaligned} proposes a versatile MoE connector, where each expert is initialized from distinct pre-trained alignment tasks, replacing standard linear projectors with task-aligned experts. However, these methods rely on purely neural Top-K routing, which limits adaptability to noisy or structurally ambiguous inputs.
While MoE connectors have shown promise in other domains, their structured routing capabilities remain underexplored in the context of real-world tables. TableMoE addresses this gap by combining role-aware neuro-symbolic planning with uncertainty-weighted fusion for fine-grained expert specialization. TableMoE extends this line of work by introducing a Neuro-Symbolic Routing mechanism that integrates semantic token-role prediction with confidence-aware soft routing, enabling structured expert reasoning in complex, multimodal table scenarios.

\section{TableMoE}
\label{tab_moe}

\begin{figure}[htpb]
\begin{center}
\resizebox{0.8\columnwidth}{!}{%
  \includegraphics{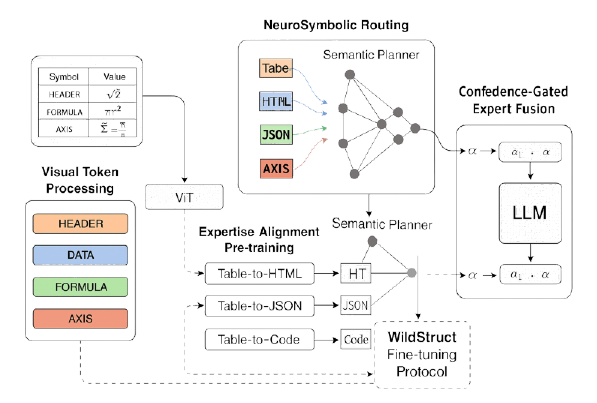}
}
\caption{Overview of the TableMoE architecture. The model encodes a table (image or structured) and text query, then uses Neuro-Symbolic Routing to ge specialized experts (HTML, JSON, Code) for reasoning. Outputs from experts are fused via a confidence-weighted mechanism, and final answers are produced by the LLM decoder. A semantic planner graph and token roles (header, data, etc.) guide the expert routing.}
\label{fig:arch}
\end{center}
\end{figure}

Figure~\ref{fig:arch} presents an overview of the TableMoE architecture. The model integrates a vision encoder, a Mixture-of-Connector-Experts (MoCE) module, with four specialized experts (HTML, JSON, Code, General) and a neuro-symbolic router, and an LLM backbone. 
Given a table image, the Vision Transformer (ViT) encodes visual tokens. On top of this representation, MoCE proceeds in two stages: (1) semantic role prediction and symbolic expert routing via the Neuro-Symbolic Router, and (2) expert-specific representation transformation. Then, the processed image and encoded text are fused into a joint embedding for final LLM decoding.



\begin{figure}[t]
\centering
\includegraphics[width=\linewidth]{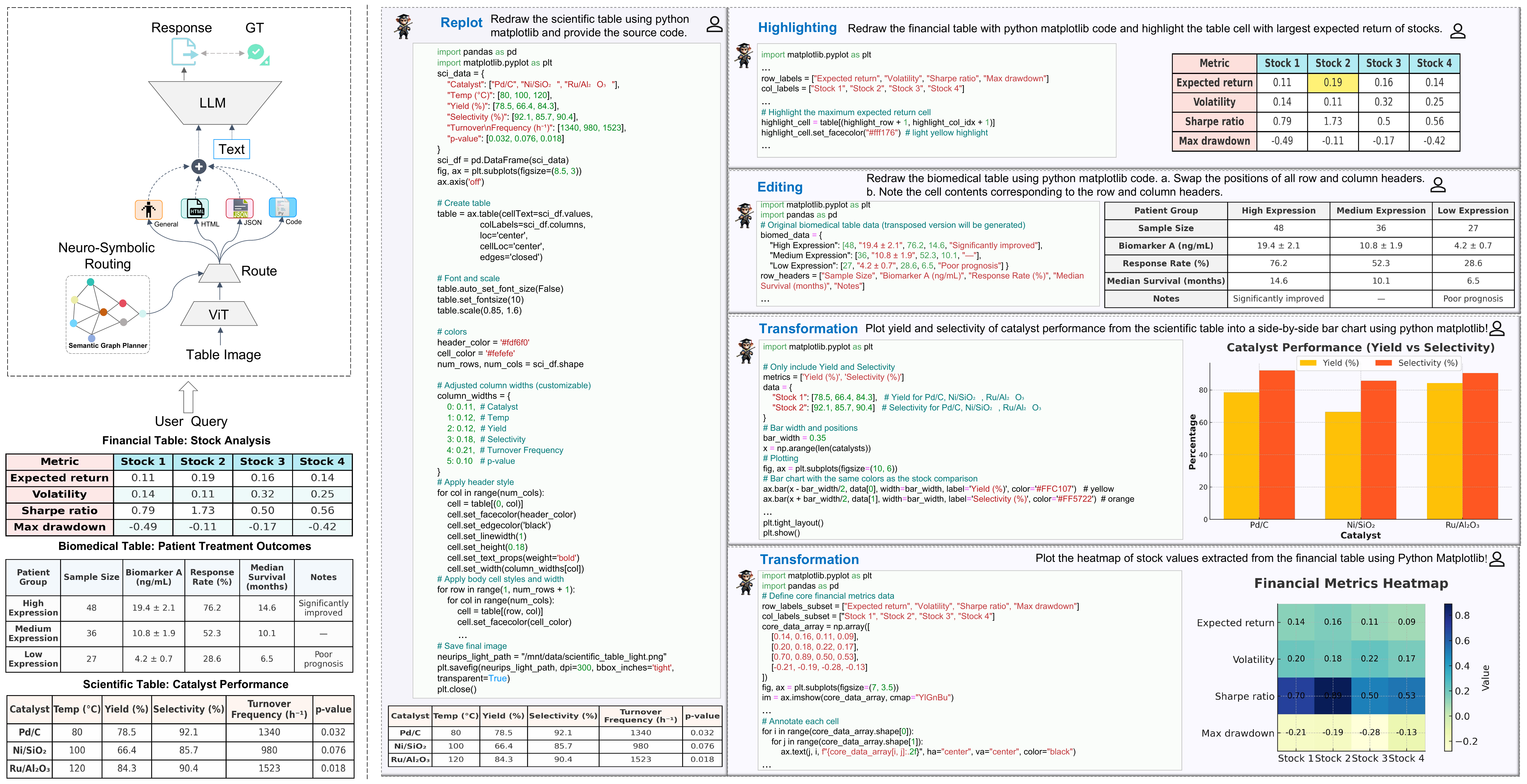}
\caption{Overview and capabilities of TableMoE. We introduce a Neuro-Symbolic MoE architecture that routes table tokens to specialized experts (HTML, JSON, Code, General) via a symbolic graph reasoning planner. Shown examples illustrate TableMoE’s ability to perform multimodal table understanding, such as highlighting, editing, replotting, and transformation, by generating executable \texttt{matplotlib} code from table images. 
}
\label{fig:tablemoe_overview}
\end{figure}


\subsection{Alignment-Driven Expert Pretraining}
We propose expert-aligned pretraining as a new paradigm for Mixture-of-Experts (MoE): each connector is pretrained on a distinct symbolic abstraction (HTML, JSON, or Code), transforming MoE into a modular neuro-symbolic reasoning engine. This disentangled initialization bypasses the tug-of-war in multi-task learning \citep{Hadsell2020EmbracingCC, zhu2024unimedunifiedmedicalgeneralist} and enables structured generalization across spatial, structural, and executable table semantics.
\\
\\
\textbf{Warm-Start Alignment}.
We initialize a two-layer MLP as the language-vision connector, following the warm-up recipe of CuMo \citep{li2024cumoscalingmultimodalllm}. Specifically, we pre-train a vanilla linear connector on LLaVA-558K \citep{liu2023visualinstructiontuning}, and refer to this as the \textbf{warm-start} connector. The resulting parameters serve as initialization for subsequent connector alignments tailored to specific tabular expertise. Additionally, we incorporate the warm-start connector as one of the multi-head connector experts, leveraging its general visual priors from natural image pretraining to enhance cross-domain generalization.
\\
\\
\textbf{Table-to-HTML Alignment}.
 Table-to-HTML alignment training is designed to help TableMoE understand table structures. In this stage, the model is required to convert visual tables into HTML structures along with all cell data. This alignment process enables the model to gain a deeper understanding of the structural information of tables and the relationships between each data point and the table's layout. 
\\
\\
\textbf{Table-to-JSON Alignment}.
This alignment comprises two tasks: (1) extracting JSON-formatted triples $\langle$row ID, column ID, cell value$\rangle$ from table images to assist TableMoE in understanding the spatial positions of table elements and their relative positional relationships, and (2) extracting fine-grained visual configurations (e.g., figure size, hex
color codes, font type, line style) to enable faithful table reconstruction via using \texttt{Matplotlib}.
\\
\\
\textbf{Table-to-Code Alignment}.
The alignment with code trains TableMoE to translate visual tables into executable Python \texttt{matplotlib} code, bridging low-level visual patterns with high-level symbolic abstraction. The model learns to generate structured scripts that faithfully capture tabular content, spatial layout, and visual attributes, including cell values, multi-row headers, and formatting cues. Beyond grid tables, the code spans diverse visualization types such as bar plots and heatmaps, adapting to both semantic and numeric structures. This alignment provides an interpretable and verifiable representation of visual tables, enhancing controllability in downstream editing, replotting, and reasoning tasks.


\subsection{Neuro-Symbolic Routing}
\label{sec:neuro-symbolic-routing}

We propose \textit{NeuroSym-MoE}, a neuro-symbolic routing paradigm that unifies neural perception and symbolic abstraction for robust multimodal table understanding. Rather than relying solely on visual embeddings, the model first predicts latent semantic roles for each token (e.g., \texttt{HEADER}, \texttt{DATA}, \texttt{FORMULA}, \texttt{AXIS}). These roles form a symbolic expert-routing graph through a compatibility mapping with expert types. This graph serves as a semantic planner that directs token-to-expert assignments, enabling interpretable expert specialization. The resulting structure-aware routing facilitates robust coordination and specialization across diverse layouts, especially under noisy or degraded table conditions.

To operationalize this, TableMoE introduces a Neuro-Symbolic Mixture-of-Connector-Experts (MoCE), which adaptively routes token embeddings to specialized experts based on semantic roles and confidence-aware gating. This design enhances interpretability, robustness, and modular reasoning.

Given a table image $I$ and accompanying text $T$, a vision encoder $f_v(\cdot)$ extracts token-level embeddings:
\begin{equation}
\mathbf{x} = f_v(I), \quad \mathbf{x} \in \mathbb{R}^{B \times N \times D}~.
\end{equation}

A role classifier $f_{\text{role}}$ predicts a soft role distribution over $R$ predefined table-centric roles for each token:
\begin{equation}
\tilde{\mathbf{r}}_i = \text{Softmax}(f_{\text{role}}(\mathbf{x}_i)), \quad \tilde{\mathbf{r}}_i \in \mathbb{R}^{R}~.
\end{equation}

These role vectors are mapped to expert affinity scores via a learned compatibility matrix $\mathbf{E} \in \mathbb{R}^{R \times E}$:
\begin{equation}
\mathbf{a}_i = \tilde{\mathbf{r}}_i \cdot \mathbf{E}, \quad \mathbf{a}_i \in \mathbb{R}^E~.
\end{equation}

To suppress noisy or ambiguous routing, we compute the normalized Shannon entropy of $\tilde{\mathbf{r}}_i$ as a confidence score:
\begin{equation}
\alpha_i = 1 - \frac{H(\tilde{\mathbf{r}}_i)}{\log R}, \quad H(\tilde{\mathbf{r}}_i) = -\sum_k \tilde{\mathbf{r}}_{i,k} \log \tilde{\mathbf{r}}_{i,k}~.
\end{equation}

This coefficient $\alpha_i$ scales the routing logits to modulate expert diversity under uncertainty:
\begin{equation}
\mathbf{w}_{i,e} = \frac{\exp(\alpha_i a_{i,e})}{\sum_{e'} \exp(\alpha_i a_{i,e'})}~.
\end{equation}

Each expert $f_e(\cdot)$ then processes $\mathbf{x}_i$ independently. The weighted outputs are fused as:
\begin{equation}
\hat{\mathbf{x}}_i = \sum_{e=1}^E \mathbf{w}_{i,e} \cdot f_e(\mathbf{x}_i)~.
\end{equation}

The fused token representation $\hat{\mathbf{x}}$ is concatenated with the encoded textual prompt $\mathbf{t} = f_t(T)$ and passed into an autoregressive decoder for final output. This hybrid mechanism grounds symbolic layout perception in neural expert dynamics, yielding interpretable and layout-robust multimodal reasoning.

\begin{algorithm}[t]
\caption{Neuro-Symbolic Routing Forward Pass}
\label{alg:ns-routing}
\begin{algorithmic}[1]  
\Require Table image $I$, text $T$, vision encoder $f_v$, role classifier $f_{\text{role}}$, compatibility matrix $\mathbf{E}$, expert set $\{f_e\}_{e=1}^E$, text encoder $f_t$, LLM decoder $\text{Decoder}$
\State $\mathbf{x} \leftarrow f_v(I)$ \Comment{Extract visual tokens $\in \mathbb{R}^{B \times N \times D}$}
\State $\tilde{\mathbf{R}} \leftarrow \text{Softmax}(f_{\text{role}}(\mathbf{x}))$ \Comment{Predict token role probabilities}
\State $\mathbf{A} \leftarrow \tilde{\mathbf{R}} \cdot \mathbf{E}$ \Comment{Compute expert affinities}
\State $\boldsymbol{\alpha} \leftarrow 1 - \text{Entropy}(\tilde{\mathbf{R}}) / \log R$ \Comment{Estimate confidence suppression}
\State $\mathbf{W} \leftarrow \text{Softmax}(\boldsymbol{\alpha} \odot \mathbf{A})$ \Comment{Final routing weights}
\For{$e = 1$ to $E$}
    \State $\mathbf{Y}^{(e)} \leftarrow f_e(\mathbf{x})$ \Comment{Apply expert transformation}
\EndFor
\State $\hat{\mathbf{x}} \leftarrow \sum_e \mathbf{W}_{(:,e)} \odot \mathbf{Y}^{(e)}$ \Comment{Fuse expert outputs}
\State $\mathbf{t} \leftarrow f_t(T)$ \Comment{Encode accompanying text}
\State \Return $\text{Decoder}([\hat{\mathbf{x}}; \mathbf{t}])$ \Comment{LLM decoding for output}
\end{algorithmic}
\end{algorithm}

\subsection{Curriculum-Guided Neuro-Symbolic Annealing}
\label{sec:annealing}

To bridge the gap between synthetic table understanding and real-world complexity, we introduce \textbf{Curriculum-Guided Neuro-Symbolic Annealing}, a fine-tuning protocol that progressively escalates structural complexity and dynamically refines symbolic expert coordination through entropy-aware routing. Inspired by recent advances in curriculum learning \citep{zhuang2024mathpumaprogressiveupwardmultimodal, wang2021surveycurriculumlearning, wu2021curriculawork} and uncertainty-based expert modulation \citep{vanlioglu2025entropyguidedsequenceweightingefficient, qu2024emoeexpansivematchingexperts, schroder-etal-2022-revisiting}, this strategy reinforces robustness and interpretability in multimodal table reasoning. Effective learning in this stage hinges on the availability of high-quality, semantically rich training data that faithfully reflects structural diversity.

\textbf{Progressive \textit{WildStruct} Exposure.} Fine-tuning begins with clean, structured tables (e.g., HTML, JSON) and incrementally introduces degradation (e.g., OCR artifacts, missing headers, symbolic clutter). As the difficulty increases, the neuro-symbolic router adapts expert reliance—for example, defaulting to the warm-start expert when layout collapses or shifting toward the HTML expert when structural reorganization is beneficial.

\textbf{Annealed Routing Confidence.} To stabilize symbolic routing under noise, we anneal the entropy-based confidence suppression term from Section~\ref{sec:neuro-symbolic-routing}, gradually amplifying the influence of high-confidence expert paths over the course of fine-tuning. This annealing controls the impact of the confidence-aware planner, allowing the model to transition from generalist exploration to expert specialization.

\textbf{Program-of-Thought as Primary Supervision.} We supervise the code expert using Program-of-Thought (PoT) execution traces \citep{wang2023pot}, which reflect interpretable and verifiable symbolic reasoning over table semantics. Additionally, we inject Chain-of-Thought (CoT) rationales \citep{NEURIPS2022_9d560961} as auxiliary decoder supervision, encouraging concise textual justifications. Ablations reveal that PoT yields superior symbolic grounding and robustness under structural degradation, while CoT contributes complementary gains in transparency.

Our protocol is end-to-end differentiable and free of test-time rules or solvers. Together, these mechanisms guide TableMoE toward emergent symbolic generalization—scalable, interpretable, and resilient to noisy real-world tabular data.

\section{Dataset Construction and Benchmark Design}\label{data_construct}
To facilitate training and evaluation of TableMoE, we construct two novel resources: the \textbf{TableMoE-Align} pretraining corpus and the \textbf{\textit{WildStruct}} fine-tuning and benchmark suite.
\subsection{\textbf{TableMoE-Align}}
We construct the alignment corpus by aggregating tabular images and data from FinTabNet~\citep{zheng2020globaltableextractorgte}, PubTabNet~\citep{zhong2020imagebasedtablerecognitiondata}, TableBank~\citep{li2020tablebankbenchmarkdatasettable}, and WTW \citep{long2021parsingtablestructureswild}, covering a wide spectrum of table structures and rendering noise. Dataset statistics are summarized in \Cref{tab:tablemoe_align_data}.

\subsection{\textit{WildStruct} Neuro-Symbolic Annealing}
We construct \textit{WildStruct}-style fine-tuning sets from the FinQA \citep{chen-etal-2021-finqa} and TAT-QA \citep{zhu-etal-2021-tat} training corpora, yielding WMMFinQA  and WMMTatQA , tailored for neuro-symbolic annealing with progressive degradation curricula. Dataset composition is summarized in Table~\ref{tab:nsa_training_data}. For all \textit{WildStruct} sets, we employ split and evaluation metrics similar to the originals.

\subsection{\textit{WildStruct} Benchmark Suite}
As existing QA benchmarks assume idealized layouts, we introduce \textbf{WildStruct}, a stress-test suite featuring real-world degradations in layout, structure, and semantics. The dataset statistics are shown in \Cref{tab:benchmarks}. See \Cref{subsec:w&p_bench} for benchmark curation details.

\section{Experiments}
We conduct extensive experiments to evaluate TableMoE against state-of-the-art methods. We first describe our experimental setup, including baseline models and implementation details, then present results on various benchmarks, followed by ablation studies.

\subsection{Experiment settings}
\textbf{Implementation details} 
We employ two instruction-tuned large language models as the backbone of TableMoE: LLaMA 3.1–8B–Instruct \citep{grattafiori2024llama3herdmodels} and Qwen2.5–VL–7B–Instruct \citep{bai2025qwen25vltechnicalreport}. For the visual encoder at first type backbone, we employ CLIP-ViT-Large-Patch14-336 \citep{radford2021learningtransferablevisualmodels}.

We observed that some input sequences during MLP connector pretraining exceeded 4096 tokens, prompting us to increase the LLM's maximum length to 8192 for both training and inference. 

\vspace{0.5em}
\noindent
\textbf{Evaluation Benchmarks}
 We benchmark TableMoE and all baselines on test datasets, as summarized in \Cref{tab:benchmarks}.
 
\vspace{0.5em}
\noindent
\textbf{Evaluation Metrics}
We evaluate performance using exact match for numerical answers (within 0.5\% relative error) and SBERT-based semantic similarity \citep{reimers2019sentencebertsentenceembeddingsusing} for textual answers. Final answers are extracted via the pipeline from \citet{chen2023theoremqatheoremdrivenquestionanswering} and compared to ground truth. All results are averaged over three runs. To ensure reliability, 10\% of predictions are manually audited to resolve ambiguous or failure-prone cases.

\begin{table}[t]
\centering

\begin{minipage}{0.48\linewidth}
\centering
\caption{\textbf{Results on \textit{WildStruct}-Bench.} TableMoE shows robustness under structural and visual degradation.}
\label{tab:wildstruct}
\resizebox{\linewidth}{!}{
\begin{tabular}{l|ccc|c}
\toprule
\makecell[c]{\multirow{2}{*}[-1.0ex]{\hspace*{1.5em} Method}} & \multicolumn{3}{c|}{\textit{WildStruct}-Bench} & \makecell[c]{\multirow{2}{*}[-1.0ex]{\hspace*{0.5em}Avg}} \\
\cmidrule(lr){2-4} 
 & WMMTabDialog & WMMFinQA & WMMTatQA & \\
\midrule
Human Expert & 87.44 & 85.96 & 90.98 & 88.13 \\
\midrule
LLaVA-Llama-3-8B & 14.72 & 18.96 & 21.47 & 18.38 \\
TabPedia (7B) & 4.69 & 8.87 & 5.76 & 6.44 \\
Table-LLaVA (7B) & 13.02 & 15.40 & 10.26 & 12.89 \\
InternVL2\_5-78B & 19.05 & 38.45 & 66.51 & 41.34 \\
Qwen2-VL-72B & 21.39 & 36.53 & 70.48 & 42.80 \\
Qwen2.5-VL-72B & 36.94 & 75.41 & 81.78 & 64.71 \\
GPT-4 Turbo & 40.39 & 73.93 & 83.94 & 66.09 \\
GPT-4o & 63.10 & 77.16 & \cellcolor{whitesmoke}84.97 & 75.08 \\
\midrule
\rowcolor{lightcyan}TableMoE+CoT & \cellcolor{whitesmoke}70.13 & \cellcolor{whitesmoke}78.73 & 83.28 & \cellcolor{whitesmoke}77.38 \\
\rowcolor{lightcyan}TableMoE+PoT & \cellcolor{gainsboro}72.25 & \cellcolor{gainsboro}81.98 & \cellcolor{gainsboro}86.94 & \cellcolor{gainsboro}80.31 \\
\bottomrule
\end{tabular}
}
\end{minipage}
\hfill
\begin{minipage}{0.48\linewidth}
\centering
\caption{\textbf{Results on MMMU-Table.} TableMoE demonstrates domain-generalized table understand and reasoning.}
\label{tab:mmmu}
\resizebox{\linewidth}{!}{
\begin{tabular}{l|ccc|c}
\toprule
\makecell[c]{\multirow{2}{*}[-1.0ex]{\hspace*{1.5em} Method}} & \multicolumn{3}{c|}{MMMU (Table)} & \makecell[c]{\multirow{2}{*}[-1.0ex]{\hspace*{0.5em}Avg}} \\
\cmidrule(lr){2-4} 
 & Finance & Economics & Others & \\
\midrule
Qwen2-VL-72B & 33.94 & 51.46 & 43.23 & 42.88 \\
GPT-4o & 35.77 & 51.46 & 44.72 & 43.98 \\
\rowcolor{lightcyan}TableMoE & 50.91 & 57.31 & 55.06 & 54.53 \\
\bottomrule
\end{tabular}
}
\end{minipage}
\end{table}

\subsection{Main Results}
\noindent
As shown in Table~\ref{tab:wildstruct}, TableMoE consistently outperforms all strong vision-language baselines on the \textit{WildStruct-Bench}, achieving an average accuracy of 80.31 with program-of-thought (PoT) reasoning. This surpasses GPT-4o by +5.2 points and narrows the gap to human expert performance to within 8 points. Notably, TableMoE+PoT attains new state-of-the-art scores on WMMFinQA (81.98) and WMMTatQA (86.94), reflecting robust numerical and symbolic reasoning. Even under dialog-based interactions (WMMTabDialog), TableMoE+CoT yields 70.13, outperforming GPT-4o by a clear +7.0 margin. These gains highlight the strength of our neuro-symbolic routing and alignment-aware MoE pretraining in handling structurally degraded and semantically complex tables. Moreover, as shown in Table~\ref{tab:mmmu}, TableMoE achieves an average accuracy of 54.53\% on the MMMU-Table benchmark, outperforming GPT-4o by +10.55 points. This substantial margin underscores TableMoE's robust generalization capabilities across diverse domains and complex table reasoning tasks.

\begin{table}[h]
\caption{Ablation results on WMMFinQA (test). Removing any component of TableMoE degrades performance, especially neuro-symbolic routing and expert fusion.}
\label{tab:ablation}
\centering
\resizebox{0.56\linewidth}{!}{%
\begin{tabular}{lcc}
\toprule
\textbf{Model Variant} & Avg (\%) & $\Delta$ from full \\
\midrule
Full TableMoE model & 81.98 & -- \\
\;\; w/o NSA (neural-only fine-tuning) & 78.28 & $-3.70$ \\
\;\; w/o symbolic graph (neural routing only) & 76.61 & $-5.37$ \\
\;\; w/o token role prediction & 77.51 & $-4.47$ \\
\;\; w/o confidence fusion (equal weights) & 75.71 & $-6.27$ \\
\;\; w/o code expert & 75.01 & $-6.97$ \\
\;\; w/o HTML expert & 77.31 & $-4.67$ \\
\;\; w/o JSON expert & 79.01 & $-2.97$ \\
\;\; w/o any expert (LLM only) & 63.51 & $-18.47$ \\
\bottomrule
\end{tabular}%
}
\end{table}

\subsection{Ablation Study}
We conduct an ablation study on WMMFinQA to assess the contribution of each component in TableMoE. As shown in Table~\ref{tab:ablation}, removing any module results in performance degradation, with the most significant drops attributed to neuro-symbolic routing and expert fusion mechanisms.

Removing the \textbf{symbolic routing graph}, replacing it with purely neural routing, reduces performance by 3.7 points, highlighting the importance of role-aware expert selection. Further, ablating only the \textit{role predictor} yields a 2.8-point drop, while removing the graph structure alone causes a 3.0-point loss. These results underscore that symbolic token roles and structured compatibility are both necessary for informed routing.

Disabling the \textbf{confidence-aware fusion} (i.e., using uniform weights) leads to a 4.6-point decline. In this variant, the model often over-relies on erroneous expert outputs, such as overconfident but incorrect code generations, resulting in brittle behavior. The entropy-based suppressor mitigates such failure cases by dynamically attenuating low-confidence outputs.

Among the individual \textbf{connector experts}, ablating the code expert yields the largest drop ($-5.3$), reflecting its centrality for executing symbolic-numeric reasoning steps. Removing the HTML expert causes a 3.0-point decline, as structural alignment is essential for row-header and layout-based reasoning. Removing the JSON expert causes only a minor drop (1.3), suggesting its role may be partially subsumed by HTML on this task, though it may prove more critical in other settings.

Finally, removing the entire expert module, leaving only the base LLM, leads to a dramatic 16.8-point drop. This confirms that TableMoE's modular expert architecture is not only beneficial but essential for high-fidelity table reasoning.

In summary, the ablation confirms that each design element of TableMoE, neuro-symbolic routing, uncertainty-aware fusion, and expert diversity, contributes meaningfully to performance. Together, they enable robust reasoning under structural complexity.

\subsection{Qualitative Analysis and Discussion}
To enhance interpretability for human readers, we present CoT-style outputs in these case study examples rather than raw PoT code traces. While PoT yields executable reasoning steps, CoT rationales are more intuitive and accessible to non-technical audiences.
\vspace{0.5em}
\noindent
\paragraph{Precision in Multilingual Symbolic Reasoning.}
Table~\ref{tab:wmmtab_rsn_chat_01} showcases a complex Chinese financial bookkeeping table requiring precise symbolic reasoning and aggregation. The task involves summing debits and credits and expressing the result in uppercase Chinese currency—both numerically exact and linguistically constrained. All baseline MLLMs, including Qwen2-VL and GPT-4o, fail, hallucinating inflated totals due to visual misreading and weak symbolic grounding. In contrast, TableMoE correctly resolves \begin{CJK*}{UTF8}{gbsn}\small 壹拾柒万捌仟贰佰叁拾玖元玖角玖分\end{CJK*} (¥178,239.99), guided by low-entropy role predictions (e.g., \begin{CJK*}{UTF8}{gbsn}\small 合计\end{CJK*}) and confident Code expert activation. This example demonstrates TableMoE’s cross-lingual precision, symbolic abstraction, and robustness in culturally specific, visually degraded tabular settings.

\textbf{Robustness Under Blur and Symbolic Clutter.}
In Table~\ref{tab:WMMTatQA_inf_rsn_03}, we evaluate model responses to a heavily blurred financial table. Qwen2.5-VL-72B confidently generates hallucinated revenue figures and fabricated fiscal years absent from the image, underscoring its lack of uncertainty modeling and gating mechanisms. Conversely, GPT-4o and our TableMoE both refrain from conjecture, instead requesting clarification. Notably, TableMoE's abstention stems from a symbolic decision boundary: its neuro-symbolic routing induces high entropy in role predictions, activating confidence suppression and redirecting processing to the warm-start connector. This mechanism enables TableMoE to recognize uncertainty and avoid erroneous commitments. Such behavior emerges from our entropy-aware annealing strategy and confidence-gated MoE design, demonstrating robustness in scenarios that challenge traditional visual feature extractors. Further analysis is provided in Appendix~\ref{app:ana}.

\section{Conclusion}
We presented TableMoE, a neuro-symbolic mixture-of-experts model for multimodal table understanding. By combining alignment-driven expert pretraining, symbolic-aware routing, and confidence-weighted fusion, TableMoE achieves new state-of-the-art results on diverse table reasoning tasks, including challenging settings with visual and noisy tables. The model not only improves accuracy but also provides interpretable intermediate representations (HTML, JSON, code) that shed light on its reasoning process. This work demonstrates the promise of structured expert reasoning in enhancing the capabilities of large language models for complex data understanding tasks.


\clearpage
\bibliography{references}

\begin{thebibliography}{40}
\providecommand{\natexlab}[1]{#1}
\providecommand{\url}[1]{\texttt{#1}}
\expandafter\ifx\csname urlstyle\endcsname\relax
  \providecommand{\doi}[1]{doi: #1}\else
  \providecommand{\doi}{doi: \begingroup \urlstyle{rm}\Url}\fi

\bibitem[Aaron~Grattafiori and etal.(2024)]{grattafiori2024llama3herdmodels}
A.~D. Aaron~Grattafiori and etal.
\newblock The llama 3 herd of models, 2024.
\newblock URL \url{https://arxiv.org/abs/2407.21783}.

\bibitem[Chen et~al.(2023)Chen, Yin, Ku, Lu, Wan, Ma, Xu, Wang, and Xia]{chen2023theoremqatheoremdrivenquestionanswering}
W.~Chen, M.~Yin, M.~Ku, P.~Lu, Y.~Wan, X.~Ma, J.~Xu, X.~Wang, and T.~Xia.
\newblock Theoremqa: A theorem-driven question answering dataset, 2023.
\newblock URL \url{https://arxiv.org/abs/2305.12524}.

\bibitem[Chen et~al.(2021)Chen, Chen, Smiley, Shah, Borova, Langdon, Moussa, Beane, Huang, Routledge, and Wang]{chen-etal-2021-finqa}
Z.~Chen, W.~Chen, C.~Smiley, S.~Shah, I.~Borova, D.~Langdon, R.~Moussa, M.~Beane, T.-H. Huang, B.~Routledge, and W.~Y. Wang.
\newblock {F}in{QA}: A dataset of numerical reasoning over financial data.
\newblock In \emph{Proceedings of the 2021 Conference on Empirical Methods in Natural Language Processing}, pages 3697--3711. Association for Computational Linguistics, Nov. 2021.
\newblock \doi{10.18653/v1/2021.emnlp-main.300}.
\newblock URL \url{https://aclanthology.org/2021.emnlp-main.300}.

\bibitem[Deng et~al.(2020)Deng, Sun, Lees, Wu, and Yu]{deng2020turltableunderstandingrepresentation}
X.~Deng, H.~Sun, A.~Lees, Y.~Wu, and C.~Yu.
\newblock Turl: Table understanding through representation learning, 2020.
\newblock URL \url{https://arxiv.org/abs/2006.14806}.

\bibitem[Fu et~al.(2025)Fu, Liu, Yang, Corring, Lu, Yang, Roth, Florencio, and Zhang]{fu2025refocusvisualeditingchain}
X.~Fu, M.~Liu, Z.~Yang, J.~Corring, Y.~Lu, J.~Yang, D.~Roth, D.~Florencio, and C.~Zhang.
\newblock Refocus: Visual editing as a chain of thought for structured image understanding, 2025.
\newblock URL \url{https://arxiv.org/abs/2501.05452}.

\bibitem[Hadsell et~al.(2020)Hadsell, Rao, Rusu, and Pascanu]{Hadsell2020EmbracingCC}
R.~Hadsell, D.~Rao, A.~A. Rusu, and R.~Pascanu.
\newblock Embracing change: Continual learning in deep neural networks.
\newblock \emph{Trends in Cognitive Sciences}, 24:\penalty0 1028--1040, 2020.
\newblock URL \url{https://api.semanticscholar.org/CorpusID:226240885}.

\bibitem[Khang and Hong(2025)]{khang2025tfloptablestructurerecognition}
M.~Khang and T.~Hong.
\newblock Tflop: Table structure recognition framework with layout pointer mechanism, 2025.
\newblock URL \url{https://arxiv.org/abs/2501.11800}.

\bibitem[Kweon et~al.(2023)Kweon, Kwon, Cho, Jo, and Choi]{kweon2023openwikitabledatasetopendomain}
S.~Kweon, Y.~Kwon, S.~Cho, Y.~Jo, and E.~Choi.
\newblock Open-wikitable: Dataset for open domain question answering with complex reasoning over table, 2023.
\newblock URL \url{https://arxiv.org/abs/2305.07288}.

\bibitem[Li et~al.(2024)Li, Wang, Zhu, Kuo, Xu, Chen, Jain, Shi, and Wen]{li2024cumoscalingmultimodalllm}
J.~Li, X.~Wang, S.~Zhu, C.-W. Kuo, L.~Xu, F.~Chen, J.~Jain, H.~Shi, and L.~Wen.
\newblock Cumo: Scaling multimodal llm with co-upcycled mixture-of-experts, 2024.
\newblock URL \url{https://arxiv.org/abs/2405.05949}.

\bibitem[Li et~al.(2020)Li, Cui, Huang, Wei, Zhou, and Li]{li2020tablebankbenchmarkdatasettable}
M.~Li, L.~Cui, S.~Huang, F.~Wei, M.~Zhou, and Z.~Li.
\newblock Tablebank: A benchmark dataset for table detection and recognition, 2020.
\newblock URL \url{https://arxiv.org/abs/1903.01949}.

\bibitem[Liu et~al.(2023)Liu, Li, Wu, and Lee]{liu2023visualinstructiontuning}
H.~Liu, C.~Li, Q.~Wu, and Y.~J. Lee.
\newblock Visual instruction tuning, 2023.
\newblock URL \url{https://arxiv.org/abs/2304.08485}.

\bibitem[Long et~al.(2021)Long, Wang, Xue, Gao, Yang, Wang, and Xia]{long2021parsingtablestructureswild}
R.~Long, W.~Wang, N.~Xue, F.~Gao, Z.~Yang, Y.~Wang, and G.-S. Xia.
\newblock Parsing table structures in the wild, 2021.
\newblock URL \url{https://arxiv.org/abs/2109.02199}.

\bibitem[Masry et~al.(2022)Masry, Long, Tan, Joty, and Hoque]{masry2022chartqabenchmarkquestionanswering}
A.~Masry, D.~X. Long, J.~Q. Tan, S.~Joty, and E.~Hoque.
\newblock Chartqa: A benchmark for question answering about charts with visual and logical reasoning, 2022.
\newblock URL \url{https://arxiv.org/abs/2203.10244}.

\bibitem[OpenAI(2024)]{openai2024gpt-4o}
OpenAI.
\newblock Hello gpt-4o, May 2024.
\newblock URL \url{https://openai.com/index/hello-gpt-4o}.

\bibitem[Qu et~al.(2024)Qu, Wellnitz, Tropsha, and Oliva]{qu2024emoeexpansivematchingexperts}
Y.~Qu, J.~Wellnitz, A.~Tropsha, and J.~Oliva.
\newblock Emoe: Expansive matching of experts for robust uncertainty based rejection, 2024.
\newblock URL \url{https://arxiv.org/abs/2406.01825}.

\bibitem[Radford et~al.(2021)Radford, Kim, Hallacy, Ramesh, Goh, Agarwal, Sastry, Askell, Mishkin, Clark, Krueger, and Sutskever]{radford2021learningtransferablevisualmodels}
A.~Radford, J.~W. Kim, C.~Hallacy, A.~Ramesh, G.~Goh, S.~Agarwal, G.~Sastry, A.~Askell, P.~Mishkin, J.~Clark, G.~Krueger, and I.~Sutskever.
\newblock Learning transferable visual models from natural language supervision, 2021.
\newblock URL \url{https://arxiv.org/abs/2103.00020}.

\bibitem[Raja et~al.(2021)Raja, Mondal, and Jawahar]{raja2021visualunderstandingcomplextable}
S.~Raja, A.~Mondal, and C.~V. Jawahar.
\newblock Visual understanding of complex table structures from document images, 2021.
\newblock URL \url{https://arxiv.org/abs/2111.07129}.

\bibitem[Reimers and Gurevych(2019)]{reimers2019sentencebertsentenceembeddingsusing}
N.~Reimers and I.~Gurevych.
\newblock Sentence-bert: Sentence embeddings using siamese bert-networks, 2019.
\newblock URL \url{https://arxiv.org/abs/1908.10084}.

\bibitem[Schr{\"o}der et~al.(2022)Schr{\"o}der, Niekler, and Potthast]{schroder-etal-2022-revisiting}
C.~Schr{\"o}der, A.~Niekler, and M.~Potthast.
\newblock Revisiting uncertainty-based query strategies for active learning with transformers.
\newblock In S.~Muresan, P.~Nakov, and A.~Villavicencio, editors, \emph{Findings of the Association for Computational Linguistics: ACL 2022}, pages 2194--2203, Dublin, Ireland, May 2022. Association for Computational Linguistics.
\newblock \doi{10.18653/v1/2022.findings-acl.172}.
\newblock URL \url{https://aclanthology.org/2022.findings-acl.172/}.

\bibitem[Shuai~Bai and etal.(2025)]{bai2025qwen25vltechnicalreport}
K.~C. Shuai~Bai and etal.
\newblock Qwen2.5-vl technical report, 2025.
\newblock URL \url{https://arxiv.org/abs/2502.13923}.

\bibitem[Smock et~al.(2022)Smock, Pesala, and Abraham]{Smock_2022_CVPR}
B.~Smock, R.~Pesala, and R.~Abraham.
\newblock Pubtables-1m: Towards comprehensive table extraction from unstructured documents.
\newblock In \emph{Proceedings of the IEEE/CVF Conference on Computer Vision and Pattern Recognition (CVPR)}, pages 4634--4642, June 2022.

\bibitem[Vanlioglu(2025)]{vanlioglu2025entropyguidedsequenceweightingefficient}
A.~Vanlioglu.
\newblock Entropy-guided sequence weighting for efficient exploration in rl-based llm fine-tuning, 2025.
\newblock URL \url{https://arxiv.org/abs/2503.22456}.

\bibitem[Wang et~al.(2023{\natexlab{a}})]{wang2023pot}
B.~Wang et~al.
\newblock Program of thoughts prompting: Disentangling computation from reasoning for numerical reasoning tasks.
\newblock \emph{arXiv preprint arXiv:2305.10601}, 2023{\natexlab{a}}.

\bibitem[Wang et~al.(2021)Wang, Chen, and Zhu]{wang2021surveycurriculumlearning}
X.~Wang, Y.~Chen, and W.~Zhu.
\newblock A survey on curriculum learning, 2021.
\newblock URL \url{https://arxiv.org/abs/2010.13166}.

\bibitem[Wang et~al.(2023{\natexlab{b}})Wang, Wei, Schuurmans, Le, Chi, Narang, Chowdhery, and Zhou]{wang2023selfconsistencyimproveschainthought}
X.~Wang, J.~Wei, D.~Schuurmans, Q.~Le, E.~Chi, S.~Narang, A.~Chowdhery, and D.~Zhou.
\newblock Self-consistency improves chain of thought reasoning in language models, 2023{\natexlab{b}}.
\newblock URL \url{https://arxiv.org/abs/2203.11171}.

\bibitem[Wei et~al.(2022)Wei, Wang, Schuurmans, Bosma, ichter, Xia, Chi, Le, and Zhou]{NEURIPS2022_9d560961}
J.~Wei, X.~Wang, D.~Schuurmans, M.~Bosma, b.~ichter, F.~Xia, E.~Chi, Q.~V. Le, and D.~Zhou.
\newblock Chain-of-thought prompting elicits reasoning in large language models.
\newblock In S.~Koyejo, S.~Mohamed, A.~Agarwal, D.~Belgrave, K.~Cho, and A.~Oh, editors, \emph{Advances in Neural Information Processing Systems}, volume~35, pages 24824--24837. Curran Associates, Inc., 2022.
\newblock URL \url{https://proceedings.neurips.cc/paper_files/paper/2022/file/9d5609613524ecf4f15af0f7b31abca4-Paper-Conference.pdf}.

\bibitem[Wu et~al.(2021)Wu, Dyer, and Neyshabur]{wu2021curriculawork}
X.~Wu, E.~Dyer, and B.~Neyshabur.
\newblock When do curricula work?, 2021.
\newblock URL \url{https://arxiv.org/abs/2012.03107}.

\bibitem[Xu et~al.(2025)Xu, Qu, Qi, Du, Xu, Yuan, and Guo]{xu2025chartmoemixturediverselyaligned}
Z.~Xu, B.~Qu, Y.~Qi, S.~Du, C.~Xu, C.~Yuan, and J.~Guo.
\newblock Chartmoe: Mixture of diversely aligned expert connector for chart understanding, 2025.
\newblock URL \url{https://arxiv.org/abs/2409.03277}.

\bibitem[Yang et~al.(2025)Yang, Zhang, Liu, Freitas, and Lin]{yang2025doestablesourcematter}
B.~Yang, Y.~Zhang, D.~Liu, A.~Freitas, and C.~Lin.
\newblock Does table source matter? benchmarking and improving multimodal scientific table understanding and reasoning, 2025.
\newblock URL \url{https://arxiv.org/abs/2501.13042}.

\bibitem[Yang et~al.(2023)Yang, Hu, Liu, Huang, and Gu]{Yang_2023}
F.~Yang, L.~Hu, X.~Liu, S.~Huang, and Z.~Gu.
\newblock A large-scale dataset for end-to-end table recognition in the wild.
\newblock \emph{Scientific Data}, 10\penalty0 (1), Feb. 2023.
\newblock ISSN 2052-4463.
\newblock \doi{10.1038/s41597-023-01985-8}.
\newblock URL \url{http://dx.doi.org/10.1038/s41597-023-01985-8}.

\bibitem[Yue et~al.(2024)Yue, Ni, Zhang, Zheng, Liu, Zhang, Stevens, Jiang, Ren, Sun, Wei, Yu, Yuan, Sun, Yin, Zheng, Yang, Liu, Huang, Sun, Su, and Chen]{Yue_2024_CVPR}
X.~Yue, Y.~Ni, K.~Zhang, T.~Zheng, R.~Liu, G.~Zhang, S.~Stevens, D.~Jiang, W.~Ren, Y.~Sun, C.~Wei, B.~Yu, R.~Yuan, R.~Sun, M.~Yin, B.~Zheng, Z.~Yang, Y.~Liu, W.~Huang, H.~Sun, Y.~Su, and W.~Chen.
\newblock Mmmu: A massive multi-discipline multimodal understanding and reasoning benchmark for expert agi.
\newblock In \emph{Proceedings of the IEEE/CVF Conference on Computer Vision and Pattern Recognition (CVPR)}, pages 9556--9567, June 2024.

\bibitem[Zhao et~al.(2024{\natexlab{a}})Zhao, Feng, Liu, Tang, Wei, Wu, Liao, Ye, Liu, Li, and Huang]{zhao2024tabpedia}
W.~Zhao, H.~Feng, Q.~Liu, J.~Tang, S.~Wei, B.~Wu, L.~Liao, Y.~Ye, H.~Liu, H.~Li, and C.~Huang.
\newblock Tabpedia: Towards comprehensive visual table understanding with concept synergy, 2024{\natexlab{a}}.

\bibitem[Zhao et~al.(2024{\natexlab{b}})Zhao, Liu, Long, Zhang, Zhao, and Cohan]{zhao2024financemath}
Y.~Zhao, H.~Liu, Y.~Long, R.~Zhang, C.~Zhao, and A.~Cohan.
\newblock Financemath: Knowledge-intensive math reasoning in finance domains, 2024{\natexlab{b}}.
\newblock URL \url{https://arxiv.org/abs/2311.09797}.

\bibitem[Zheng et~al.(2024)Zheng, Feng, Si, She, Lin, Jiang, and Wang]{zheng2024multimodaltableunderstanding}
M.~Zheng, X.~Feng, Q.~Si, Q.~She, Z.~Lin, W.~Jiang, and W.~Wang.
\newblock Multimodal table understanding, 2024.
\newblock URL \url{https://arxiv.org/abs/2406.08100}.

\bibitem[Zheng et~al.(2020)Zheng, Burdick, Popa, Zhong, and Wang]{zheng2020globaltableextractorgte}
X.~Zheng, D.~Burdick, L.~Popa, X.~Zhong, and N.~X.~R. Wang.
\newblock Global table extractor (gte): A framework for joint table identification and cell structure recognition using visual context, 2020.
\newblock URL \url{https://arxiv.org/abs/2005.00589}.

\bibitem[Zhong et~al.(2019)Zhong, Tang, and Yepes]{zhong2019publaynetlargestdatasetdocument}
X.~Zhong, J.~Tang, and A.~J. Yepes.
\newblock Publaynet: largest dataset ever for document layout analysis, 2019.
\newblock URL \url{https://arxiv.org/abs/1908.07836}.

\bibitem[Zhong et~al.(2020)Zhong, ShafieiBavani, and Yepes]{zhong2020imagebasedtablerecognitiondata}
X.~Zhong, E.~ShafieiBavani, and A.~J. Yepes.
\newblock Image-based table recognition: data, model, and evaluation, 2020.
\newblock URL \url{https://arxiv.org/abs/1911.10683}.

\bibitem[Zhu et~al.(2021)Zhu, Lei, Huang, Wang, Zhang, Lv, Feng, and Chua]{zhu-etal-2021-tat}
F.~Zhu, W.~Lei, Y.~Huang, C.~Wang, S.~Zhang, J.~Lv, F.~Feng, and T.-S. Chua.
\newblock {TAT}-{QA}: A question answering benchmark on a hybrid of tabular and textual content in finance.
\newblock In \emph{Proceedings of the 59th Annual Meeting of the Association for Computational Linguistics and the 11th International Joint Conference on Natural Language Processing (Volume 1: Long Papers)}, pages 3277--3287, Online, Aug. 2021. Association for Computational Linguistics.
\newblock \doi{10.18653/v1/2021.acl-long.254}.
\newblock URL \url{https://aclanthology.org/2021.acl-long.254}.

\bibitem[Zhu et~al.(2024)Zhu, Hu, Mo, Li, and Wu]{zhu2024unimedunifiedmedicalgeneralist}
X.~Zhu, Y.~Hu, F.~Mo, M.~Li, and J.~Wu.
\newblock Uni-med: A unified medical generalist foundation model for multi-task learning via connector-moe, 2024.
\newblock URL \url{https://arxiv.org/abs/2409.17508}.

\bibitem[Zhuang et~al.(2024)Zhuang, Huang, Zhang, and Zeng]{zhuang2024mathpumaprogressiveupwardmultimodal}
W.~Zhuang, X.~Huang, X.~Zhang, and J.~Zeng.
\newblock Math-puma: Progressive upward multimodal alignment to enhance mathematical reasoning, 2024.
\newblock URL \url{https://arxiv.org/abs/2408.08640}.

\end{thebibliography}


\clearpage
\appendix
\section*{Appendix}
\label{app_sec}

\section{Design and Analysis of Neuro-Symbolic Routing in TableMoE}
\vskip -0.2in

\begin{table}[htbp]
\centering
\caption{\textbf{Token Role Taxonomy in WMMFinQA (Training Set).} 
We define 9 semantic roles for table tokens to support interpretable and structure-aware expert routing. Example tokens illustrate their usage in real financial tables.}
\label{tab:token_roles}
\resizebox{\linewidth}{!}{
\begin{tabular}{l|p{6.5cm}|p{6.0cm}}
\toprule
\textbf{Role Type} & \textbf{Description} & \textbf{Example Tokens} \\
\midrule
\textbf{HEADER}     & Semantic field title or descriptor               & \texttt{revenue}, \texttt{operating income} \\
\textbf{DATA}       & Core numerical or categorical values             & \texttt{7834}, \texttt{327900}, \texttt{51} \\
\textbf{AXIS}       & Time series or categorical span identifiers      & \texttt{2012}, \texttt{Q4}, \texttt{december 31} \\
\textbf{UNIT}       & Units of measure or financial indicators         & \texttt{\%}, \texttt{\$}, \texttt{in millions} \\
\textbf{TOTAL}      & Summary/aggregated values in row or column       & \texttt{total}, \texttt{sum}, \texttt{total change} \\
\textbf{FORMULA}    & Symbolic/derived expressions                     & $A = P \times \left(1 + \frac{r}{n} \right)^{nt}$, \texttt{LIBOR + 2.05\%}
 \\
\textbf{ANNOTATION} & Clarifying text about a value or cell            & \texttt{( in millions )}, \texttt{( 100 \% )} \\
\textbf{TEXT}       & Natural language or verbose remarks              & \texttt{fidelity national}, \texttt{deferred taxes} \\
\textbf{EMPTY}      & Placeholder or structurally empty cells          & \texttt{""}, \textnormal{—} (em dash), \texttt{--} (double hyphen)\\
\bottomrule
\end{tabular}
}
\vskip -0.2in
\end{table}

\subsection{Token Role Taxonomy and Distribution}
To support interpretable and structured expert routing in TableMoE, we construct a fine-grained taxonomy of semantic token roles. Drawing from linguistic and layout priors in financial tables, we define a set of 9 interpretable role types: \textsc{Header}, \textsc{Data}, \textsc{Axis}, \textsc{Unit}, \textsc{Total}, \textsc{Formula}, \textsc{Annotation}, \textsc{Text}, and \textsc{Empty}. Each role captures a unique functional property of the token, such as whether it denotes a temporal axis, a numeric value, a symbolic computation, or contextual metadata.

Table~\ref{tab:token_roles} provides concise definitions and real-world examples of each role, illustrating their relevance in financial tabular corpora. For instance, tokens like \texttt{LIBOR + 2.05\%} fall under \textsc{Formula}, while \texttt{(in millions)} is categorized as \textsc{Annotation}. Notably, we explicitly model visually ambiguous or structural tokens such as \textnormal{—} and \texttt{--} as \textsc{Empty}, enabling the routing mechanism to bypass them in fusion.

\begin{figure}[htbp]
\centering
\includegraphics[width=\linewidth]{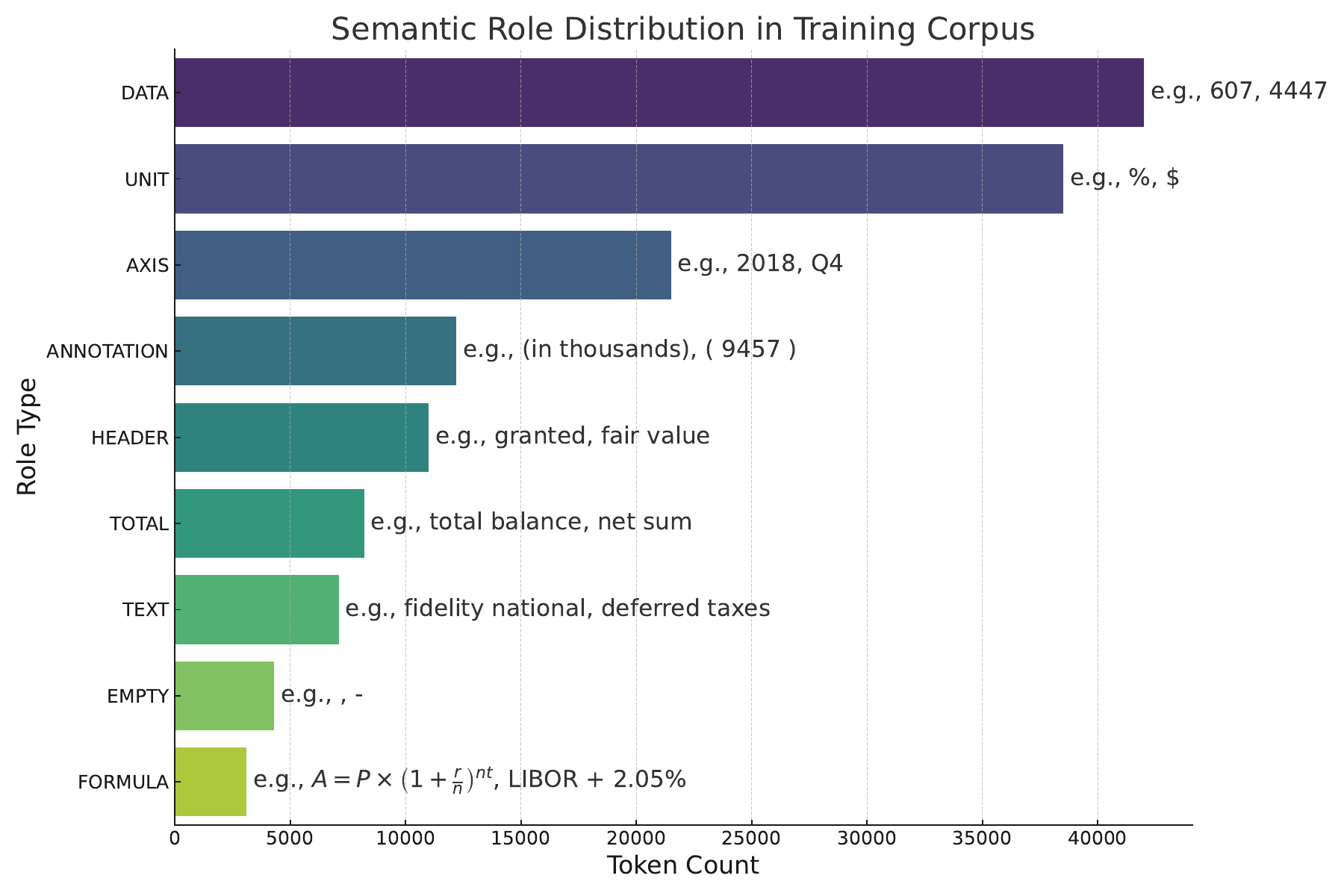}
\vspace{-0.5em}
\caption{
\textbf{Token Role Taxonomy and Distribution in WMMFinQA (Training Set)}.
Our semantic token-role system categorizes tokens into 9 interpretable types (e.g., HEADER, DATA, UNIT, etc.), facilitating role-aware expert routing. The accompanying table visually demonstrates each role in context.
}
\label{fig:token_roles}
\end{figure}

Figure~\ref{fig:token_roles} shows the empirical distribution of token roles in the WMMFinQA training set. We observe a significant skew toward \textsc{Data} and \textsc{Unit}, with rarer but semantically critical roles such as \textsc{Formula} and \textsc{Total} underrepresented. This motivates the use of confidence-aware routing and specialized experts, \textbf {as tokens associated with rare roles often require distinct reasoning strategies}. The imbalance also informs our entropy-suppression strategy, ensuring robustness in routing under token-level uncertainty.


\subsection{Neuro-Symbolic Annealing Objective}

\begin{figure}[t]
  \centering
  \resizebox{0.7\linewidth}{!}{%
    \includegraphics{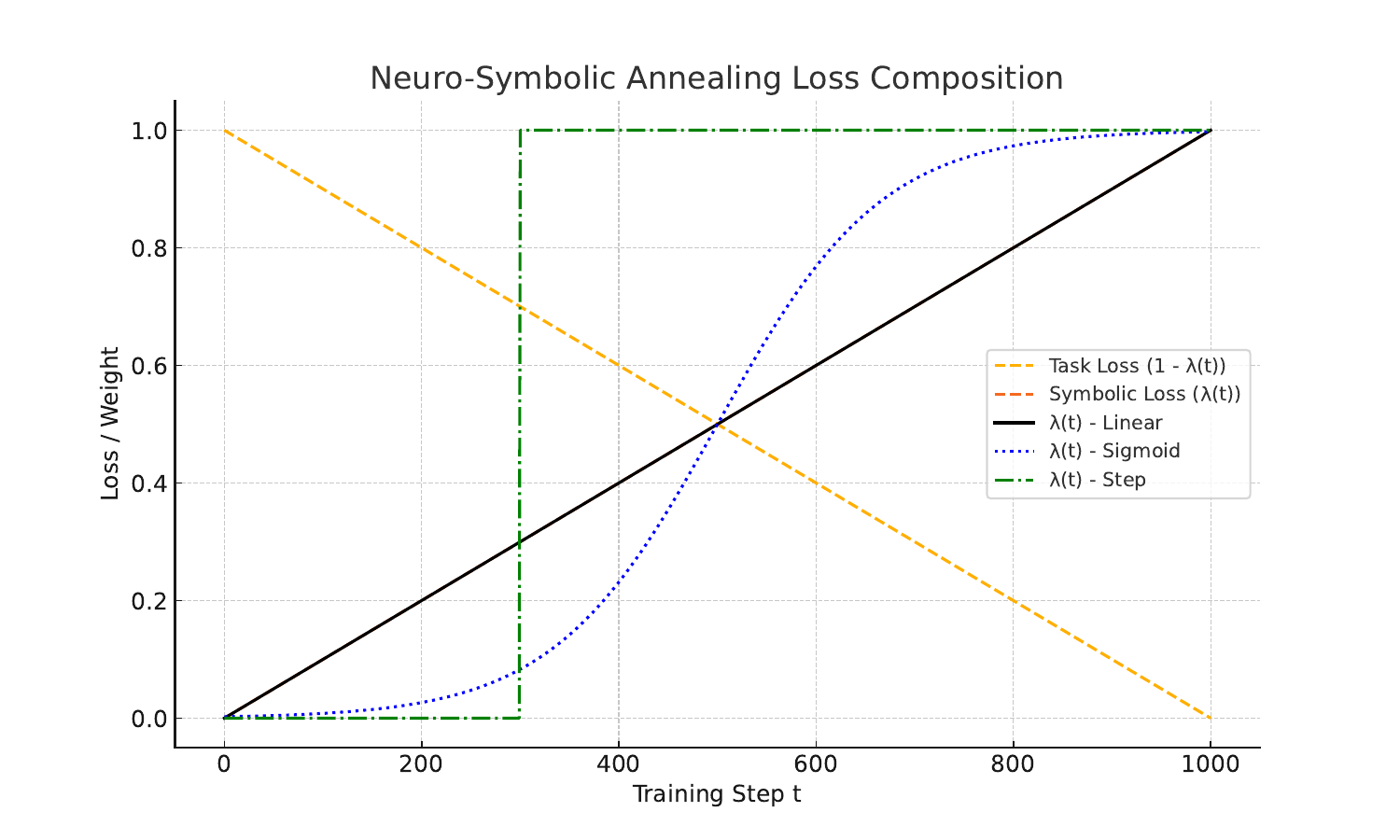}
  }
  \caption{
    Neuro-Symbolic Annealing (NSA) loss scheduling visualization.
    During training, the annealing coefficient $\lambda(t)$ gradually shifts emphasis from the task-specific loss $\mathcal{L}_{\text{task}}$ 
    to symbolic objectives, including role classification $\mathcal{L}_{\text{role}}$ and structural alignment $\mathcal{L}_{\text{struct}}$.
    We illustrate three $\lambda(t)$ schedules: linear, sigmoid, and step-wise.
  }
  \label{fig:nsa_loss_schedule}
\end{figure}

We formulate the training objective of Neuro-Symbolic Annealing (NSA) as a time-dependent loss function that smoothly interpolates between task-driven neural learning and structure-aware symbolic supervision. Let $t$ denote the training epoch (or step), and let $\lambda(t)$ be a monotonic weighting schedule controlling the symbolic influence.

We define the total NSA loss as:
\begin{equation}
\mathcal{L}_{\text{NSA}}(t) = (1 - \lambda(t)) \cdot \mathcal{L}_{\text{task}} + \lambda(t) \cdot \left(\lambda_1 \mathcal{L}_{\text{role}} + \lambda_2 \mathcal{L}_{\text{struct}} \right)
\end{equation}

Here:
\begin{itemize}
\item $\mathcal{L}_{\text{task}}$ denotes the main supervision signal, e.g., QA loss or decoder loss;
\item $\mathcal{L}_{\text{role}}$ supervises token role classification (e.g., \texttt{HEADER}, \texttt{AXIS}, etc.);
\item $\mathcal{L}_{\text{struct}}$ supervises symbolic structure compatibility (e.g., layout graph alignment);
\item $\lambda_1$, $\lambda_2$ are static hyperparameters balancing symbolic subcomponents;
\item $\lambda(t)$ is the symbolic annealing schedule, defined as:
\end{itemize}
\begin{equation}
\lambda(t) = \frac{1}{1 + \exp\left(\frac{e - t}{s}\right)}
\end{equation}

Finally, we present the condensed formulation:
\begin{equation}
\mathcal{L}_{\text{NSA}}(t) = (1 - \lambda(t)) \cdot \mathcal{L}_{\text{neural}} + \lambda(t) \cdot \mathcal{L}_{\text{symbolic}}
\end{equation}
\begin{equation}
\mathcal{L}_{\text{symbolic}} = \lambda_1 \mathcal{L}_{\text{role}} + \lambda_2 \mathcal{L}_{\text{struct}}
\end{equation}

\section{More Implementation Details}


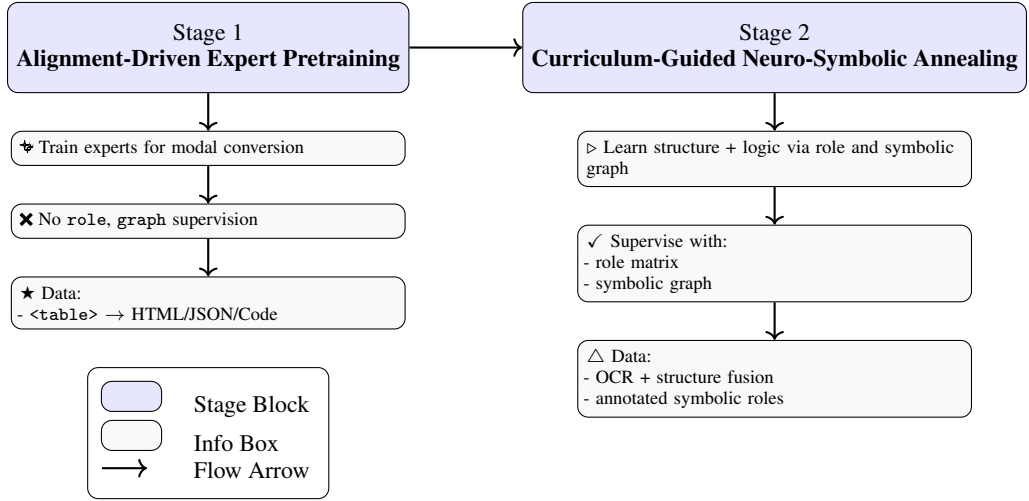
\begin{figure}[ht]
\centering
\begin{tikzpicture}[
    stagebox/.style={draw, rounded corners, fill=blue!10, minimum width=4.8cm, minimum height=1.2cm, font=\small, align=center}, 
    info/.style={draw, rounded corners, fill=gray!5, font=\scriptsize, text width=5cm, align=left}, 
    legend/.style={rectangle, draw=black, rounded corners, inner sep=5pt, font=\scriptsize, fill=white}, 
    arrow/.style={->, thick},
    node distance=0.5cm and 1.5cm
]

\node[stagebox] (s1) {Stage 1\\\textbf{Alignment-Driven Expert Pretraining}}; 
\node[info, below=of s1] (s1a) {✨ Train experts for modal conversion};
\node[info, below=of s1a] (s1b) {❌ No \texttt{role}, \texttt{graph} supervision};


\node[info, below=of s1b] (s1c) {\ding{72} Data:\\- \texttt{<table>} $\rightarrow$ HTML/JSON/Code};

\node[stagebox, right=of s1] (s2) {Stage 2\\\textbf{Curriculum-Guided Neuro-Symbolic Annealing}};
\node[info, below=of s2] (s2a) {$\triangleright$ Learn structure + logic via role and symbolic graph};
\node[info, below=of s2a] (s2b) {$\checkmark$ Supervise with:\\- role matrix\\- symbolic graph};
\node[info, below=of s2b] (s2c) {$\triangle$ Data:\\- OCR + structure fusion\\- annotated symbolic roles};

\draw[arrow] (s1) -- (s2);
\draw[arrow] (s1) -- (s1a); 
\draw[arrow] (s1a) -- (s1b);
\draw[arrow] (s1b) -- (s1c);

\draw[arrow] (s2) -- (s2a);
\draw[arrow] (s2a) -- (s2b);
\draw[arrow] (s2b) -- (s2c);

\node[legend, below=0.5cm of s1c, font=\footnotesize, align=left] (legend) {
  \begin{tabular}{@{}ll@{}}
    \tikz{\node[stagebox, minimum width=0.8cm, minimum height=0.4cm] {};} & Stage Block \\
    \tikz{\node[draw, rounded corners, fill=gray!5, minimum width=0.8cm, minimum height=0.4cm] {};} & Info Box \\
    \tikz{\draw[->, thick] (0,0) -- +(0.6cm, 0);} & Flow Arrow \\
  \end{tabular}
};

\end{tikzpicture}
\caption{Two-stage Neuro-Symbolic Expert Training Flow. Stage 1 focuses on modal alignment; Stage 2 incorporates symbolic reasoning via role and graph supervision.}
\label{fig:training_flow}
\end{figure}

\subsection{Baselines}
The version of GPT-4o used in all experiments is gpt-4o-2024-08-06, and the version of GPT-4 Turbo with Vision used in all experiments is gpt-4-turbo-2024-04-09.

\subsection{Experimental Settings}
We detail the two-stage training pipeline of TableMoE, which is designed to align modality-specialized experts and refine symbolic routing under structural noise. In Stage 1, we conduct expert alignment pretraining using 1.2M table quadruples across general, HTML, JSON, and code formats to endow each expert with modality-specific structural priors. In Stage 2, we apply \textit{Neuro-Symbolic Annealing}, a fine-tuning strategy that jointly optimizes task loss, token-role supervision, and structural consistency using real-world datasets (TabRecSet \citep{Yang_2023}, WMMFinQA, WMMTatQA). This stage leverages role-aware routing, confidence suppression, and symbolic weight annealing to robustly fuse expert outputs under visual and semantic uncertainty. Table~\ref{tab:training_settings} summarizes the training hyperparameters and key design choices for both stages.

\subsection{Curriculum Training Design}

To extend the generalization of structure-aware symbolic routing to low-resource languages and real-world tables in the wild, we augment both the expert alignment and curriculum-guided annealing stages with two bilingual, visually diverse corpora—\textbf{WTW} \citep{long2021parsingtablestructureswild} and \textbf{TabRecSet} \citep{Yang_2023}. This enables TableMoE to learn robust multimodal alignment across both linguistic and structural regimes, particularly in the early stages of curriculum training and token-role–supervised Neuro-Symbolic Annealing. Notably, this enhancement does not modify our annealing protocol but enriches symbolic grounding and significantly improves generalization to real-world, Chinese-centric table reasoning tasks such as \textbf{WMMTabDialog}.



\begin{table}[htbp]
\centering
\small
\caption{Training Hyperparameters for TableMoE Expert Alignment Pretraining and Neuro-Symbolic Annealing}
\label{tab:training_settings}
\resizebox{\linewidth}{!}{
\begin{tabular}{l|c|c}
\toprule
\textbf{Parameter} & \textbf{Expert Pretraining (Align)} & \textbf{Neuro-Symbolic Annealing} \\
\midrule
Dataset Size & 1.2M Table Quadruples & 68K Table QA Samples (incl. CoT \& PoT) \\
Modality Types & General, HTML, JSON, Code & All + Token Role + Graph \\
Optimizer & AdamW ($\beta_1{=}0.9$, $\beta_2{=}0.95$, $\epsilon{=}1\mathrm{e}^{-8}$) & AdamW ($\beta_1{=}0.9$, $\beta_2{=}0.95$, $\epsilon{=}1\mathrm{e}^{-8}$) \\
Learning Rate & 5e-5 & 2e-5 \\
LR Schedule & cosine decay & cosine decay\\
Batch Size & 128 & 64 \\
Gradient Accumulation & 4 & 4 \\
Warmup Ratio & 0.1 & 0.03 \\
Numerical Precision & bfloat16& bfloat16\\
Epoch & N/A (step-based) & 3 \\
Checkpointing & Every 2K Steps & Every 1K Steps \\
Weight Decay & 0.05& 0.05\\
Backbone & LLaMA-3.1 / Qwen2.5-VL & LLaMA-3.1 / Qwen2.5-VL \\
Max Input Length & 8192 & 8192 \\
LoRA Usage & No & Yes (r=64, $\alpha$=64, dropout=0.05) \\
Deepspeed Stage& 2 & 3 \\
Expert Training & Expert-Specific Modal Alignment & Symbolic-Aware Joint Training \\
Routing Strategy & Softmax Token Matching & Annealed Symbolic-Neural Fusion \\
Loss Terms & Modal Alignment Loss & $\mathcal{L}_{\text{NSA}} = \lambda(t) \cdot \mathcal{L}_{\text{neural}} + (1 - \lambda(t)) \cdot \mathcal{L}_{\text{symbolic}}$ \\
GPU Hours (8 $\times$ A100-80G) & 30 & 10\\
\bottomrule
\end{tabular}}
\end{table}

\section{More Experimental Results}

\begin{figure}[t]
    \centering
    \includegraphics[width=0.85\linewidth]{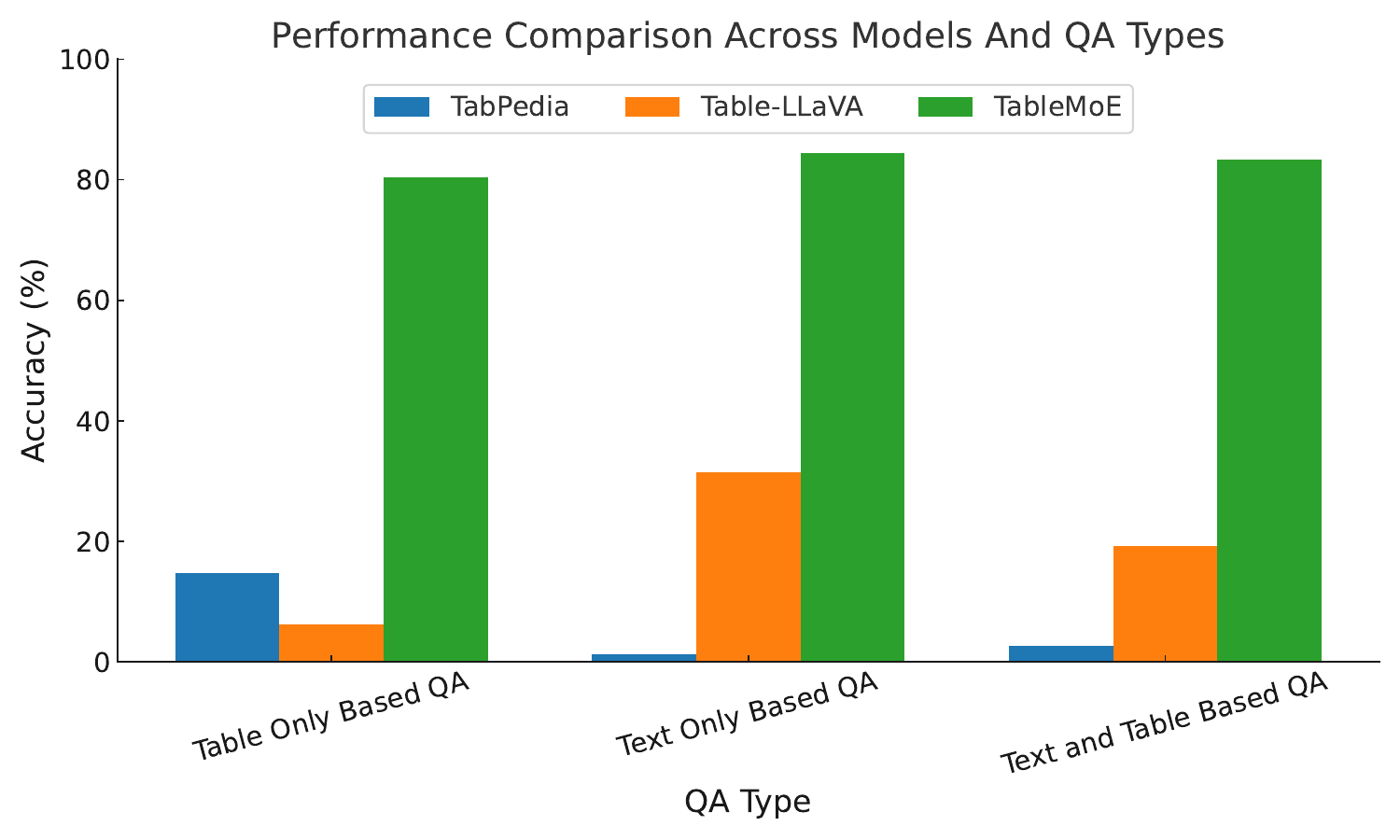}
    \caption{
        \textbf{Model Performance on WMMFinQA under \textit{WildStruct} Perturbations.}
        We compare \textbf{TabPedia}, \textbf{Table-LLaVA}, and \textbf{TableMoE} across Table-, Text-, and Table+Text-based QA. 
        WMMFinQA features five challenging \textit{WildStruct} styles: incomplete headers or fonts, skewed layouts, blurred contents, watermarked noise, and compositional degradation.
        While baseline models degrade sharply, \textbf{TableMoE} remains robust across all settings.
    }
    \label{fig:performance_comparison_wildstruct_wmmfinqa}
\end{figure}

\subsection{Benchmarking Domain-Specific Baselines on Real-World Table-Centric Reasoning Tasks}


FinanceMath test set is a knowledge-intensive QA benchmark designed for evaluating both single-table and multi-table reasoning in image-grounded contexts \cite{zhao2024financemath}. As shown in Table~\ref{tab:financemath_wmmfinancemath_rsn_com}, reasoning over table images imposes substantially greater challenges for multimodal large language models compared to purely textual QA, further highlighting the inherent complexity of structured visual grounding. Consequently, both Table-LLaVA \citep{zheng2024multimodaltableunderstanding} and TabPedia \citep{zhao2024tabpedia} perform suboptimally, likely due to the architectural rigidity of their monolithic designs (see Section~\ref{subsec:mmtur}) and their reliance on single-table training paradigms. Interestingly, the 13B variant of Table-LLaVA underperforms its 7B counterpart, suggesting that simply scaling model size without structural adaptation fails to enhance, and may even impair, performance on downstream table-based reasoning tasks.

\textbf{WildStruct Degradation Analysis.} The WMMFinanceMath benchmark isolates the effects of \textit{WildStruct}, a family of real-world structural and visual degradations encompassing incomplete header fonts, skewed layouts, low-resolution blur, and watermark occlusions. When evaluated on these degraded yet realistic table images, both models experience a substantial drop in performance: Table-LLaVA (7B) suffers a 20.97\% relative decrease in accuracy, while TabPedia drops by 16.78\%. Notably, these degradations exclusively affect the \textit{w/ Table Image} scores,  indicating a pronounced fragility in their visual-table encoders. This performance collapse underscores the necessity of a structurally-aware vision-language connector capable of robustly grounding semantics in noisy and layout-complex table images.

Moreover, the WMMFinQA benchmark targets \textit{WildStruct} degradations, realistic structural and visual perturbations including incomplete headers or fonts, skewed layouts, blurred content, watermarked noise, and compositional degradation.
As illustrated in Figure~\ref{fig:performance_comparison_wildstruct_wmmfinqa}, these distortions severely impair the performance of TabPedia and Table-LLaVA, with significant drops across table-only and cross-modal QA settings.
In contrast, TableMoE maintains high accuracy under all conditions, revealing strong robustness to layout and visual noise.
This divergence highlights the necessity of structurally-aware multimodal connectors that can faithfully align vision and semantics amid noisy, misaligned, or compositionally complex table images.

These phenomena reveal a critical limitation: domain-specific multimodal models like Table-LLaVA and TabPedia are particularly sensitive to noisy, structure-rich input distributions. Despite being fine-tuned on high-quality table corpora, their performance collapses when confronted with the structural entropy of in-the-wild financial tables, calling into question their robustness in real-world deployment scenarios.

\textbf{Implications.} These findings underscore the necessity of structurally adaptive routing and symbolic inductive bias in modeling complex tables. Unlike general image-text tasks, table-centric QA demands models that can reason over discrete layout semantics and maintain interpretability under structural noise. The TableMoE framework, as discussed in Section~\ref{tab_moe}, addresses this gap through role-guided expert allocation and entropy-aware annealing, achieving superior robustness across WildStruct variants.

\begin{table}[htbp]
\centering
\caption{Performance comparison on FinanceMath and WMMFinanceMath benchmarks.}
\label{tab:financemath_wmmfinancemath_rsn_com}
\small
\resizebox{\textwidth}{!}{
\begin{tabular}{l|ccc|ccc|c}
\toprule
\multirow{2}{*}{Models} &\multicolumn{3}{c|}{FinanceMath (Relax Acc @0.005)} &\multicolumn{3}{c|}{WMMFinanceMath (Relax Acc @0.005)} & \multirow{2}{*}{WildStruct $\bigtriangledown$} \\
\cmidrule(lr){2-4} \cmidrule(lr){5-7}
& w/ Table Image & w/o Table Image & Avg & w/ Table Image & w/o Table Image & Avg\\
\midrule
TabPedia (7B)& \cellcolor{blue!10}0.0447 & 0.0553 & 0.0510 & \cellcolor{red!10} 0.0372 & 0.0553 & 0.0480 & -16.78\% (-0.0075)\\
Table-LLaVA (7B) & \cellcolor{blue!10} 0.0596 & 0.1173 & 0.0940 &\cellcolor{red!10}0.0471&0.1173 &0.0890& -20.97\% (-0.0125)\\
Table-LLaVA (13B)& \cellcolor{blue!10} 0.0317 & 0.0466 & 0.0406 &\cellcolor{red!10}0.0246 &0.0466& 0.0377& -22.40\% (-0.0071)\\
\bottomrule
\end{tabular}
}
\end{table}

\section{Further Analysis}
\label{app:ana}

\subsection{Research Questions and Answers}
To provide a deep understanding of TableMoE's design, generalization, and performance, we summarize our findings through the following refined research questions (RQs):

\textbf{RQ1:} \textit{What are the benefits of Neuro-Symbolic Annealing (NSA) over conventional fine-tuning?}\\
\textbf{Answer:} NSA introduces semantic routing guided by token roles and expert compatibility, enabling symbolic interpretability and specialization. It significantly improves robustness in degraded scenarios, outperforming traditional neural-only fine-tuning by 3.7 accuracy points on WMMFinQA.

\textbf{RQ2:} \textit{Is token role classification introduced during expert alignment pretraining or during NSA? Why?}\\
\textbf{Answer:} Token role prediction is integrated into NSA. While MLP expert alignment focuses on modality-specific structure reconstruction (e.g., HTML, JSON), NSA introduces structural semantics via token roles to supervise routing and enhance symbolic grounding.

\textbf{RQ3:} \textit{How does token role supervision help TableMoE handle visual degradation and incomplete structure?}\\
\textbf{Answer:} Role supervision enables semantic routing that is robust to missing fonts, partial borders, and layout skew. Table~\ref{tab:wmmtab_inf_rsn_02} shows a good example for this case. It guides expert selection even when low-level visual cues are unreliable, improving robustness on WildStruct-Bench under real-world degradations.

\textbf{RQ4:} \textit{Can TableMoE generalize beyond financial tables to open-domain table and chart reasoning tasks?} \\
\textbf{Answer:} Yes. TableMoE is designed as a domain-agnostic architecture and is pretrained on the large-scale \textbf{TableMoE-Align} dataset, which comprises 1.2M high-quality table–HTML–JSON–code quadruples spanning finance, science, biomedicine, and industrial domains. This alignment-driven pretraining equips the model with rich structural and symbolic priors. As shown in \Cref{tab:chartqa_financemath_reasoning_comparison}, TableMoE demonstrates strong generalization to out-of-domain benchmarks such as MMMU and ChartQA, outperforming baseline models and showcasing robust adaptability across diverse real-world tabular and visual reasoning scenarios.

\textbf{RQ5:} \textit{How does the WildStruct dataset’s finance focus affect generalization?}\\
\textbf{Answer:} 
While \textit{WildStruct} is constructed from financial documents to systematically benchmark visual-structural robustness, TableMoE is not trained solely on financial data. Our Neuro-Symbolic Annealing corpus also incorporates the TabRecSet (ZH/EN) dataset, which spans a wide range of real-world domains, including education, administrative forms and web pages. This heterogeneous supervision enables the model to acquire symbolic reasoning capabilities that are structurally grounded but semantically agnostic. As a result, TableMoE exhibits strong generalization even to out-of-domain tables, confirming the transferability of its neuro-symbolic inductive biases. Future expansions of \textit{WildStruct} to healthcare and scientific domains will further validate domain transfer under complex symbolic layouts.

\textbf{RQ6:} \textit{What are the limitations of using synthetic alignment data in pretraining?}\\
\textbf{Answer:} Although the alignment data used in pretraining originate from real-world tables, it lacks typical visual noise and linguistic diversity, which limits generalization. To address this, we incorporate the WTW dataset \citep{long2021parsingtablestructureswild} during alignment, as it provides structurally degraded tables (e.g., blur, occlusion, layout loss) along with bilingual coverage (Chinese and English). This integration enhances TableMoE’s robustness to visual perturbations and cross-lingual variation, bridging the gap between synthetic alignment and real-world complexity.

\textbf{RQ7:} \textit{How do biases introduced by GPT-generated PoT and RoT steps affect TableMoE’s performance, and what strategies can reduce these biases?}\\
\textbf{Answer:} GPT-generated programs may reflect stylistic bias or spurious shortcuts. We enforce program diversity via beam sampling, adversarial filtering, and prompt rephrasing, improving consistency and generalization of symbolic reasoning.

\textbf{RQ8:} \textit{How does TableMoE scale with different backbone LLM sizes?}\\
\textbf{Answer:} We ablate the size of the backbone using LLaMA-2-7B and LLaMA-2-13B, and find that neuro-symbolic routing consistently delivers a +2.9 point improvement in WMMFinQA. This indicates that TableMoE’s structural inductive bias enhances reasoning robustness independently of model scale, offering complementary gains to pure capacity increases.

\textbf{RQ9:} \textit{Are the improvements of TableMoE valid under fair fine-tuning settings compared to zero-shot and fine-tuned baselines?}\\
\textbf{Answer:} 
Yes. To ensure fairness, we fine-tune strong vision-language baselines (e.g., Qwen2.5-VL-7B) on the same WildStruct corpus. Even under matched fine-tuning protocols, TableMoE consistently outperforms these models, confirming that the observed improvements stem from its symbolic routing and role-aware MoE pretraining—not from discrepancies in training regimes. This validates the robustness and generality of our approach under controlled, comparable settings.

\textbf{RQ10:} \textit{How does TableMoE compare to domain-specific baselines for table understanding?}\\
\textbf{Answer:} TableMoE significantly outperforms prior multimodal table understanding models like TabPedia (7B) and Table-LLaVA (7B), which directly encode visual tables through large vision-language models. Despite their ability to process table images, these models remain brittle under structural degradation, scoring only 6.44\% and 12.89\% average accuracy on \textit{WildStruct}-Bench (see Table~\ref{tab:wildstruct}). In contrast, TableMoE achieves 80.31\% with symbolic Program-of-Thought prompting, narrowing the gap with GPT-4o and human performance. This gain stems from its modular expert design and neuro-symbolic routing, which confer superior robustness in noisy, multi-format, and domain-shifted tables—surpassing both monolithic and modality-agnostic baselines.

\textbf{RQ11:} \textit{What is the contribution of Program-of-Thought (PoT) reasoning to TableMoE?}\\
\textbf{Answer:} PoT bridges the expert outputs with symbolic program generation. It improves numerical and logical reasoning, raising WMMFinQA scores from 78.73 to 81.98 and enabling interpretable multi-step inference in structurally noisy tables.

\textbf{RQ12:} \textit{What emergent capabilities does TableMoE exhibit?}\\
\textbf{Answer:} TableMoE exhibits compelling emergent behaviors beyond supervised objectives. It reliably suppresses hallucinations under visual degradation by deferring uncertain predictions—enabled by entropy-aware neuro-symbolic routing (see Table~\ref{tab:WMMTatQA_inf_rsn_03}). 
Despite no explicit supervision, TableMoE generalizes zero-shot to new domains for downstream tasks (e.g., psychology, materials; see Table~\ref{tab:mmmu}) and resolves cross-lingual field alignment in mixed Chinese-English tables. These behaviors stem from the synergy of symbolic role grounding and modular expert specialization, indicating strong out-of-distribution robustness and reasoning plasticity.

\subsection{Robustness Under Incomplete Table Scenarios}
To assess the reliability of state-of-the-art multimodal large language models (MLLMs) under real-world interface imperfections, we conduct a comparative case study using cropped (incomplete) and full (complete) table screenshots (Table~\ref{tab:incom_com_tab_rsn_study}). Although both versions contain sufficient information to answer tax- and depreciation-related queries, we observe a stark divergence in model performance.

When reasoning over incomplete tables, GPT-4 Turbo and GPT-4o fail to infer the correct \textit{original value excluding tax} and \textit{monthly depreciation}, despite the tax rate and total price being present. The absence of the auxiliary cue "Price with Tax Included" appears to disrupt their mental model of the layout, leading to invalid base assumptions. Given the complete image, however, both models recover correct results, confirming their reliance on semantic layout anchors.

This phenomenon illustrates a broader vulnerability: MLLMs are sensitive to layout incompleteness—even if required values exist visually, missing context hinders structure disambiguation and reasoning alignment. These breakdowns are especially common in mobile screenshots and web forms.

TableMoE addresses this by leveraging symbolic token roles to recover structural semantics. Through roles like \textsc{Price}, \textsc{Tax Credit}, and \textsc{Deduction}, our routing layer retains expert reasoning capabilities even under visual truncation, providing resilience in incomplete real-world conditions.

\subsection{Qualitative Failures and Hallucinations in WildStruct Tables}
To further evaluate reasoning robustness, we conduct detailed qualitative analyses spanning Table~\ref{tab:wmmtab_rsn_chat_01} to Table~\ref{tab:WMMTatQA_inf_rsn_04}.

Notably, Table~\ref{tab:wmmtab_checkbox_rsn_01} demonstrates a failure case where GPT-4 Turbo and GPT-4o hallucinate incorrect answers, such as fabricated names and totals, when facing structurally complex checkbox-style tables. These hallucinations often arise from misinterpreting layout cues or overfitting to familiar patterns in pretraining.

Such results highlight the unique difficulty posed by \textit{WildStruct} scenarios: irregular, nested, or partially structured table images elicit brittle behaviors from MLLMs. In contrast, TableMoE’s symbolic routing and semantic structure modeling allow it to remain faithful to visual semantics and avoid factual drift. This confirms that token-level structure induction and symbolic supervision are vital for grounded multimodal reasoning in the wild.

\section{Further Study}

\subsection{TableMoE Built on Other MLLMs}

\begin{table}[htbp]
\centering
\caption{Performance comparison on ChartQA and FinanceMath benchmarks.}
\label{tab:chartqa_financemath_reasoning_comparison}
\small
\resizebox{\linewidth}{!}{
\begin{tabular}{l|ccc|ccc}
\toprule
\multirow{2}{*}{Models} & \multicolumn{3}{c|}{ChartQA (Relax Acc @0.005)} & \multicolumn{3}{c}{FinanceMath (Relax Acc @0.005)} \\
\cmidrule(lr){2-4} \cmidrule(lr){5-7}
                        & Human & Aug & Avg & w/ Table Image & w/o Table Image & Avg \\
\midrule
Qwen2.5-VL-7B                & 0.6177 & 0.1264 & 0.2444 & 0.0893 & 0.3802 & 0.2630 \\
Qwen2.5-VL-7B + TableMoE     & 0.6980 & 0.2018 & 0.3204 & 0.1723 & 0.5120 & 0.3751 \\
\bottomrule
\end{tabular}
}
\end{table}
To assess the generality and architectural adaptability of TableMoE, we instantiate it as a modular reasoning layer atop two distinct vision-language backbones: LLaMA3.1–8B with ViT and Qwen2.5–VL–7B–Instruct. In this experiment, we apply TableMoE to Qwen2.5-VL-7B using only 10\% of the alignment data (see \Cref{tab:tablemoe_align_data}), together with the training sets of ChartQA \citep{masry2022chartqabenchmarkquestionanswering} and WMMFinQA \citep{chen-etal-2021-finqa}. As a controlled baseline, we fine-tune the original Qwen2.5-VL-7B on the same fine-tuning data without TableMoE architecture.

We assess both models on two clean, plain-image benchmarks: (1) ChartQA \citep{masry2022chartqabenchmarkquestionanswering}, a well-established dataset for evaluating chart reasoning, and (2) FinanceMath \citep{zhao2024financemath}, where we convert its markdown-formatted tables into canonical table images—free from \textit{WildStruct} degradation. These benchmarks allow us to verify whether TableMoE maintains its reasoning effectiveness in structured, in-distribution settings.

As shown in \Cref{tab:chartqa_financemath_reasoning_comparison}, TableMoE consistently improves over the backbone model. On ChartQA, the gains stem from the specialized reasoning capabilities introduced by our alignment-pretrained experts (HTML, JSON, and Code). On FinanceMath, TableMoE demonstrates its ability to generalize symbolic reasoning even in domains not characterized by structural noise. These results affirm that TableMoE provides a modular and portable reasoning scaffold, improving MLLM performance even under clean table-image distributions.

\begin{table}[htbp]
\centering
\small
\caption{\textbf{Statistics of synthetic Table-to-X alignment data used for pretraining TableMoE.} Sample counts are based on original dataset availability with adjustments for alignment conversions. A nearly balanced subset of 1.2M samples (600K HTML, 400K JSON, 200K Code) is selected for pretraining across structural modalities.}
\label{tab:tablemoe_align_data}
\begin{tabular}{l|c|l|c|c|c}
\toprule
\textbf{Source} & \textbf{Data Type} & \textbf{Task Type} & \textbf{Samples} & \textbf{WildStruct} & \textbf{Language} \\
\midrule
\multicolumn{6}{c}{\textit{Alignment Training (per task type)}} \\
\midrule
FinTabNet \citep{zheng2020globaltableextractorgte} & synthetic & table to HTML & 113K & \usym{2717} & English \\
& synthetic & table to JSON & 113K & \usym{2717} & English \\
& synthetic & table to code & 113K & \usym{2717} & English \\
\midrule
PubTabNet \citep{zhong2020imagebasedtablerecognitiondata} & synthetic & table to HTML & 568K & \usym{2717} & English \\
& synthetic & table to JSON & 568K & \usym{2717} & English \\
& synthetic & table to code & 568K & \usym{2717} & English \\
\midrule
TableBank \citep{li2020tablebankbenchmarkdatasettable} & synthetic & table to HTML & 145K & \usym{2717} & English \\
& synthetic & table to JSON & 145K & \usym{2717} & English \\
& synthetic & table to code & 145K & \usym{2717} & English \\
\midrule
WTW \citep{long2021parsingtablestructureswild} & synthetic & table to HTML & 14.5K & \usym{2713} & Chinese+English \\
& synthetic & table to JSON & 14.5K & \usym{2713} & Chinese+English \\
& synthetic & table to code & 14.5K & \usym{2713} & Chinese+English \\
\midrule
\multicolumn{4}{l|}{\textbf{Total}} & \multicolumn{2}{c}{\textbf{2.5M}} \\
\multicolumn{4}{l|}{\textbf{Usage:}  HTML = \textbf{600K} JSON = \textbf{400K} Code = \textbf{200K}} & \multicolumn{2}{c}{\textbf{1.2M}} \\
\bottomrule
\end{tabular}
\end{table}

\section{Extended Dataset and Benchmark Design}
\label{data_construct}

To support both training and evaluation of TableMoE, we introduce two key resources: the \textbf{TableMoE-Align} corpus for expert modality alignment pretraining and the \textbf{\textit{WildStruct}} suite for fine-tuning and benchmark evaluation.

\subsection{TableMoE-Align Pretraining Corpus}

As summarized in Table~\ref{tab:tablemoe_align_data}, we construct \textbf{TableMoE-Align} by aggregating raw tables from FinTabNet \citep{zheng2020globaltableextractorgte}, PubTabNet \citep{zhong2020imagebasedtablerecognitiondata}, TableBank \citep{li2020tablebankbenchmarkdatasettable}, and WTW \citep{long2021parsingtablestructureswild}. Using GPT-4o \citep{openai2024gpt-4o}, we generate 2.5M high-fidelity table--HTML--JSON--code quadruples. From this, we select a nearly balanced 1.2M-sample subset (600K HTML, 400K JSON, 200K Code) to pretrain alignment across structured modalities.




\subsection{\textit{WildStruct} and Public Evaluation Benchmarks}
\label{subsec:w&p_bench}
\begin{figure*}[htbp]
\centering
\vskip -0.3in
\hspace{-190mm}
\begin{minipage}{0.93\textwidth}
\centering
\includegraphics[width=\linewidth]{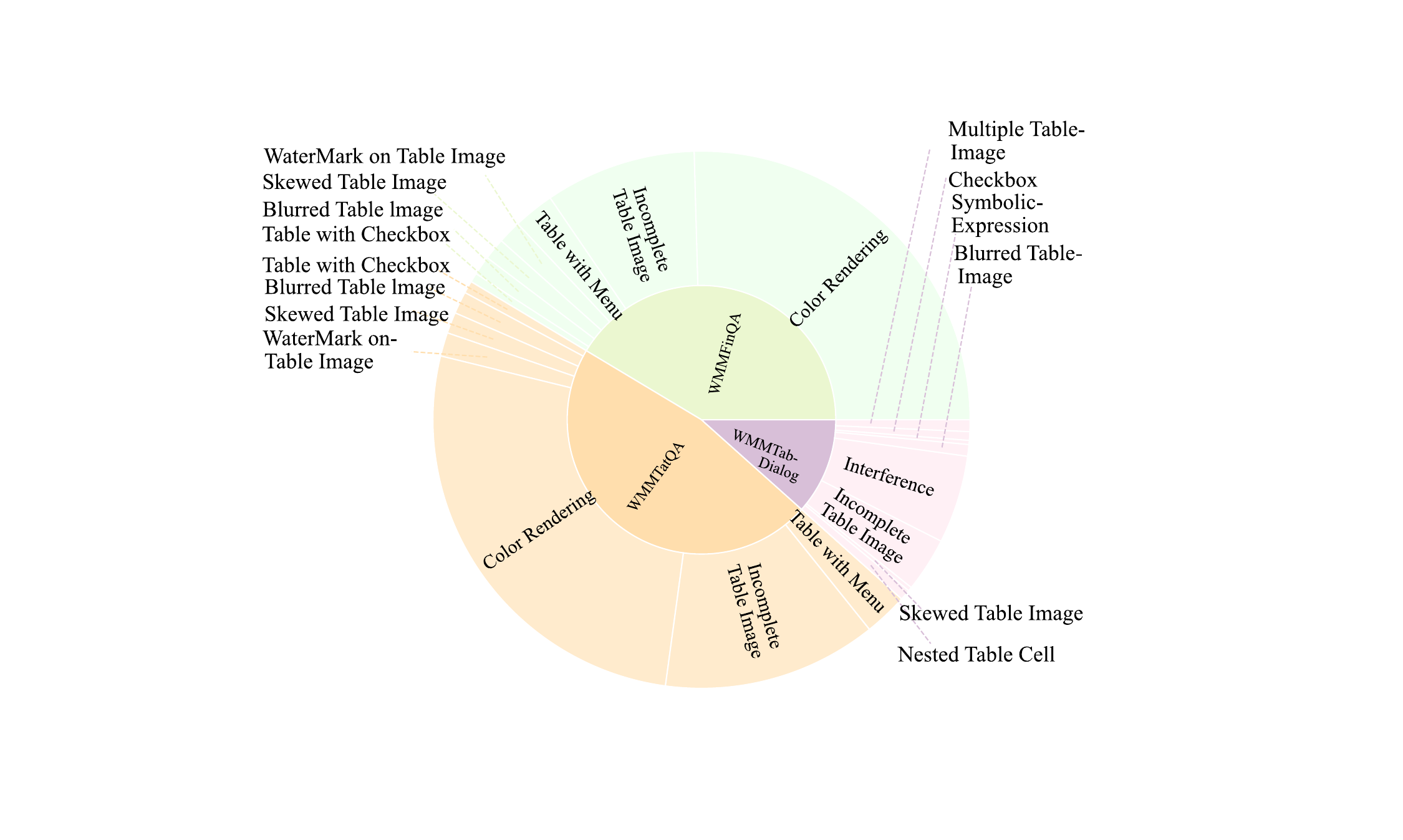}
\vskip -0.4in
\captionof{subfigure}{Table Image Type}
\label{fig:sub1}
\end{minipage}
\hspace{-73mm}
\begin{minipage}{0.93\textwidth}
\centering
\includegraphics[width=\linewidth]{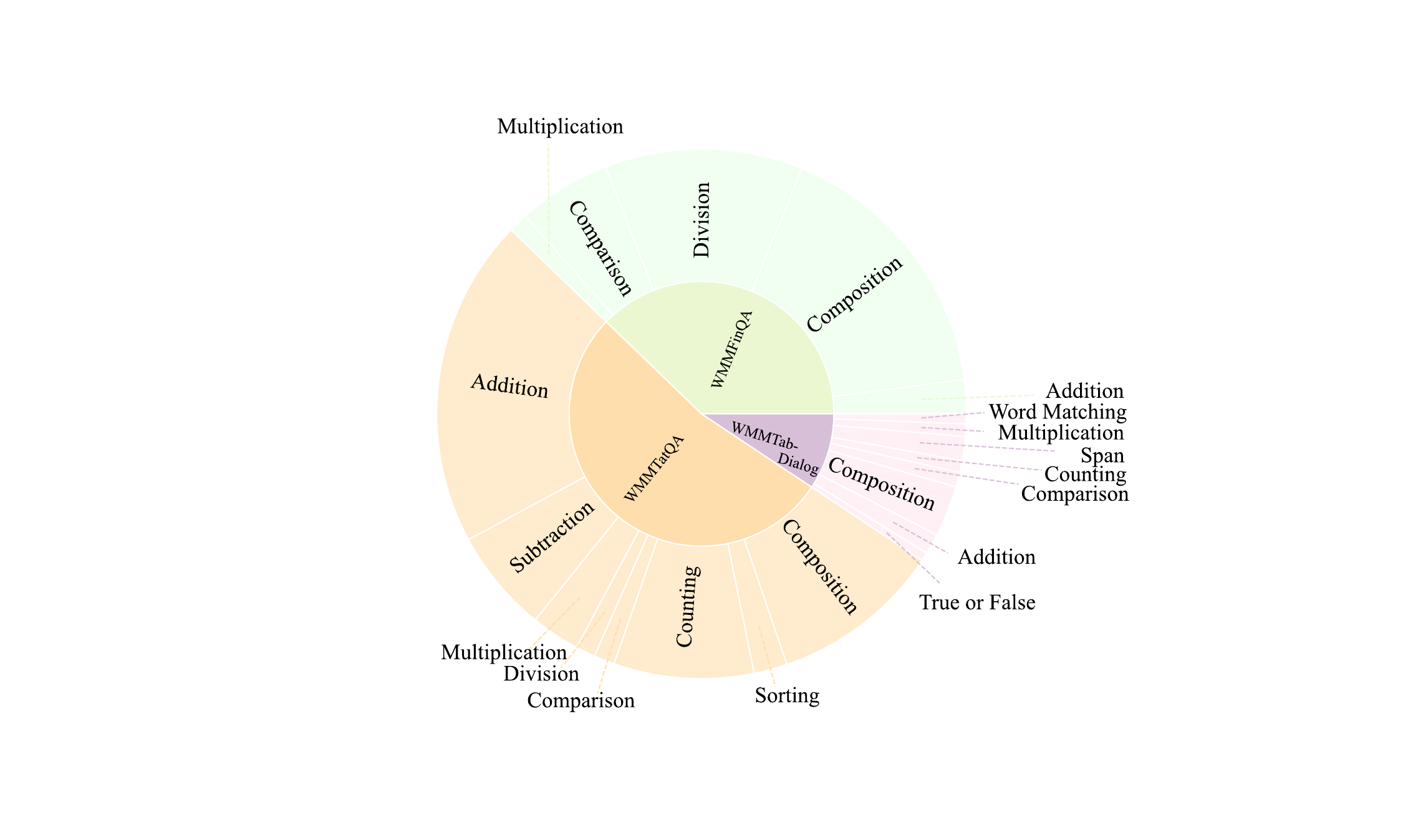}
\vskip -0.4in
\captionof{subfigure}{Numerical Reasoning Type}
\label{fig:sub2}
\end{minipage}
\hspace{-190mm}
\caption{Fine-grained Taxonomy of Table Image and Numerical Reasoning in \textit{WildStruct} Benchmarks}
\label{fig:fine_grained_img_nr_type}
\end{figure*}

To assess model generalization in realistic and structurally complex scenarios, we introduce the \textbf{\textit{WildStruct}} benchmark suite, covering both held-in and held-out evaluation settings. This suite targets table-centric reasoning under diverse task types and modalities, with a focus on financial mathematics, multi-turn dialogue, and multilingual content. Our held-in benchmarks, \textbf{WMMFinQA} and \textbf{WMMTatQA}, support model adaptation through supervised fine-tuning on in-domain financial QA and hybrid text-table reasoning tasks. In contrast, \textbf{WMMTabDialog} and \textbf{WMMFinanceMATH} serve as held-out and out-of-domain testbeds, emphasizing multi-turn reasoning, symbolic robustness, and knowledge-intensive queries in both English and Chinese. To evaluate cross-domain generalization, we incorporate two public benchmarks: \textbf{FinanceMATH} \citep{zhao2024financemath}, with tables rendered from markdown into plain images, and the table split of \textbf{MMMU} \citep{Yue_2024_CVPR}, covering diverse table reasoning tasks across domains. Table~\ref{tab:benchmarks} provides a comparative summary of all benchmark corpora, including their structural modality, task focus, language coverage, and scale.

\begin{table}[t]
  \caption{\textbf{Evaluation Benchmarks Overview.} We categorize benchmarks by provenance (ours vs. public), role in evaluation (held-in vs. held-out), and task characteristics.}
  \label{tab:benchmarks}
  \centering
  \begin{threeparttable}
  \resizebox{\linewidth}{!}{
  \begin{tabular}{l c c p{3.6cm} >{\centering\arraybackslash}p{2.7cm} c c}
    \toprule
    \textbf{Corpus} & \textbf{WildStruct} & \textbf{Generalization Settings} & \textbf{Task Type} & \textbf{Multimodal Table Reasoning} & \textbf{Lang.} & \textbf{\# Samples} \\
    \midrule
    WMMFinQA & \usym{2713} & Held-In  & Table-Centric QA for Financial Mathematics & \usym{2713} & EN & 1147 \\
    WMMTatQA & \usym{2713} & Held-In  & Text+Table Numerical QA & \usym{2713} & EN & 1663 \\
    WMMTabDialog & \usym{2713} & Held-Out or Out-of-Domain & Dialog-Based Table QA (Multi-Turn) & \usym{2713} & ZH & 231 \\
    WMMFinanceMATH\tnote{1} & \usym{2713} & Held-Out or Out-of-Domain& Knowledge-Intensive Financial and Healthcare Math QA & \usym{2713} & EN & 1000 \\
    \midrule
    FinanceMATH \citep{zhao2024financemath}\tnote{2} & \usym{2717} & Held-Out or Out-of-Domain & Knowledge-Intensive Financial and Healthcare Math QA & \usym{2717} & EN & 1000 \\
    MMMU \citep {Yue_2024_CVPR} (Table) & \usym{2717} & Held-Out or Out-of-Domain & Multi-Domain Table-Centric Reasoning & \usym{2713} & EN & 2463 \\
    \bottomrule
  \end{tabular}
  }
  \begin{tablenotes}[flushleft]
    \footnotesize
    \item[1] We convert markdown-format tables of FinanceMATH \citep{zhao2024financemath} into \textit{WildStruct}-style table images with synthetic \\ \setlength{\parindent}{2em}  noise, preserving the original data content.
    \item[2] We render original markdown tables into plain table images and formulate multimodal queries for evaluation.
  \end{tablenotes}
  \end{threeparttable}
\end{table}

\begin{figure*}[htbp]
        \small
	\centering
	\vspace{-0.15in}
	\begin{minipage}{1\linewidth}	
		\subfigure[incomplete table image with some column information missing]{
			\label{fig:1}
			\includegraphics[width=0.47\linewidth,height=1.2in]{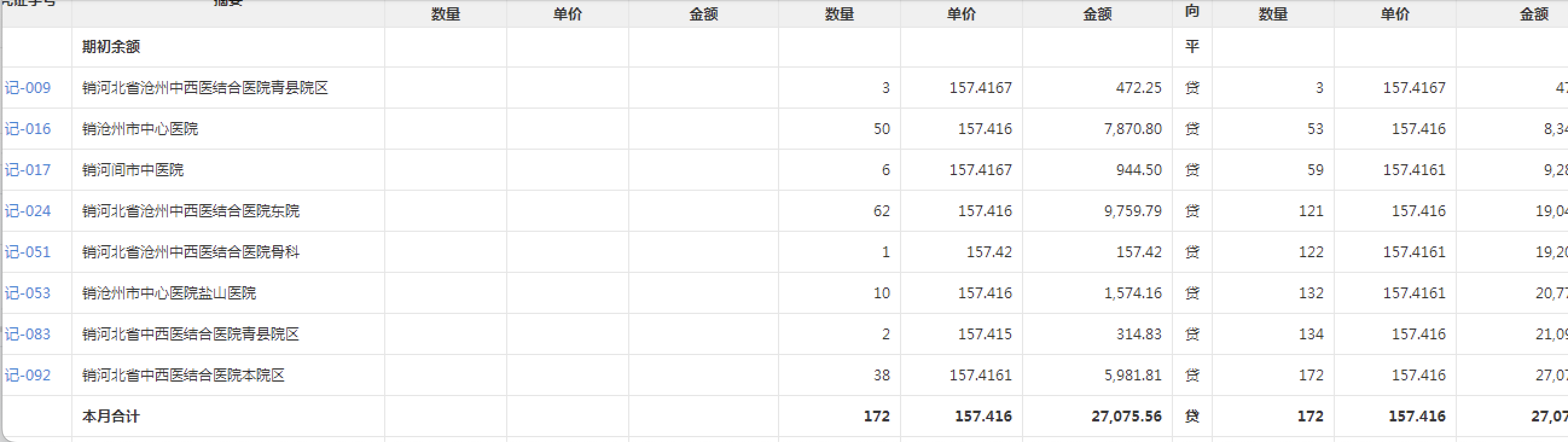}	
		}\noindent
            \hspace{.15in}
		\subfigure[incomplete table image with row header and  column header missing]{
			\label{fig:2}
			\includegraphics[width=0.47\linewidth,height=1.2in]{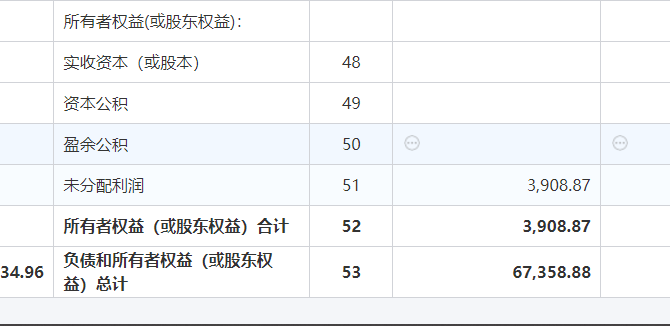}
		}
	\end{minipage}
	\vskip -0.3cm 
	\begin{minipage}{1\linewidth }
            \subfigure[table image with nested table cell]{
			\label{fig:3}
			\includegraphics[width=0.47\linewidth,height=1.2in]{{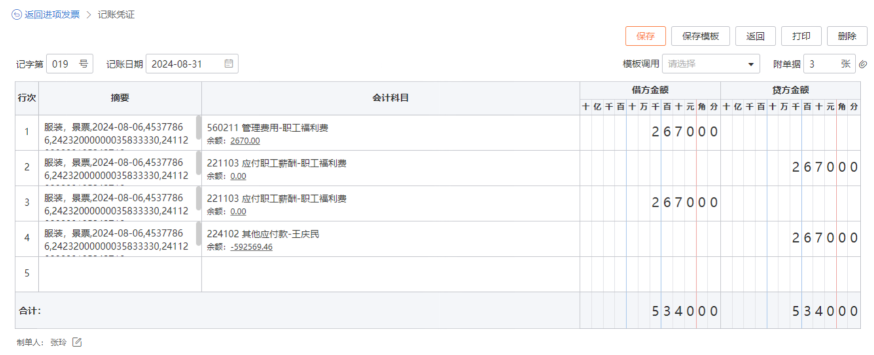}
			}
		}\noindent
            \hspace{.15in}
		\subfigure[table image with unwanted noise]{
			\label{fig:4}
			\includegraphics[width=0.47\linewidth,height=1.2in]{{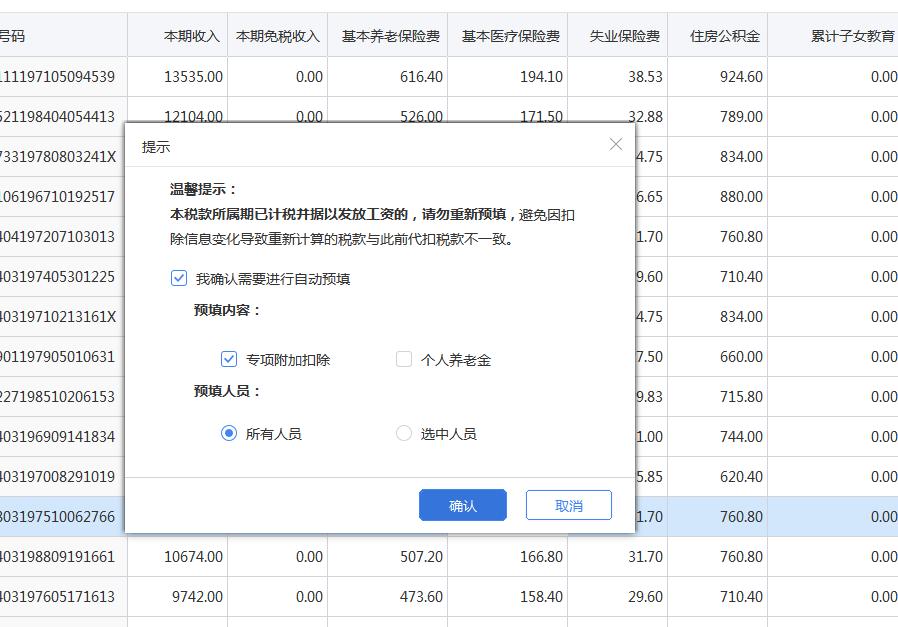}
			}
		}
	\end{minipage}
 \vspace{-0.15in}
	\begin{minipage}{1\linewidth}	
		\subfigure[table image with menus inside and outside table cells]{
			\label{fig:5}
			\includegraphics[width=0.47\linewidth,height=1.2in]{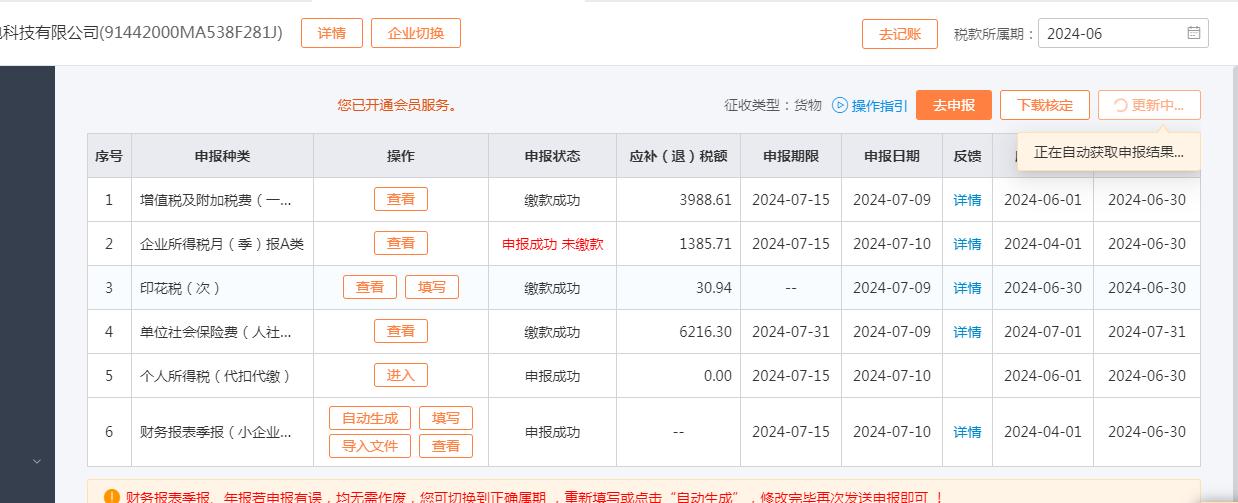}	
		}\noindent
            \hspace{.15in}
		\subfigure[Table image with watermark]{
			\label{fig:6}
			\includegraphics[width=0.47\linewidth,height=1.2in]{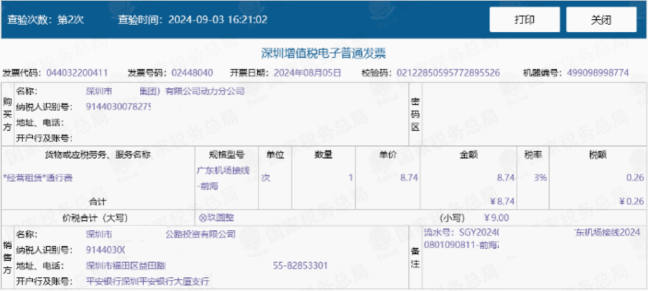}
		}
	\end{minipage}
	\vskip 0.1cm 
	\begin{minipage}{1\linewidth }
		\subfigure[table image with checkbox in table cell]{
			\label{fig:7}
			\includegraphics[width=0.47\linewidth,height=1.2in]{{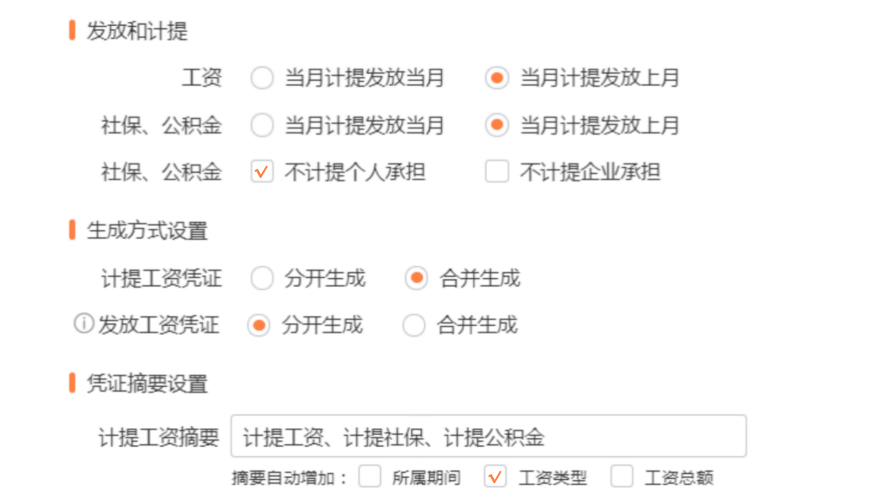}
			}
		}\noindent
            \hspace{.15in}
		\subfigure[skewed table image]{
			\label{fig:8}
			\includegraphics[width=0.47\linewidth,height=1.2in]{{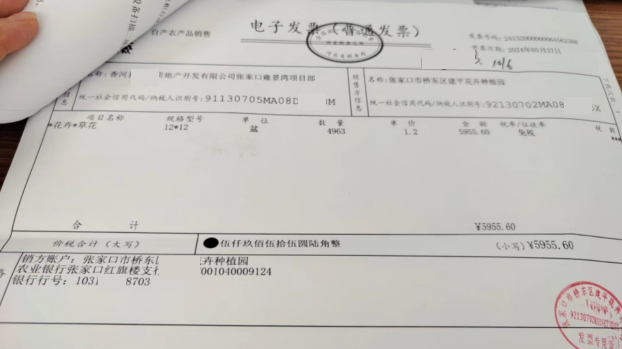}
			}
		}
	\end{minipage}
 \vspace{-0.15in}
	\begin{minipage}{1\linewidth}	
		\subfigure[table image with symbolic expression in table cell (as shown by the blue underline)]{
			\label{fig:9}
			\includegraphics[width=0.47\linewidth,height=1.2in]{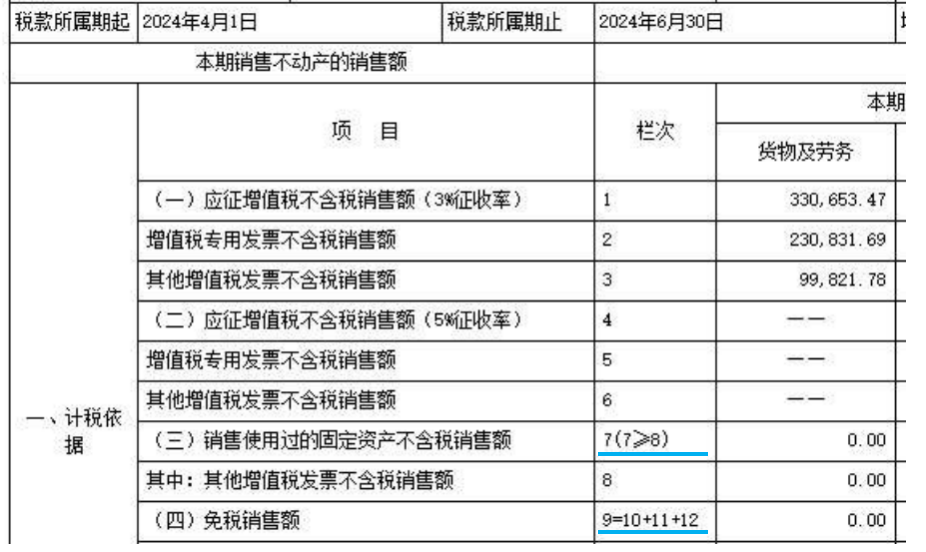}	
		}\noindent
            \hspace{.15in}
		\subfigure[table image with irrelevant background]{
			\label{fig:10}
			\includegraphics[width=0.47\linewidth,height=1.2in]{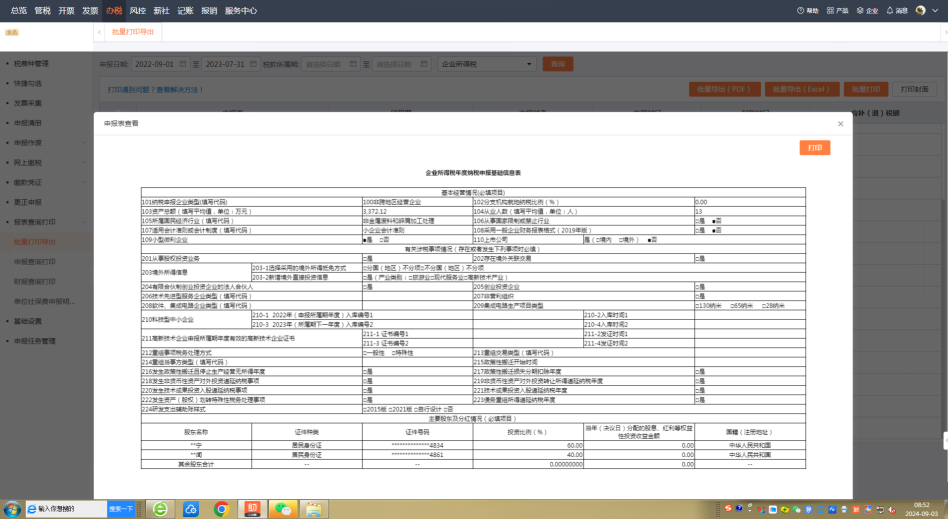}
		}
	\end{minipage}
	\vskip +0.2cm 
	\begin{minipage}{1\linewidth }
		\subfigure[table image with checkbox outside table]{
			\label{fig:11}
			\includegraphics[width=0.47\linewidth,height=1.2in]{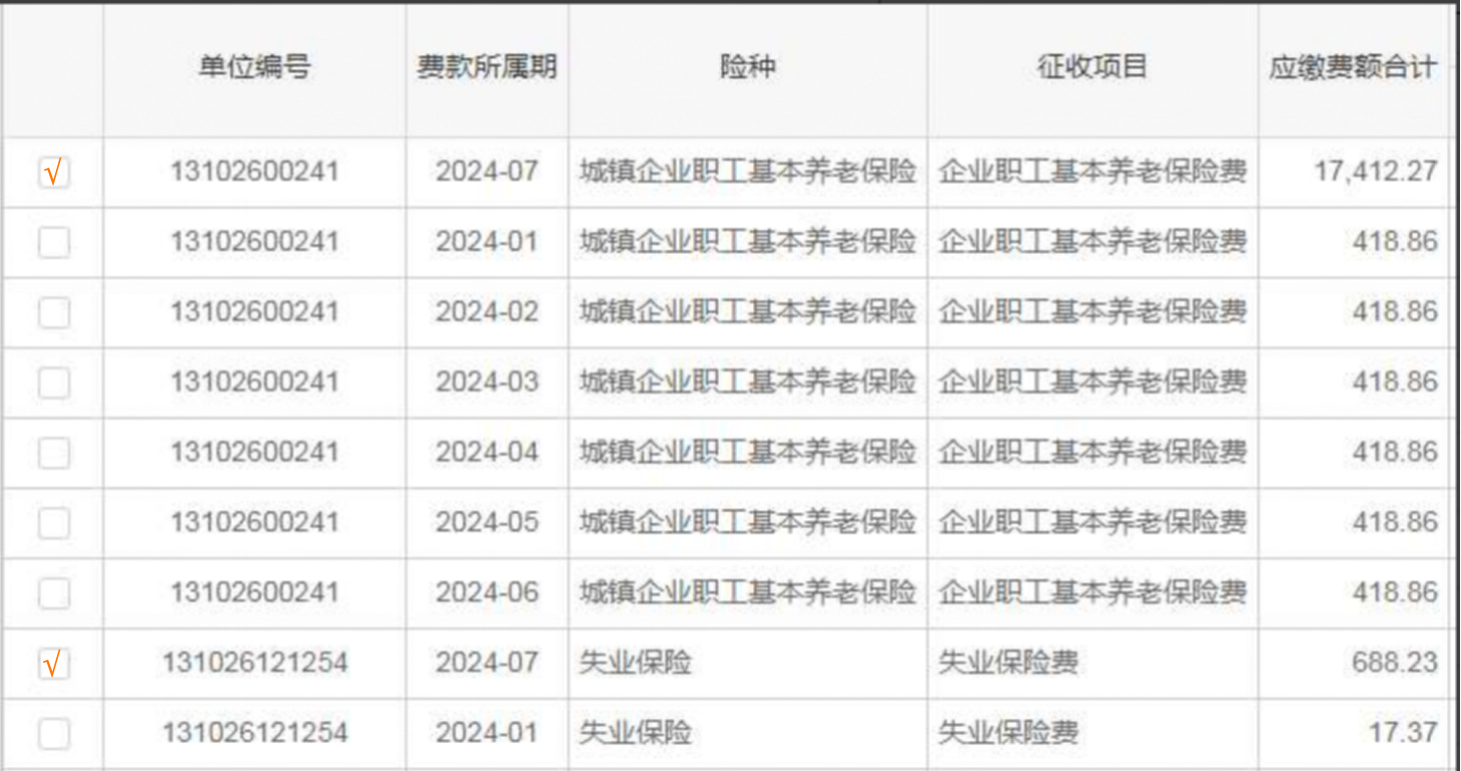}
		}\noindent
            \hspace{.15in}
		\subfigure[blurred table image]{
			\label{fig:12}
			\includegraphics[width=0.47\linewidth,height=1.2in]{{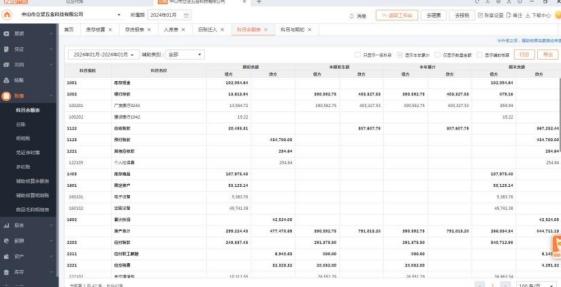}
			}
		}
	\end{minipage}

	\vspace{-0.05in}	
	\caption{\textit{WildStrcut} images demonstration of WMMTabDialog-bench.}
	\vspace{-0.2in}		
	\label{fig:WMMTabDialog_bench}
\end{figure*}

\paragraph{WMMTabDialog Benchmark}
To assess structured multimodal reasoning in practical scenarios, we introduce \textbf{WMMTabDialog}—a held-out benchmark featuring Chinese financial-tax dialogues grounded in complex, real-world table images. These samples reflect enterprise-level workflows where users consult in-the-wild tables for tax-related questions. Each sample is authored and verified by three certified tax professionals following protocols from FinQA \citep{chen-etal-2021-finqa} and TAT-QA \citep{zhu-etal-2021-tat}, ensuring coverage of diverse numerical reasoning patterns (see Figure~\ref{fig:fine_grained_img_nr_type}) and realistic \textit{WildStruct} layouts (see Figure~\ref{fig:WMMTabDialog_bench}). Only samples that meet strict criteria for domain specificity, logical soundness, completeness and answerability are retained. Unlike crowdsourced corpora, WMMTabDialog is curated through a domain-expert-only pipeline, which is essential for high-quality. \textbf{Held out from all training, WMMTabDialog serves as a zero-shot benchmark} to evaluate generalization under multilingual, structurally degraded, and semantically complex table conditions.

\paragraph{WMMFinQA and WMMTatQA Benchmarks}
We employ the similar approach, illustrated in \Cref{para:wmm_finqa_tatqa_cura}, to construct WMMFinQA and WMMTatQA Benchmarks.

\begin{table}[htbp]
\centering
\caption{\textbf{Training Datasets for Neuro-Symbolic Annealing in TableMoE.} Diverse symbolic-rich corpora supervise expert routing.}
\label{tab:nsa_training_data}
\resizebox{\linewidth}{!}{
\begin{tabular}{l|c|c|p{5.5cm}|p{4.5cm}}
\toprule
\textbf{Dataset} & \textbf{Samples} & \textbf{Language} & \textbf{Task Type} & \textbf{Remarks} \\
\midrule
WMMFinQA & 6,251 & English & Table-centric Mathematical QA & High-quality synthetic training data with \textit{WildStruct} format and symbolic program annotation \\
WMMTatQA & 2,201 (13,215 turns) & English & Multi-turn Table-centric Mathematical QA & Conversational numerical reasoning; structurally degraded \textit{WildStruct} format \\
TabRecSet (ZH) \citep{Yang_2023} & 30,000 & Chinese & TSR + TCR + TQA & Covers multilingual, scanned, and distorted formats; samples may contain both tasks \\
TabRecSet (EN) \citep{Yang_2023} & 30,000 & English & TSR + TCR + TQA & OCR-derived complex layout tables with bounding box and structure annotation \\
\bottomrule
\end{tabular}}
\end{table}

\begin{figure}[htbp]
\centering
\includegraphics[width=1.0\textwidth]{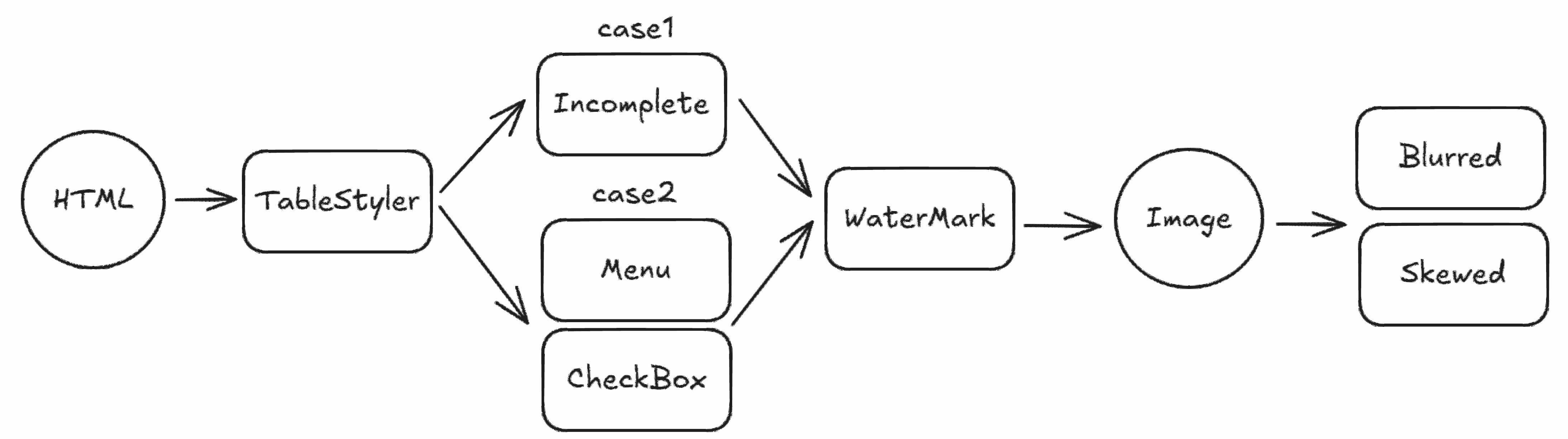}
\caption{Table Image Rendering Pipeline for WMMFinQA and WMMTatQA.}
\label{fig:tab_rendering}
\end{figure}

\subsection{Neuro-Symbolic Annealing Training Corpus} \label{subsec:nsa_corpus}

\paragraph{WMMFinQA and WMMTatQA Fine-tuning Dataset} \label{para:wmm_finqa_tatqa_cura}
We transform FinQA \citep{chen-etal-2021-finqa} and TAT-QA \citep{zhu-etal-2021-tat} into \textbf{WMMFinQA} and \textbf{WMMTatQA} by rendering incomplete, degraded or noisy table images and aligning CoT and PoT traces. Generation steps include CSS-based HTML rendering, image post-processing, GPT-4-assisted rationale synthesis, human verification, role annotation, etc.

\begin{enumerate}
\item Designing CSS templates for realistic table rendering (e.g., zebra stripes, menus, checkboxes).
\item Generating incomplete and distorted images via coordinate-aware HTML cropping, watermark injection, and synthetic noise (e.g., skew, blur), as illustrated in Figure~\ref{fig:tab_rendering}.
\item Converting HTML tables into rendered images using WebDriver and BeautifulSoup.
\item Prompting GPT-4 to generate \textit{Program-of-Thought (PoT)} and \textit{Multimodal Table Chain-of-Thought (MMTab-CoT)} rationales using curated templates, as shown in \Cref{tab:pot} and \Cref{tab:wmm-cot_gen} .
\item Applying self-consistency filtering \citep{wang2023selfconsistencyimproveschainthought} and expert validation to ensure program fidelity.
\item Combining rendered images, CoT/PoT, domain knowledge, and original QA pairs to synthesize training instances for Neuro-Symbolic Annealing.
\end{enumerate}

This procedure yields rich, visually noisy QA data with verified symbolic supervision, supporting robust and interpretable table reasoning.

\paragraph{TabRecSet Fine-Tuning Dataset}\label{para:tabrecset_cura}
To enable symbolic supervision across structure, content, and reasoning, we construct a diverse fine-tuning corpus (Table~\ref{tab:nsa_training_data}). In addition to WMMFinQA and WMMTatQA, we leverage TabRecSet \citep{Yang_2023}—the largest bi-lingual (EN/ZH) in-the-wild benchmark for table recognition, featuring 38.1K real-world tables with rich polygon-based annotations for table structure recognition (TSR) and table content recognition (TCR). To extend its utility to reasoning, we generate table question answering (TQA) annotations using GPT-4o \citep{openai2024gpt-4o}, followed by expert validation. Following rigorous quality control, we select 30K high-quality samples per language, yielding a total of 68,452 fine-tuning instances across TSR, TCR, and TQA tasks.

\section{Limitations and Broader Impacts.}\label{app:limit} Despite its modular and efficient MoCE design, TableMoE faces several limitations. First, its three connector experts are pre-trained independently, which may limit early synergy and cross-task transfer among alignment objectives. Second, reliance on synthetic alignment data (HTML/JSON/Code) may weaken generalization to highly unstructured, noisy, or handwritten tables often encountered in real-world \textit{WildStruct} scenarios. Third, our role prediction mechanism operates with fixed role taxonomies and lacks uncertainty calibration beyond entropy-based suppression, which may lead to suboptimal routing under domain shifts or low-resource table types. Addressing these challenges will require joint expert pretraining, real-world weak supervision, and adaptive symbolic abstractions in future work.
\newline TableMoE aims to improve the robustness and interpretability of multimodal table understanding, which may benefit applications in scientific analysis, financial auditing, and biomedical report parsing. However, enhanced table processing capabilities could also be misused in surveillance, automated misinformation extraction, or high-stakes decision systems lacking human oversight. Our model does not incorporate sensitive demographic attributes, but care should be taken when applying it to domains involving privacy or fairness concerns. Future work should explore safeguards for alignment in real-world deployments.

\begin{table*}[htbp]
\small
\centering
\caption{GPT-4 prompt template example that used to generate MMTab-CoT for WMMFinQA train dataset.} 
\label{tab:wmm-cot_gen}
\begin{tcolorbox}[colframe=black,colback=white]
\textcolor{blue}{System Prompt}:\newline
You are an expert in finance and taxation, skilled in addressing complex, multi-step reasoning questions within these fields. You are capable of providing precise answers to a broad range of intricate inquiries, utilizing both domain knowledge and table textual data.
\newline
\newline \textcolor{blue}{Instruction Prompt}:
\newline Given you a question, relevant text knowledge, a table in text list format and the corresponding answer, please deeply perceive and understand these information and use them to generate intermediate reasoning steps that derive the answer to the question. 
\newline
\newline You can use mathematical formulas to calculate intermediate steps and final results. Output each reasoning step on a separate line and do not add any number before each reasoning step. 
\newline
\newline If the result contains a percentage, round to the nearest single digit. 
\newline
\newline If the question is a "True" or "False" problem, or a "Yes" or "No" problem, ensure that one of them is included in the final result.
\newline
\newline Output the final result in JSON format: $\left\{\textrm{"The answer is": \textrm{$\textrm{$\left\{\textrm{YOUR ANSWER}\right\}$}$}}\right\}$ 
\newline
\newline Always take care to describe each step of reasoning as briefly as possible.
\newline
\newline Here are some demonstrations with similar questions, and intermediate reasoning steps to arrive at the answer in these demonstrations may assist you in answering the question:
\newline
\newline \textcolor{blue}{Few-shot Demonstration}:
\newline \dots
\end{tcolorbox}
\end{table*}

\begin{table*}[htbp]
  \small
  \centering
  \caption{GPT-4 prompt template used to generate PoT for the WMMFinQA train dataset.}
  \label{tab:pot}

  \begin{adjustbox}{max width=\textwidth}
  \begin{tcolorbox}[colframe=black,colback=white]
\textcolor{blue}{\textbf{System Prompt}}:\\
You are a financial expert; write a Python program that answers the given question.
The \emph{return} value of the program must be the final answer.
\medskip
\newline \textcolor{blue}{\textbf{Example program}}:\\
\begin{lstlisting}[language=Python]
def solution():
    # Define variables
    ...
    # Do math calculation
    ...
    return answer
\end{lstlisting}

\textcolor{blue}{\textbf{Instruction Prompt}}:\\
Given a question, relevant text knowledge, a table (list format), calculation steps 
(list format), a formula program, and an answer, derive the intermediate reasoning
steps, i.e.\ the Program-of-Thought, for the question.
\newline Here are demonstrations with similar questions whose intermediate steps may
help you.  
\newline Output \emph{only} the part:

\begin{lstlisting}[language=Python]
def solution():
    ...
\end{lstlisting}
  \end{tcolorbox}
  \end{adjustbox}
\end{table*}

\begin{figure}[htbp]
\centering
\subfigure[incomplete]{
\includegraphics[width=\mysize]{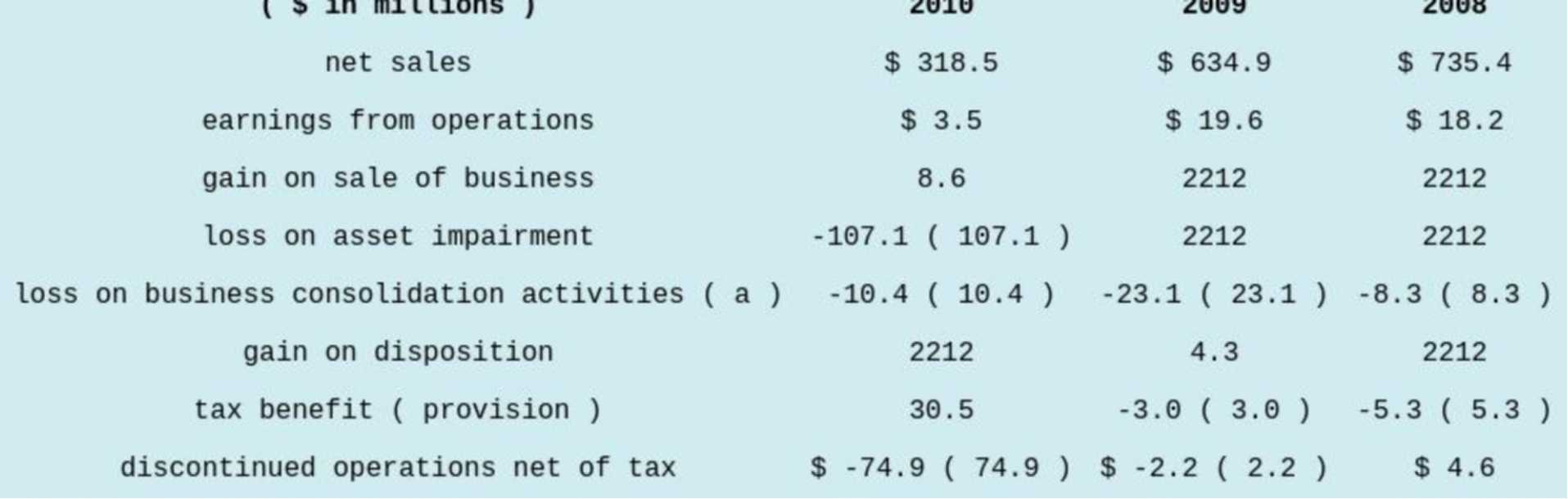}
}
\quad
\subfigure[incomplete]{
\includegraphics[width=\mysize]{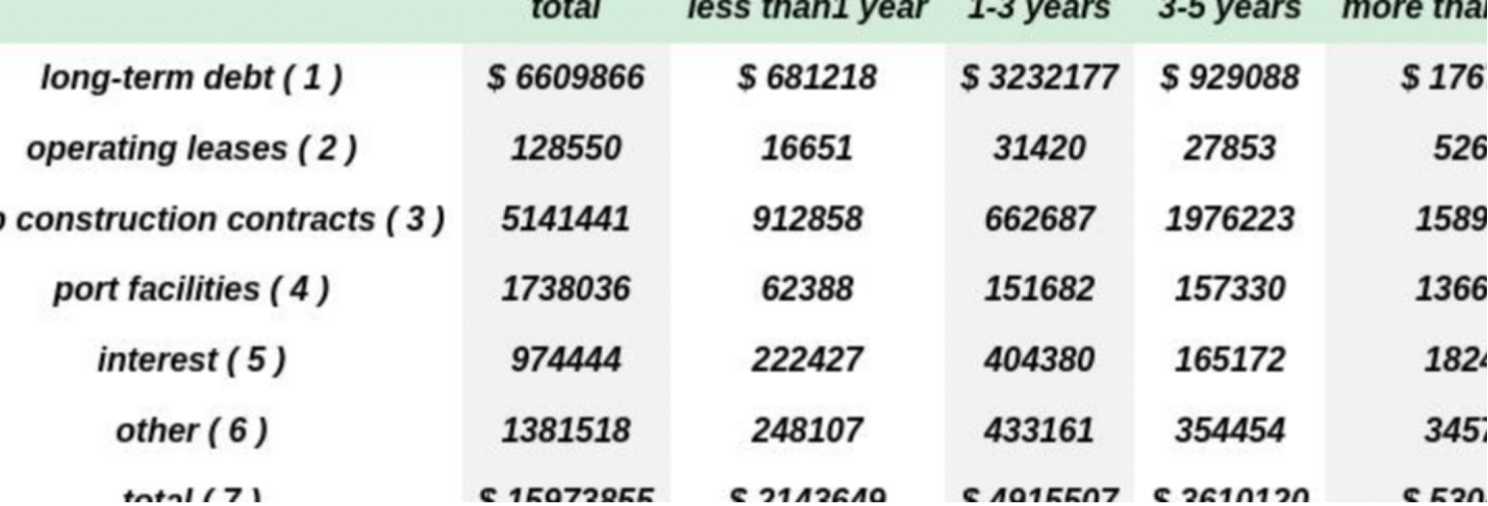}
}
\quad
\subfigure[incomplete]{
\includegraphics[width=\mysize]{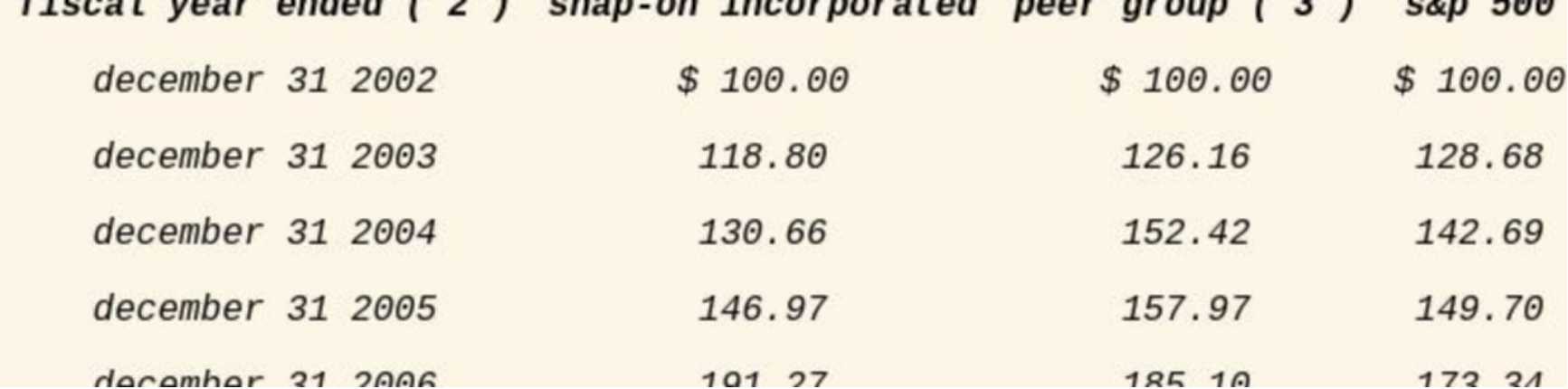}
}
\quad
\subfigure[skewed]{
\includegraphics[width=\mysize]{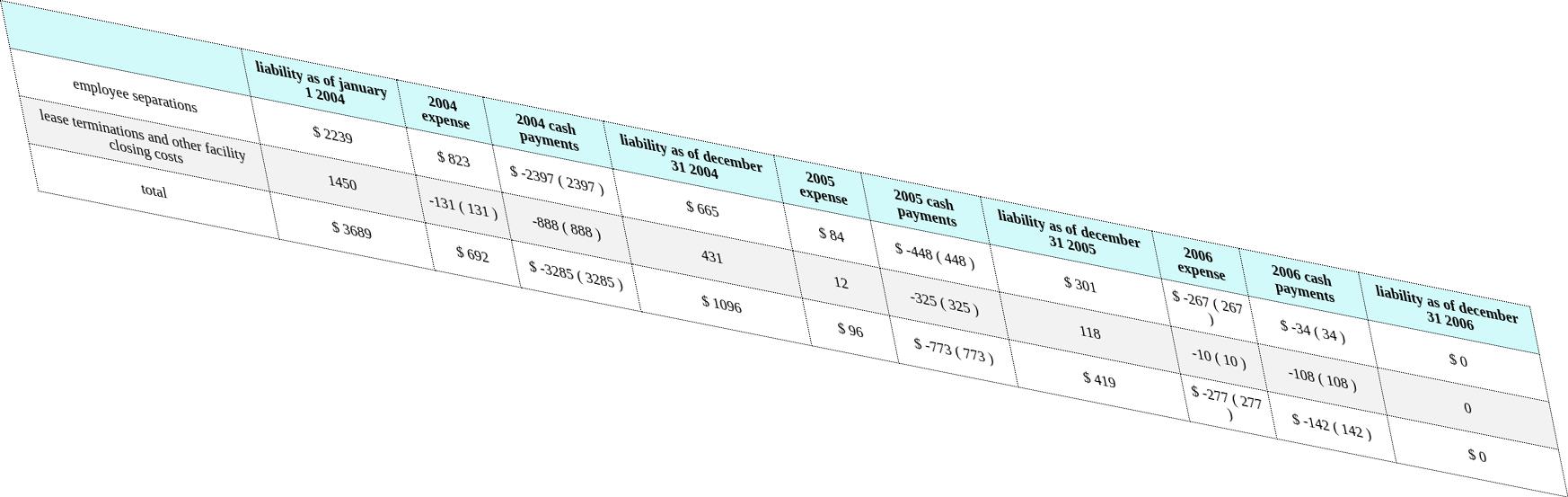}
}
\quad
\subfigure[skewed]{
\includegraphics[width=\mysize]{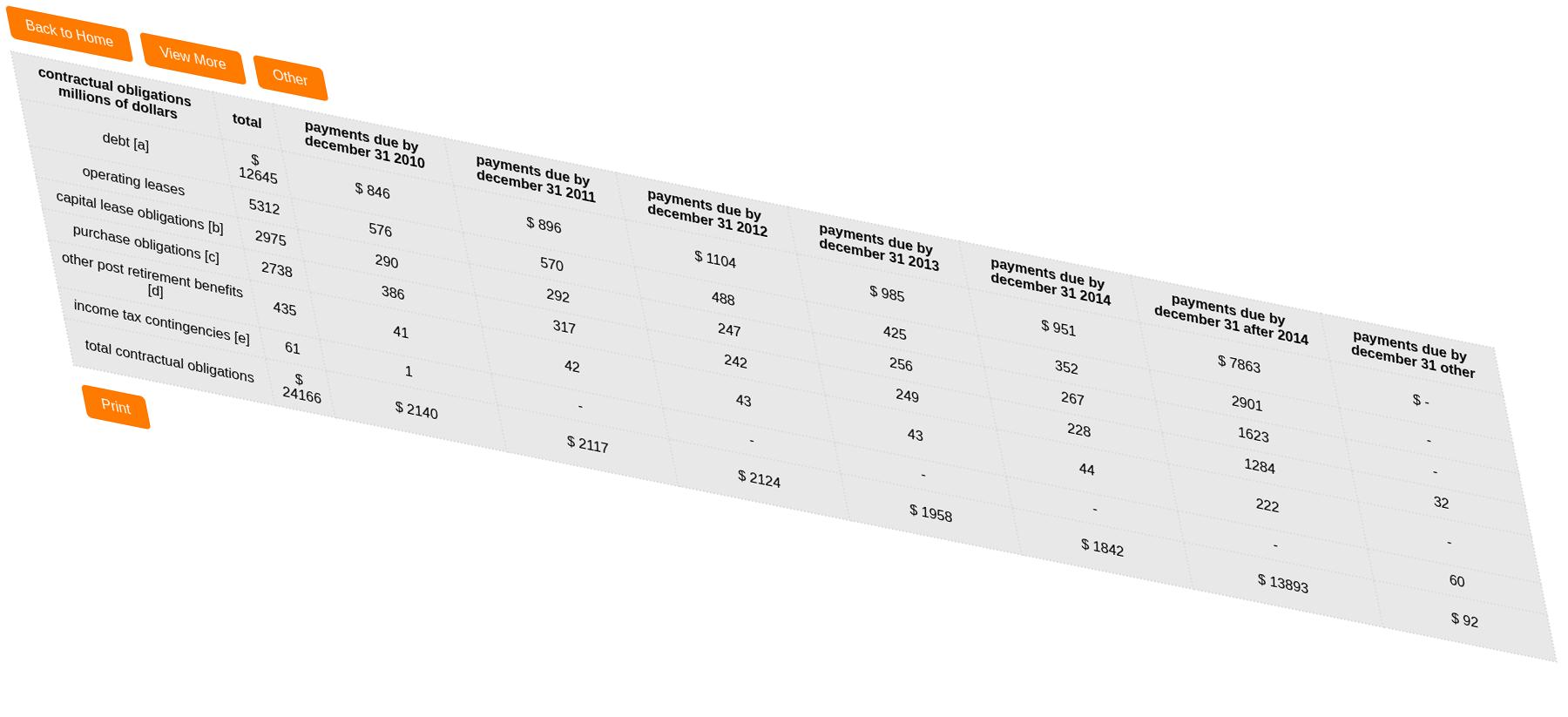}
}
\quad
\subfigure[skewed]{
\includegraphics[width=\mysize]{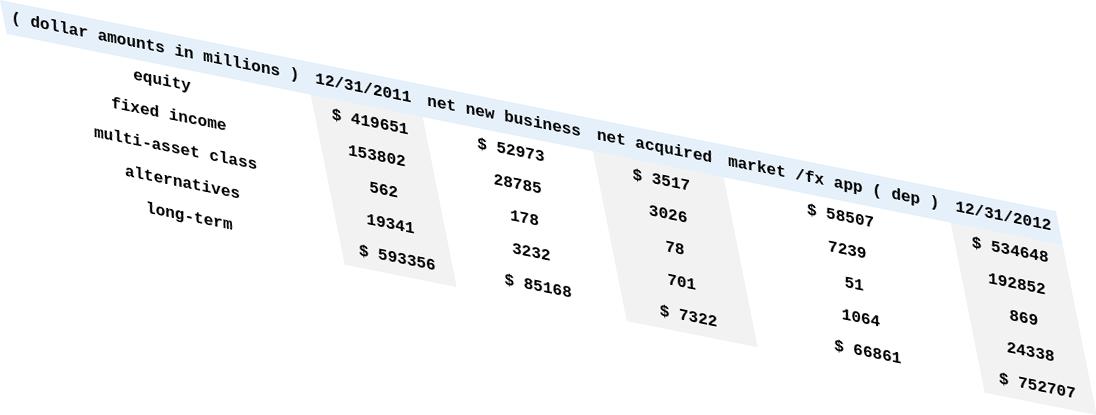}
}
\quad
\subfigure[blurred]{
\includegraphics[width=\mysize]{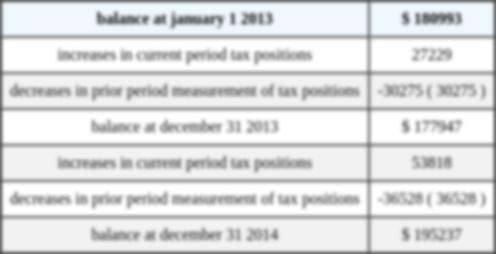}
}
\quad
\subfigure[blurred]{
\includegraphics[width=\mysize]{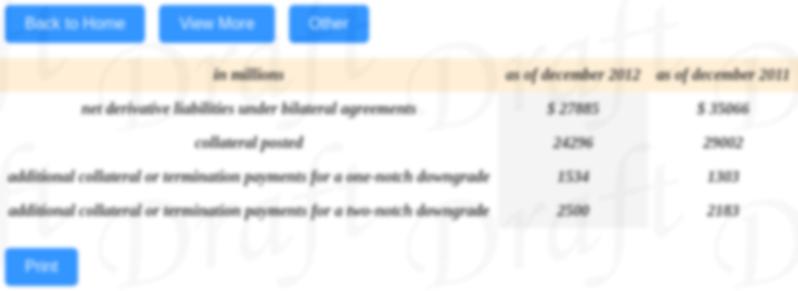}
}
\quad
\subfigure[blurred]{
\includegraphics[width=\mysize]{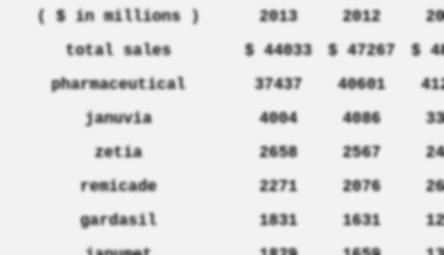}
}
\quad
\subfigure[watermark]{
\includegraphics[width=\mysize]{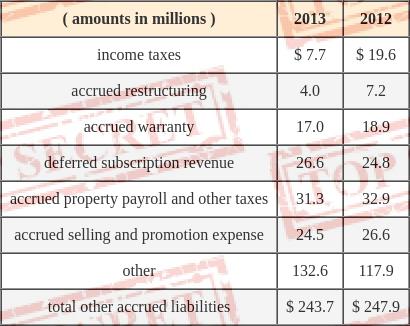}
}
\quad
\subfigure[watermark]{
\includegraphics[width=\mysize]{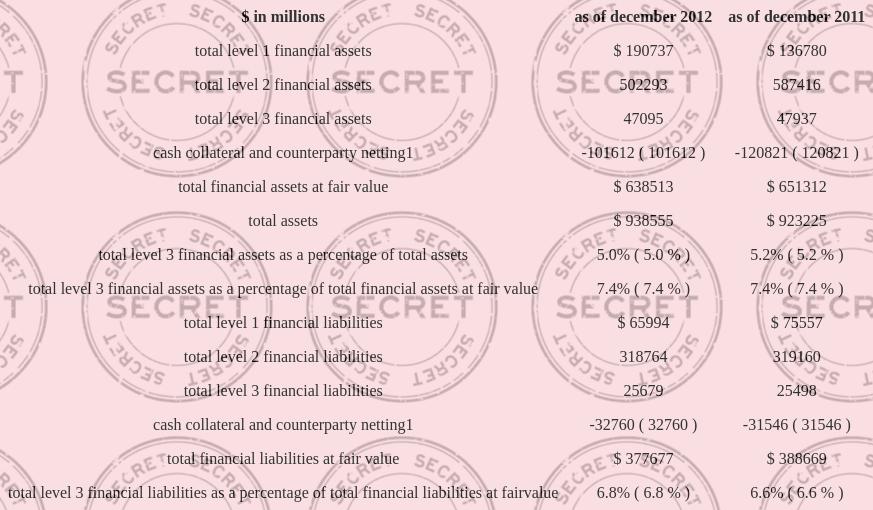}
}
\quad
\subfigure[watermark]{
\includegraphics[width=\mysize]{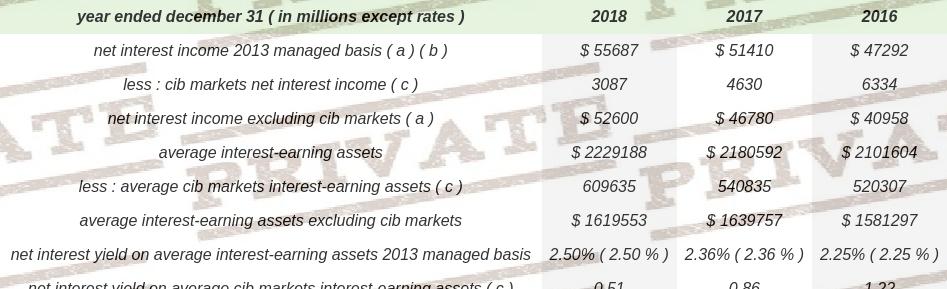}
}

\quad
\subfigure[compositional]{
\includegraphics[width=\mysize]{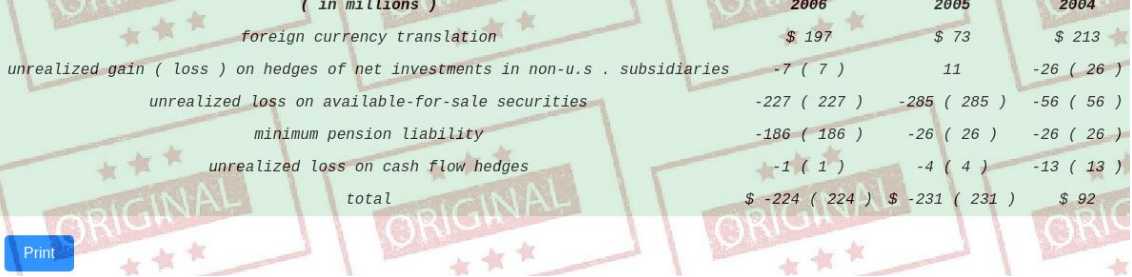}
}
\quad
\subfigure[compositional]{
\includegraphics[width=\mysize]{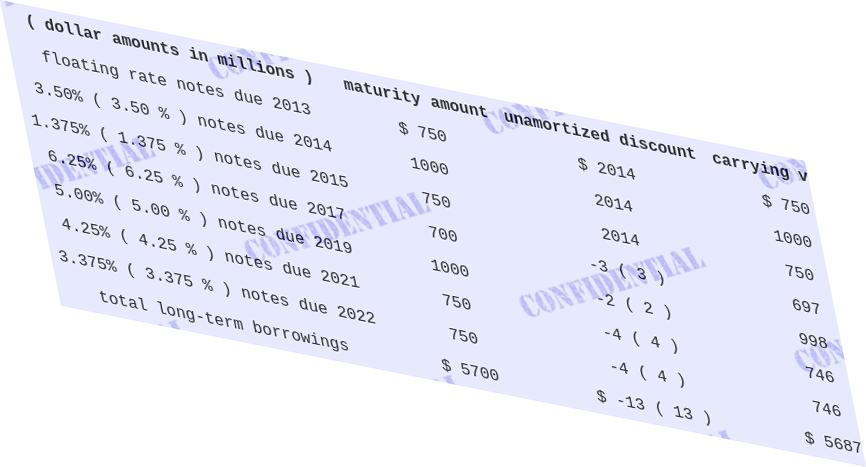}
}
\quad
\subfigure[compositional]{
\includegraphics[width=\mysize]{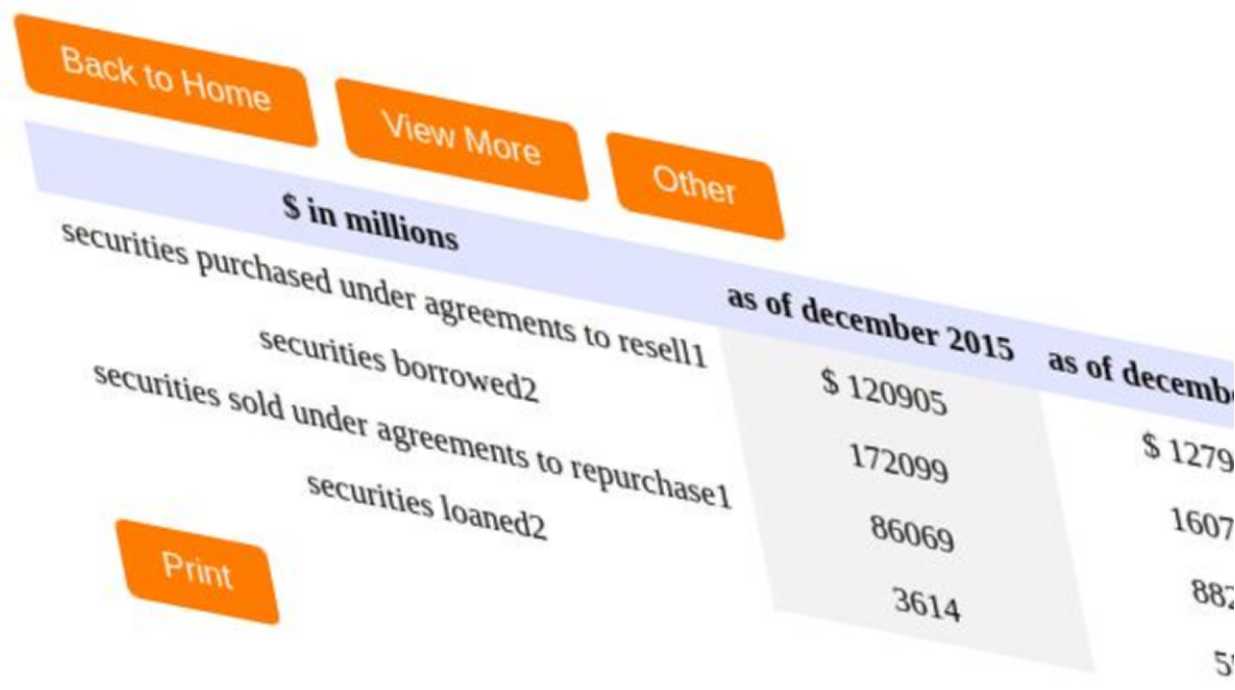}
}

\quad
\subfigure[compositional]{
\includegraphics[width=\mysize]{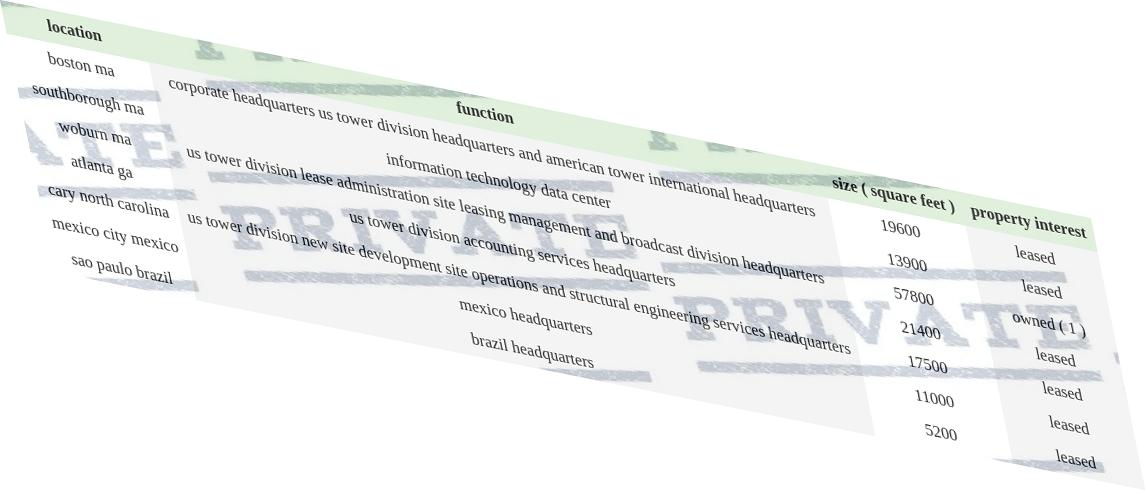}
}
\quad
\subfigure[compositional]{
\includegraphics[width=\mysize]{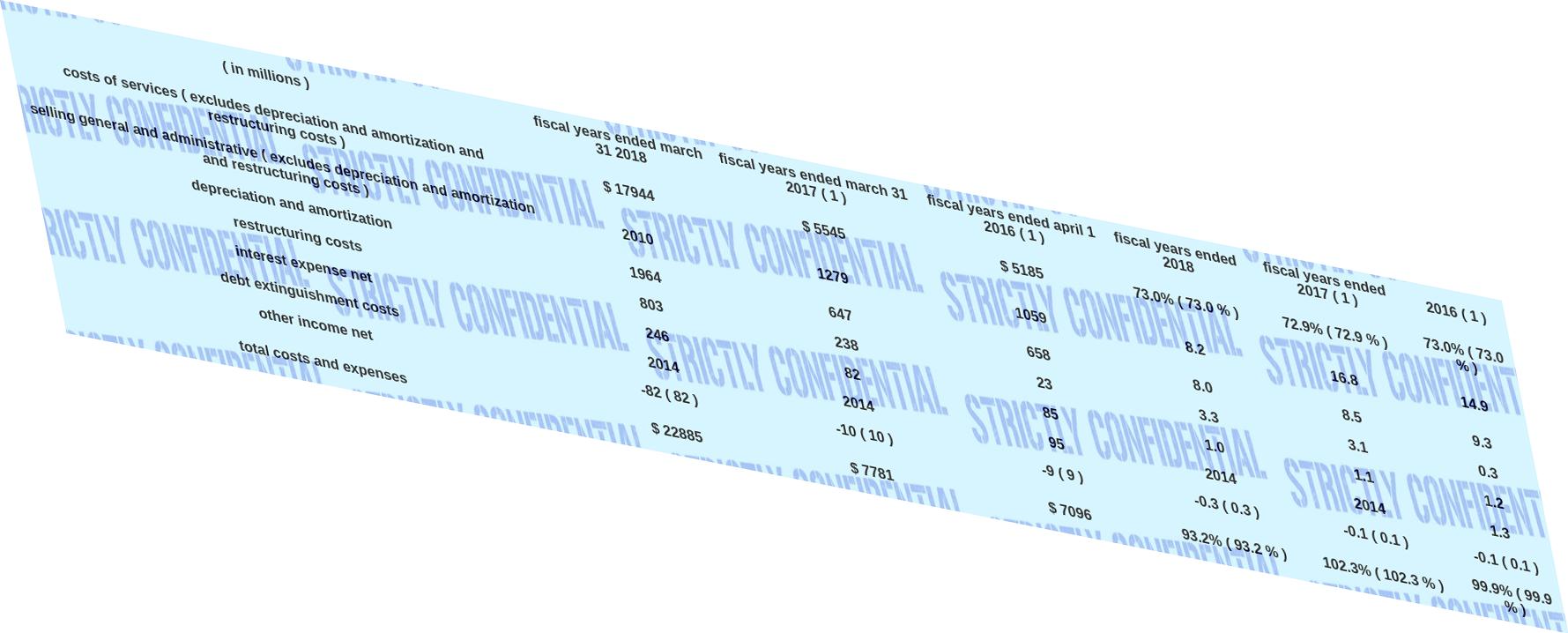}
}
\quad
\subfigure[compositional]{
\includegraphics[width=\mysize]{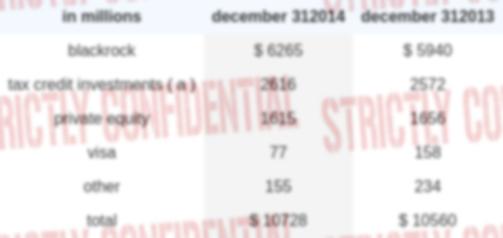}
}

\caption{\textit{WildStrcut} images demonstration of WMMFinQA-bench.}
\label{fig:WMMFinQA_bench}
\end{figure}

\begin{table}[htbp]
\small
\centering
\setlength{\abovecaptionskip}{1pt}%
\setlength{\belowcaptionskip}{0pt}%
 \caption{Baseline models' performance variation on incomplete table image versus complete table image for multimodal table reasoning tasks. For clarity, we only provide complete table image with intact information at the bottom of the table.}
 \label{tab:incom_com_tab_rsn_study}

\begin{tabular}{p{0.17\textwidth} p{0.77\textwidth}}
 \toprule
 \multicolumn{2}{l}{\textbf{Multimodal input instance: table image and domain knowledge}} \\
 \midrule[0.1pt]
& \begin{minipage}[p]{0.35\textwidth}
    \begin{figure}[H]
     \centering
       \subfigure[Incomplete screenshot table image]
{\includegraphics[scale=0.15,valign=c]{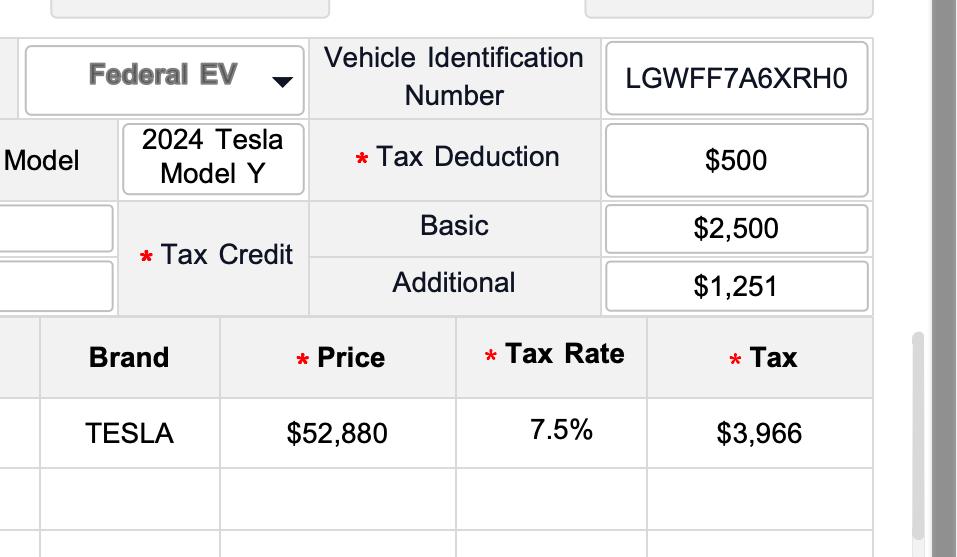}}
     \end{figure}
\end{minipage}
     \hspace{.06in}
     \ $\Rightarrow$
     \hspace{.001in}
\begin{minipage}[p]{0.35\textwidth}
\begin{figure}[H]
    \centering
     \subfigure[Original table image]
     {\includegraphics[scale=0.38,valign=c]{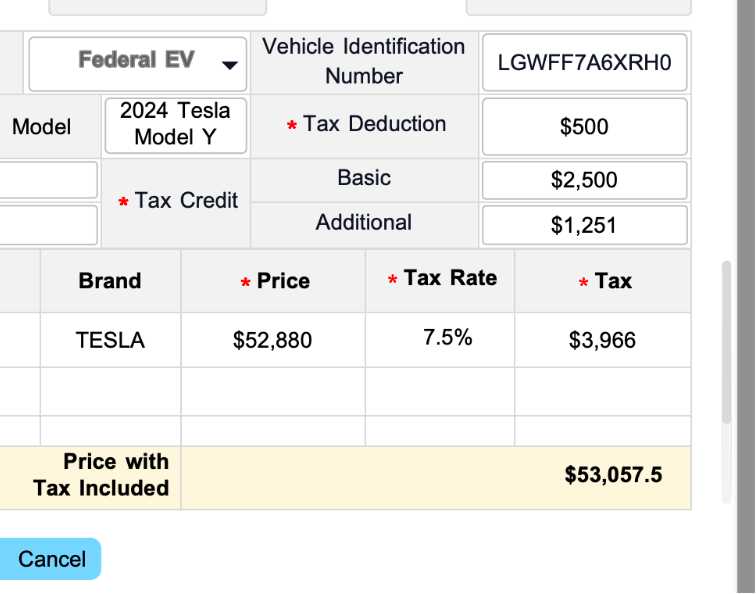}}
      \end{figure}
\end{minipage} \\  
  Domain \newline Knowledge &  \textit{Implementation Regulations of Enterprise Income Tax Law} stipulates the minimum depreciation life for cars purchased by enterprises is 4 years \ldots. Accumulated depreciation of a car is calculated as the formula: original value of fixed assets (excluding tax) = accumulated depreciation + net residual value. A deduction is an amount you subtract from your taxable price when you file so you don’t pay tax on it. By lowering your taxable price, deductions lower your tax. While a credit is an amount you subtract from the tax you owe. This can lower your tax payment \ldots.\\
\midrule[0.1pt]
 & Incomplete table image \\
\midrule[0.1pt]
 User & Given that depreciation life of the car in the image is 4 years and net residual value is 3\% of original value, what is monthly depreciation amount of the car?\\
 GPT-4 Turbo \newline with Vision & 
 \ldots \textcolor{red}{
 Original Value Excluding Tax = \$52,880 - \$3,966 = \$48,914.} \ldots.
\newline The answer is \$988.47. \textcolor{red}{\usym{2718}}\\
 User & What is the tax payable on the car in the image?\\
 GPT-4 Turbo \newline with Vision & The tax payable on the car in the image is \$3,966.
 \textcolor{red}{This is calculated based on the price of the car (\$52,880) and the tax rate (7.5\%) as shown in the image.} 
  \textcolor{red}{\usym{2718}}\\
 \midrule[0.1pt]
 User & Given that depreciation life of the car in the image is 4 years and net residual \ldots \\
GPT-4o & \ldots \textcolor{red}{
$\text{Original Value Excluding Tax} = \frac{\text{Price}}{1 + \text{Tax Rate}} = \frac{52880}{1 + 0.075} = \frac{52880}{1.075} \approx 49144.19$.}
Calculate the net residual value = Original Value Excluding Tax $\times$ 0.03 = \textcolor{red}{49144.19 $\times$ 0.03 $\approx 1474.33\$$.}
\newline Monthly depreciation amount = Total depreciation amount / Depreciation period = \textcolor{red}{$\frac{49144.19 - 1474.33}{48} = \frac{47669.86}{48} \approx 993.12\$$}
\newline The answer is \$993.12. \textcolor{red}{\usym{2718}}\\
User & What is the tax payable on the car in the image? \\
GPT-4o & \ldots. The tax payable on the car in the image is \$177.50. \textcolor{green}{\usym{2714}}\\
\midrule[0.1pt]
 & Complete table image \\
 \midrule[0.1pt]
  User & Given that depreciation life of the car in the image is 4 years and net residual value is 3\% of original value, what is monthly depreciation amount of the car?\\
 GPT-4 Turbo \newline with Vision &  \ldots. Calculate the Net Residual Value: Net Residual Value = 3\% $\times$ \$52,880 = \$1,586.40.
 \newline Calculate the Depreciable Base: Depreciable Base = \$52,880 - \$1,586.40 = \$51,293.60
 \newline Monthly Depreciation = $\frac{\$51,293.60}{48}$ = \$1,068.62. \ldots
\newline The answer is \$1,068.62. \textcolor{green}{\usym{2714}}\\
 User & What is the tax payable on the car in the image?\\
 GPT-4 Turbo \newline with Vision & The tax payable on the car in the image is \$3,966.
 \textcolor{red}{This is calculated based on the price of the car (\$52,880) and the tax rate (7.5\%), as shown in the image.} 
  \textcolor{red}{\usym{2718}}\\
 \midrule[0.1pt]
 
 User & Given that depreciation life of the car in the image is 4 years and net residual \ldots?\\
GPT-4o & \ldots. The original value excluding tax is \$52,880.
\newline Calculate the net residual value: Net residual value = \$52,880 * 0.03 = \$1,586.40
\newline Calculate the total depreciation amount: Total depreciation amount = \$52,880 - \$1,586.40 = \$51,293.60. 
\newline Calculate the monthly depreciation amount: Monthly depreciation amount = Total depreciation amount / Depreciation period = \$51,293.60 / 48 = \$1,068.62
\newline Therefore, the monthly depreciation amount of the car is \$1,068.62.
\textcolor{green}{\usym{2714}}\\
User & What is the tax payable on the car in the image? \\
GPT-4o & \ldots. The tax payable on the car in the image is \$177.50. \textcolor{green}{\usym{2714}}\\
\bottomrule
\end{tabular}
\end{table}

\begin{table*}[htbp]
\caption{Incomplete table image (incomplete screenshot taken by user) based reasoning via dialogues. \texttt{TableMoE} leverages MMTab-CoT to offer more easily interpretable and understandable responses, i.e., it indicates which modality the key information in the intermediate inference steps comes from, as shown by the highlighted colors 
 \fboxsep3pt
 \raisebox{0.15em}{\colorbox{icterine}{\rule{0pt}{0.2em}}} 
 \raisebox{0.15em}{\colorbox{guppiegreen}{\rule{0pt}{0.15em}}}
, whereas GPT-4o fails to do so. 
}
\label{tab:incom_tab_reasoning}
\begin{center}
\begin{small}
\begin{tabular}{p{0.17\textwidth} p{0.77\textwidth}}
\toprule
 \multicolumn{2}{l}{\textbf{Multimodal input instance: incomplete table image and domain knowledge}} \\
 \midrule[0.1pt]  
& \makecell[c]{\includegraphics[scale=0.17,valign=c]{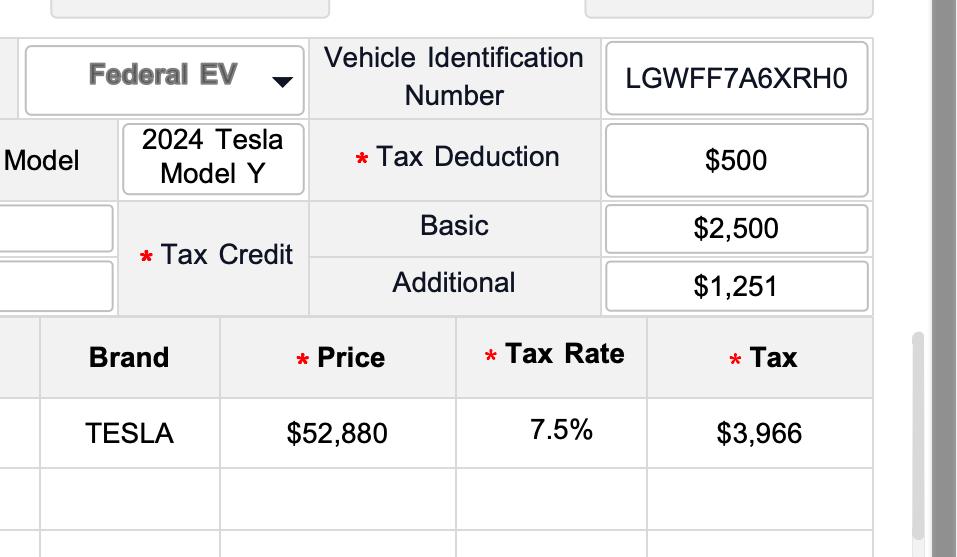}} \\  
  Domain \newline Knowledge &  \textit{Implementation Regulations of Enterprise Income Tax Law} stipulates the minimum depreciation life for cars purchased by enterprises is 4 years \ldots. Accumulated depreciation of a car is calculated as the formula: original value of fixed assets (excluding tax) = accumulated depreciation + net residual value. A deduction is an amount you subtract from your taxable price when you file so you don’t pay tax on it. By lowering your taxable price, deductions lower your tax. While a credit is an amount you subtract from the tax you owe. This can lower your tax payment \ldots.\\
\midrule[0.1pt]
 & 
 [\textit{\#Turn 1}] \\
User & Given that depreciation life of the car in the image
is 4 years and net residual value is 3\% of original value, 
what is monthly depreciation amount of the car? 
\\
TableMoE
& \textbf{Generated Rationale:}
\newline Let's think step by step. 
\newline From \colorbox{icterine}{the domain knowledge}, we know that the formula for calculating annual depreciation amount of a car is: \textcolor{blue}{Annual Depreciation Amount = (Original Value excluding Tax - Net Residual Value) / Depreciation Life}. 
\newline \colorbox{guppiegreen}{The table image} shows that \textcolor{blue}{original value of the car excluding tax is \$52880} . 
\newline Net residual value of the car = original value of the car excluding tax * 3\% = 52880 * 3\% = 1586.4.
\newline Substituting the value into annual depreciation amount formula, Annual Depreciation Amount = (52880 - 1586.4) / 4 = 12823.4.
\newline \textcolor{blue}{Monthly depreciation amount of the car = 12823.4 / 12 = 1068.6167}. 
\newline \textbf{Answer:}
\newline The answer is \$1068.62. \textcolor{green}{\usym{2714}}\\
 & 
 [\textit{\#Turn 2}]  \\
 User & What is the tax payable on the car in the image? \\
 TableMoE & 

Let's think step by step.
\newline \colorbox{icterine}{The domain knowledge} and \colorbox{guppiegreen}{the table image} provide the information to derive the formula for calculating tax payable is: \textcolor{blue}{Tax Payable = Tax - Tax Credit = 
Taxable Price * Tax Rate - Tax Credit = (Original Price - Tax Deduction) * Tax Rate - (Basic Tax Credit + Additional Tax Credit)}. \newline\colorbox{guppiegreen}{The table image} indicates \textcolor{blue}{original price is \$52880} and tax rate is 7.5\%. \colorbox{guppiegreen}{The table image} explicitly shows that tax deduction is \$500. \colorbox{guppiegreen}{The table image} also lists basic tax credit and additional tax credit as \$2500 and \$1251 respectively.
\newline Plugging in the values: Tax Payable = (52880 - 500) * 7.5\% - (2500 + 1251) = 177.5.
\newline The answer is \$177.5. \textcolor{green}{\usym{2714}}\\
\midrule[0.1pt]
 User & Given \ldots , what is the monthly depreciation amount of the car in the image? \\
 GPT-4 Turbo \newline with Vision & \ldots \textcolor{red}{
 Original Value Excluding Tax = \$52,880 - \$3,966 = \$48,914.} \ldots.
\newline The answer is \$988.47. \textcolor{red}{\usym{2718}}\\
 User & What is the tax payable on the car in the image?\\
 GPT-4 Turbo \newline with Vision & The tax payable on the car in the image is \$3,966.
 \textcolor{red}{This is calculated based on the price of the car (\$52,880) and the tax rate (7.5\%) as shown in the image.} 
  \textcolor{red}{\usym{2718}}\\
 \midrule[0.1pt]
User & Given \ldots , what is the monthly depreciation amount of the car in the image? \\
GPT-4o & \ldots \textcolor{red}{
$\text{Original Value Excluding Tax} = \frac{\text{Price}}{1 + \text{Tax Rate}} = \frac{52880}{1 + 0.075} = \frac{52880}{1.075} \approx 49144.19$.} 
\ldots. 
\newline The answer is \$993.12. \textcolor{red}{\usym{2718}}\\
User & What is the tax payable on the car in the image? \\
GPT-4o & \ldots. The tax payable on the car in the image is \$177.50. \textcolor{green}{\usym{2714}}\\
\bottomrule
\end{tabular}
\end{small}
\end{center}
\end{table*}
\begin{table}[htbp]
\small
\centering
\setlength{\abovecaptionskip}{1pt}%
\setlength{\belowcaptionskip}{0pt}%
\caption{Visualization results of an instance of multimodal table reasoning on WMMTabDialog are provided to illustrate the performance differences across various models.} 
\label{tab:wmmtab_rsn_chat_01}
\begin{tcolorbox}[colframe=black,colback=whitesmoke]
\begin{minipage}{\linewidth}
\centering
\includegraphics[scale=0.43]{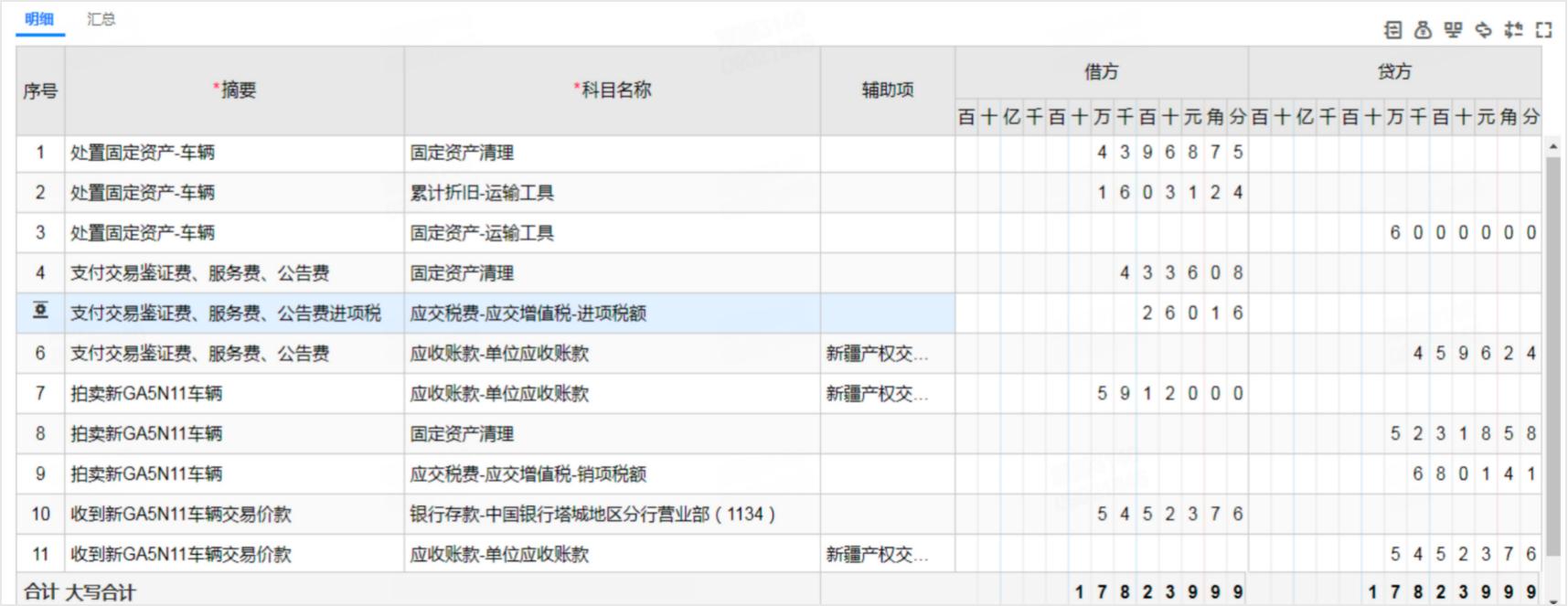}
\end{minipage}

\begin{minipage}{\linewidth}
\flushleft
\hspace*{\fill}
\hspace*{\fill}
\newline \textcolor{amber(sae/ece)}{\textbf{Question:}} 
\newline 
\vspace{0.01in}
\begin{CJK*}{UTF8}{gbsn}
\small
表中大写合计应该填写多少？
\end{CJK*}
\newline How much the total amount in words in the table should be filled?
\newline 
\newline \textcolor{forestgreen(web)}{\textbf{Ground Truth:}} \newline \begin{CJK*}{UTF8}{gbsn}
\small
壹拾柒万捌仟贰佰叁拾玖元玖角玖分
\end{CJK*}
\newline SAY YUAN ONE HUNDRED AND SEVENTY-EIGHT THOUSAND,TWO HUNDRED AND THIRTY-NINE AND CENTS NINETY-NINE ONLY
\newline
\newline\textcolor{blue}{\textbf{Qwen2-VL-72B:}} \newline 
\begin{CJK*}{UTF8}{gbsn}
\small 
表中大写合计应该填写:\textcolor{red}{壹佰柒拾捌亿贰仟叁佰玖拾玖万玖仟玖佰元}。\textcolor{red}{\usym{2718}}
\newline The total amount in words should be filled in as: \textcolor{red}{SAY YUAN AND SEVENTEEN BILLION,EIGHT HUNDRED AND TWENTY-THREE MILLION,NINE HUNDRED AND NINETY-NINE THOUSAND,NINE-HUNDRED AND CENTS ONLY}.
\textcolor{red}{\usym{2718}}
\end{CJK*}
\newline
\newline
\textcolor{blue}{\textbf{Qwen-VL-Max:}} \newline 
\begin{CJK*}{UTF8}{gbsn}
\small
根据表格中的数据，合计金额为\textcolor{red}{178239999元}，大写为\textcolor{red}{壹亿柒仟捌佰贰拾叁万玖仟玖佰玖拾玖元}。\textcolor{red}{\usym{2718}}
\newline According to the data in the table, the total amount is \textcolor{red}{178,239,999 yuan}, the total amount in words is: \textcolor{red}{SAY YUAN ONE HUNDRED AND SEVENTY-EIGHT MILLION,TWO HUNDRED AND THIRTY-NINE THOUSAND,NINE HUNDRED AND NINETY-NINE AND CENTS ONLY}. \textcolor{red}{\usym{2718}}
\end{CJK*}
\newline
\newline
\textcolor{blue}{\textbf{GPT-4 Turbo with Vision:}} \newline 
\begin{CJK*}{UTF8}{gbsn}
\small
合计对应的是\textcolor{red}{第5行的“金额”列}。根据表格中的数据，该单元格应填写的数值是\textcolor{red}{“26016”}。\textcolor{red}{\usym{2718}}\end{CJK*}
\newline The total amount corresponds to \textcolor{red}{the “Amount” column in row 5}. According to the data in the table, the cell should be filled with the value \textcolor{red}{“26016”}. \textcolor{red}{\usym{2718}}
\newline
\newline
\textcolor{blue}{\textbf{GPT-4o:}} \newline 
\begin{CJK*}{UTF8}{gbsn}
\small
根据图片中的数字合计，借方总金额\textcolor{red}{178,823,999元}。因此，合计大写金额应填写为：\textcolor{red}{壹亿柒仟捌佰贰拾捌万贰仟叁佰玖拾玖元整}。\textcolor{red}{\usym{2718}}
\end{CJK*}
\newline According to the total number in the picture, the total debit amount is \textcolor{red}{178,823,999 yuan}. Therefore, the total amount in words should be filled in as: \textcolor{red}{SAY YUAN ONE HUNDRED AND SEVENTY-EIGHT MILLION,EIGHT HUNDRED AND TWENTY-THREE THOUSAND,NINE HUNDRED AND NINETY-NINE AND CENTS ONLY}. \textcolor{red}{\usym{2718}}
\newline
\newline
\textcolor{blue}{\textbf{TableMoE \textbf{(Ours):}}} 
\newline\begin{CJK*}{UTF8}{gbsn}
\small
让我们一步一步来思考:
\newline 大写合计应该填写借方或者贷方的金额总和，表格图片显示两者金额总和均为1十7万8千2百3十9元9角9分，
\newline 因此对应的大写金额为：壹拾柒万捌仟贰佰叁拾玖元玖角玖分。\textcolor{green}{\usym{2714}}
\end{CJK*}
\newline Let's think step by step:
\newline The total amount in words at the bottom of the table image is the total amount of debit or credit.  The table image shows that the total amount of both is 178,239.99 yuan.
\newline So the corresponding amount in words is: SAY YUAN ONE HUNDRED AND SEVENTY-EIGHT THOUSAND,TWO HUNDRED AND THIRTY-NINE AND CENTS NINETY-NINE ONLY. \textcolor{green}{\usym{2714}} 
\\
\end{minipage}
\end{tcolorbox}
\end{table}

\begin{table}[htbp]
\small
\centering
\setlength{\abovecaptionskip}{1pt}%
\setlength{\belowcaptionskip}{0pt}%
\caption{Visualization results of an instance of multimodal table reasoning on WMMTabDialog are provided to illustrate the performance differences across various models.} 
\label{tab:ref_rsn_chat_02}
\begin{tcolorbox}[colframe=black,colback=whitesmoke]
\begin{minipage}{\linewidth}
\centering
\includegraphics[scale=0.15]{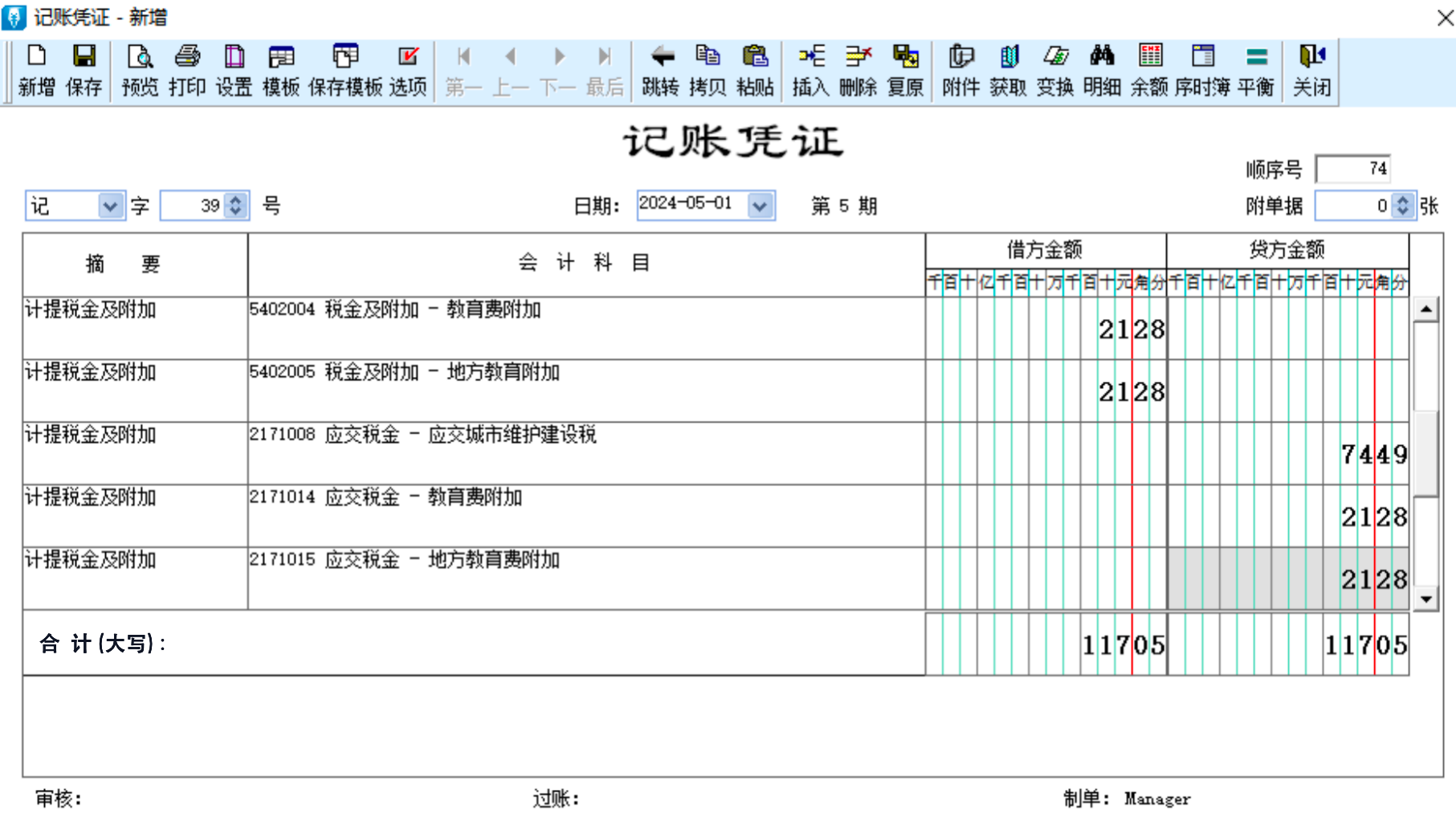}
\end{minipage}

\begin{minipage}{\linewidth}
\flushleft
\hspace*{\fill}
\hspace*{\fill}
\newline \textcolor{amber(sae/ece)}{\textbf{Question:}} 
\newline 
\begin{CJK*}{UTF8}{gbsn}
\small
表中合计(大写)应该填写多少？
\end{CJK*}
\newline How much should be filled in the total amount in words?
\newline 
\newline \textcolor{forestgreen(web)}{\textbf{Ground Truth:}} \newline \begin{CJK*}{UTF8}{gbsn}
\small
壹佰壹拾柒元伍分
\end{CJK*}
\newline SAY YUAN ONE HUNDRED AND SEVENTEEN AND CENTS FIVE ONLY
\newline
\newline\textcolor{blue}{\textbf{Qwen2-VL-72B:}} \newline 
\begin{CJK*}{UTF8}{gbsn}
\small
 表中合计(大写)应该填写“\textcolor{red}{壹万萬壹仟柒佰陆拾}” \textcolor{red}{\usym{2718}}
\newline The total (in words) in the table should be filled in as "\textcolor{red}{SAY YUAN AND ELEVEN THOUSAND,SEVEN HUNDRED AND SIXTY AND CENTS ONLY}" \textcolor{red}{\usym{2718}}
\end{CJK*}
\newline
\newline
\textcolor{blue}{\textbf{GPT-4 Turbo with Vision:}} \newline 
\begin{CJK*}{UTF8}{gbsn}
\small
在图像中显示的是一个财务软件界面，其中包含了一些账目信息。在“合计”一栏中，我们可以看到数字\textcolor{red}{11705}。在中国的财务文档中，大写金额通常用于正式文件和法律文件中，以防止篡改。\textcolor{red}{数字11705的大写形式是“壹万壹仟柒佰零伍元整”}。因此，在“表中合计(大写)”的位置，应填写“\textcolor{red}{壹万壹仟柒佰零伍元整}”。\textcolor{red}{\usym{2718}}\end{CJK*}
\newline The image shows a financial software interface that contains some accounting information. In the “Total” column, we can see the number \textcolor{red}{11705}. In Chinese financial documents, The amount in words are usually used in official and legal documents to prevent tampering. \textcolor{red}{The number 11705 in words is “\textcolor{red}SAY YUAN AND ELEVEN THOUSAND,SEVEN HUNDRED AND FIVE AND CENTS ONLY”}. Therefore, in the position of “Total (in words) in the table”, it should be filled in as “\textcolor{red}{SAY YUAN AND ELEVEN THOUSAND,SEVEN HUNDRED AND FIVE AND CENTS ONLY}”. \textcolor{red}{\usym{2718}}
\newline 
\newline
\textcolor{blue}{\textbf{GPT-4o:}} \newline 
\begin{CJK*}{UTF8}{gbsn}
\small
表中合计金额为11705，转换为大写应填写为：\textcolor{red}{壹万壹仟柒佰零伍}。\textcolor{red}{\usym{2718}}
\end{CJK*}
\newline{The total amount in the table is \textcolor{red}{11,705}, which should be written in words as: \textcolor{red}{SAY YUAN AND ELEVEN THOUSAND,SEVEN HUNDRED AND FIVE AND CENTS ONLY}.\textcolor{red}{\usym{2718}}}
\newline
\newline
\textcolor{blue}{\textbf{TableMoE \textbf{(Ours):}}} 
\newline\begin{CJK*}{UTF8}{gbsn}
\small
让我们一步一步来思考:
\newline 应该填写借方或者贷方的金额总和，表格图片显示两者金额总和均为1十7万8千2百3十9元9角9分。
\newline 因此对应的大写金额为：壹拾柒万捌仟贰佰叁拾玖元玖角玖分。\textcolor{green}{\usym{2714}}
\end{CJK*}
\newline Let’s think step by step:
\newline The total amount of debit or credit should be filled in. The table image shows that the total amount of both is 178,239.99 yuan.
\newline So the corresponding amount in words is: SAY YUAN ONE HUNDRED AND SEVENTY-EIGHT THOUSAND,TWO HUNDRED AND THIRTY-NINE AND CENTS NINETY-NINE ONLY. \textcolor{green}{\usym{2714}} 
\\
\end{minipage}
\end{tcolorbox}
\end{table}

\begin{table}[htbp]
\small
\centering
\setlength{\abovecaptionskip}{1pt}%
\setlength{\belowcaptionskip}{0pt}%
\caption{Visualization results of an instance of symbolic expression reasoning on WMMTabDialog are provided to illustrate the performance difference across various models.} 
\label{tab:wmmtab_se_rsn_01}
\begin{tcolorbox}[colframe=black,colback=whitesmoke]
\begin{minipage}{\linewidth}
\centering
\includegraphics[scale=0.41]{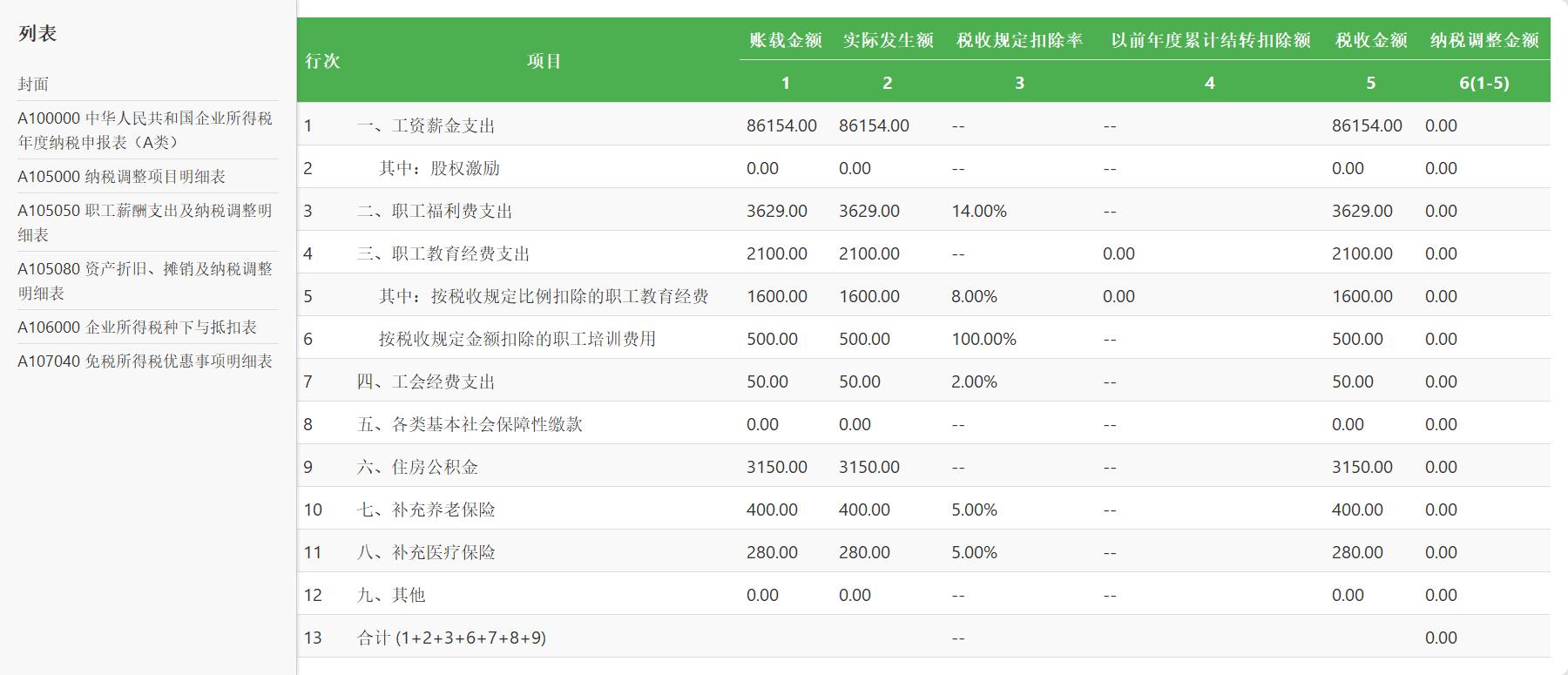}
\end{minipage}
\begin{minipage}{\linewidth}
\flushleft
\hspace*{\fill}
\hspace*{\fill}
\newline \textcolor{amber(sae/ece)}{\textbf{Question:}} 
\newline 
\begin{CJK*}{UTF8}{gbsn}
\small
账载金额合计应该填写多少？
\end{CJK*}
\newline How much the total carrying amount should be filled?
\newline 
\newline \textcolor{forestgreen(web)}{\textbf{Ground Truth:}} \newline
93483.00
\newline
\newline\textcolor{blue}{\textbf{Qwen2-VL-72B:}}  
\begin{CJK*}{UTF8}{gbsn}
\small 
\newline 账载金额合计应该填写 139748.00 元。
\newline 计算方法如下: 1. 工资薪金支出:86154.00 \quad 2. 职工福利费支出:3629.00 \quad 3. 职工教育经费支出:2100.00 \quad 4. 工会经费支出:50.00 \quad 5. 住房公积金:3150.00 \quad 6. 补充养老保险:400.00 \quad 7. 补充医疗保险:280.00.
\newline 合计: \textcolor{red}{86154.00 + 3629.00 + 2100.00 + 50.00 + 3150.00 + 400.00 + 280.00 = 139748.00}.  \textcolor{red}{\usym{2718}}
\newline The total amount of the account should be 139,748.00 yuan.
\newline Calculation method is as follows: 1. Salary and wage expenses:86154.00 \quad 2. Employee welfare expenses:3629.00 \quad 3. Employee education expenses:2100.00 \quad 4. Labor union expenses:50.00 
\quad 5. Housing Provident Fund:3150.00 
\quad 6. Supplementary Pension Insurance:400.00 
\quad 7. Supplementary Medical Insurance: 280.00.
\newline Total: \textcolor{red}{86154.00 + 3629.00 + 2100.00 + 50.00 + 3150.00 + 400.00 + 280.00 = 139748.00}.  \textcolor{red}{\usym{2718}}
\end{CJK*}
\newline
\newline
\textcolor{blue}{\textbf{GPT-4 Turbo with Vision:}} \newline 
\begin{CJK*}{UTF8}{gbsn}
\small
在图中的最后一行，显示的是合计金额，\textcolor{red}{即所有项的总和}。根据图中的数据，合计金额为：\textcolor{red}{86154.00 + 0.00 + 3629.00 + 2100.00 + 1600.00 + 500.00 + 50.00 + 0.00 + 3150.00 + 400.00 + 280.00 + 0.00 = 99963.00}。因此，账载金额合计应填写为 99963.00。\textcolor{red}{\usym{2718}}\end{CJK*}
\newline 
The total amount in the last row of the table image is \textcolor{red}{the sum of all the items} as shown. Based on the data in the image, the total amount is: \textcolor{red}{86154.00 + 0.00 + 3629.00 + 2100.00 + 1600.00 + 500.00 + 50.00 + 0.00 + 3150.00 + 400.00 + 280.00 + 0.00 = 99963.00}. Therefore, the total carrying amount should be filled as 99963.00. \textcolor{red}{\usym{2718}}
\newline
\newline
\textcolor{blue}{\textbf{GPT-4o:}} \newline 
\begin{CJK*}{UTF8}{gbsn}
\small
账载金额合计填写\textcolor{red}{86154.00}。\textcolor{red}{\usym{2718}}
\end{CJK*}
\newline The total carrying amount is \textcolor{red}{86154.00}. \textcolor{red}{\usym{2718}}
\newline
\newline
\textcolor{blue}{\textbf{TableMoE \textbf{(Ours):}}} 
\newline\begin{CJK*}{UTF8}{gbsn}
\small
让我们一步一步来思考: 根据表格中的最后一行“合计(1+2+3+6+7+8+9)”，我们需要将这些行的“预算金额”列的数值相加。
\newline从表格中提取相应的数值：
\quad- 第1行：86154.00
\quad- 第2行：0.00
\quad- 第3行：3629.00
\quad- 第6行：500.00
\quad- 第7行：50.00
\quad- 第8行：0.00
\quad- 第9行：3150.00。
\newline 将这些数值相加得到：86154.00 + 0.00 + 3629.00 + 500.00 + 50.00 + 0.00 + 3150.00 = 93483.00。
\newline因此，账载金额合计应填写为 93483.00。\textcolor{green}{\usym{2714}}
\end{CJK*}
\newline 
Let's think step by step: According to the last line of the table “Total (1+2+3+6+7+8+9)”, we need to add up the values of the “Budget Amount” columns of these lines.
\newline Extract the corresponding values from the table:
\quad - Row 1: 86154.00
\quad- Row 2: 0.00
\quad - Row 3: 3629.00
\quad - Row 6: 500.00
\quad - Row 7: 50.00
\quad - Row 8: 0.00
\quad - Row 9: 3150.00.
\newline Add these values to get:
 86154.00 + 0.00 + 3629.00 + 500.00 + 50.00 + 0.00 + 3150.00 = 93483.00.  
\newline The total of the carrying amount should therefore be entered as 93483.00. \textcolor{green}{\usym{2714}}
\end{minipage}
\end{tcolorbox}
\end{table}


\begin{table}[!htbp]
\small
\centering
\setlength{\abovecaptionskip}{1pt}%
\setlength{\belowcaptionskip}{0pt}%
\caption{
Visualization results of multimodal table-centric reasoning with checkbox-related query on WMMTabDialog to illustrate the performance differences across various models.} 
\label{tab:wmmtab_checkbox_rsn_01}
\begin{tcolorbox}[colframe=black,colback=whitesmoke]
\begin{minipage}{\linewidth}
\centering
\includegraphics[scale=0.45]{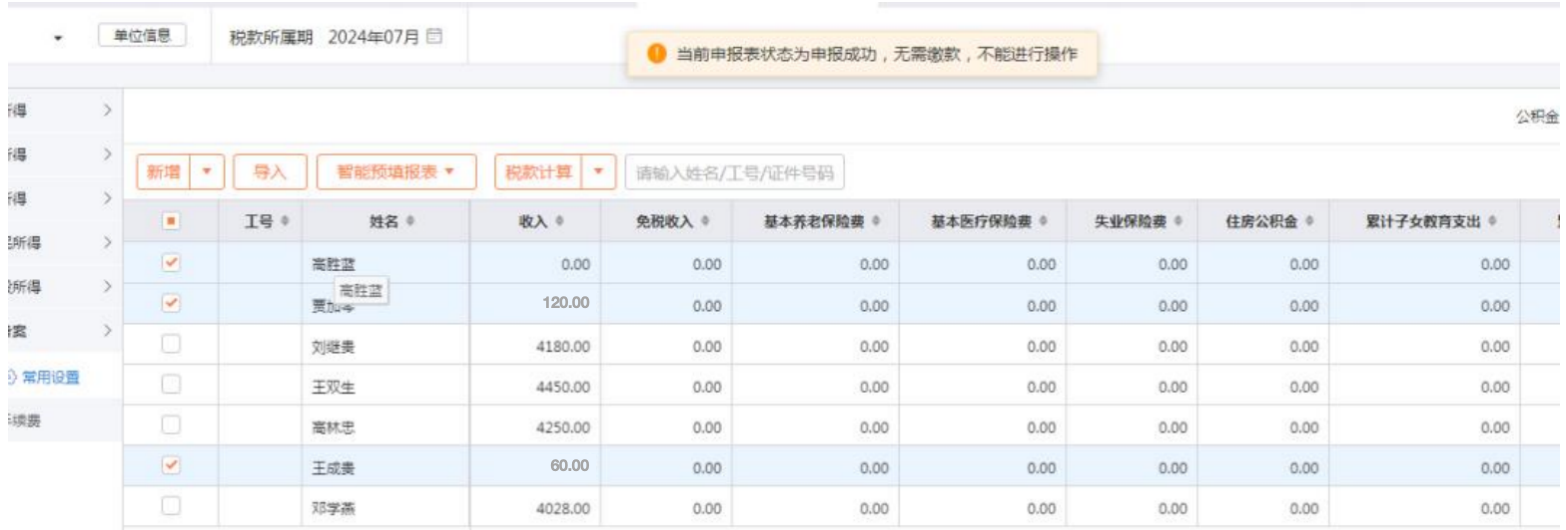}
\end{minipage}

\begin{minipage}{\linewidth}
\flushleft
\hspace*{\fill}
\hspace*{\fill}
\newline \textcolor{amber(sae/ece)}{\textbf{Question:}} 
\newline 
\begin{CJK*}{UTF8}{gbsn}
\small
根据表格图像中信息，勾选人员的收入合计是多少？
\end{CJK*}
\newline 
Based on the information in the table image, what is the total income of the people checked?
\newline 
\newline \textcolor{forestgreen(web)}{\textbf{Ground Truth:}} 
\newline 180.00
\newline
\newline\textcolor{blue}{\textbf{Qwen2.5-VL-72B-Instruct:}} \newline 
\begin{CJK*}{UTF8}{gbsn}
\small 
根据表格图像中勾选人员的收入信息，勾选人员的收入合计如下：
\newline 高胜蓝：0.00
\newline \textcolor{red}{王双生：4450.00}
\newline 王成贵：60.00
\newline 将这些数值相加：\textcolor{red}{0.00 + 4450.00 + 60.00 = 4510.00}
\newline 因此，勾选人员的收入合计是 \textcolor{red}{4510.00} \textcolor{red}{\usym{2718}}
\end{CJK*}
\newline According to the income information of the people checked in the table image, the total income of the people checked is as follows: 
\newline Gao Shenglan: 0.00
\newline \textcolor{red}{Wang Shuangsheng: 4450.00}
\newline Wang Chenggui: 60.00
\newline Add these values: \textcolor{red}{0.00 + 4450.00 + 60.00 = 4510.00}
\newline Therefore, the total income of the checked people is \textcolor{red}{4510.00} \textcolor{red}{\usym{2718}}
\newline
\textcolor{blue}{\textbf{GPT-4 Turbo with Vision:}} \newline 
\begin{CJK*}{UTF8}{gbsn}
\small
根据图片中的信息，勾选的人员（刘建华、王宝强、王志强）的收入合计为：\newline - 刘建华：4180.00 - 王宝强：4450.00 - 王志强：4250.00 \newline合计：\textcolor{red}{4180.00 + 4450.00 + 4250.00 = 12880.00} \newline所以，勾选人员收入合计是12880.00。\textcolor{red}{\usym{2718}}\end{CJK*}
\newline 
According to the information in the image, the total income of the three people checked is: \newline
- Liu Jianhua: 4180.00 - Wang Baoqiang: 4450.00 - Wang Zhiqiang: 4250.00
\newline \textcolor{red}{Total: 4180.00 + 4450.00 + 4250.00 = 12880.00}
\newline Therefore, the total income of the three people who checked is 12880.00. \textcolor{red}{\usym{2718}}
\newline
\textcolor{blue}{\textbf{GPT-4o:}} \newline 
\begin{CJK*}{UTF8}{gbsn}
\small
根据图中信息，勾选人员的收入分别是：\newline - 常艳萍：0.00 - 常艳萍：0.00 - 刘建生：4180.00\newline 收入合计为：\textcolor{red}{0.00 + 0.00 + 4180.00 = 4180.00}。\textcolor{red}{\usym{2718}}
\end{CJK*}
\newline 
According to the information in the image, the income of the three people checked are: \newline
- Chang Yanping: 0.00 - Chang Yanping: 0.00 - Liu Jiansheng: 4180.00
\newline The total income is: \textcolor{red}{0.00 + 0.00 + 4180.00 = 4180.00}. \textcolor{red}{\usym{2718}}
\newline
\textcolor{blue}{\textbf{TableMoE \textbf{(Ours):}}} 
\newline\begin{CJK*}{UTF8}{gbsn}
\small
让我们一步一步来思考: 
\newline  由图片信息可知，勾选人员为三位。 
\newline 三位勾选人员的收入分别为：高胜蓝：0.00; 贾加令：120.00；王成贵：60.00。
\newline 收入总计是：0.00 + 120.00 + 60.00 = 180.00。
\newline 因此，勾选人员收入合计为 180.00。\textcolor{green}{\usym{2714}}
\end{CJK*}
\newline Let's think about it step by step:
\newline From the table image, we can see that there are three people checked in box.
\newline The incomes of the three people who checked the box are: Gao Shenglan: 0.00; Jia Jialing: 120.00; Wang Chenggui: 60.00.
\newline Their total income is: 0.00 + 120.00 + 60.00 = 180.00.
\newline Therefore, the total income of the people who checked the box is 180.00. \textcolor{green}{\usym{2714}} 
\end{minipage}
\end{tcolorbox}
\end{table}

\begin{table}[!htbp]
\small
\centering
\setlength{\abovecaptionskip}{1pt}%
\setlength{\belowcaptionskip}{0pt}%
\caption{
Visualization results of multimodal table-centric reasoning based on incomplete table image on WMMTabDialog are provided to illustrate the performance differences across various models. In this example, domain knowledge is needed to infer the result.} 
\label{tab:wmmtab_inf_rsn_01}
\begin{tcolorbox}[colframe=black,colback=whitesmoke]
\begin{minipage}{\linewidth}
\centering
\includegraphics[scale=0.13]{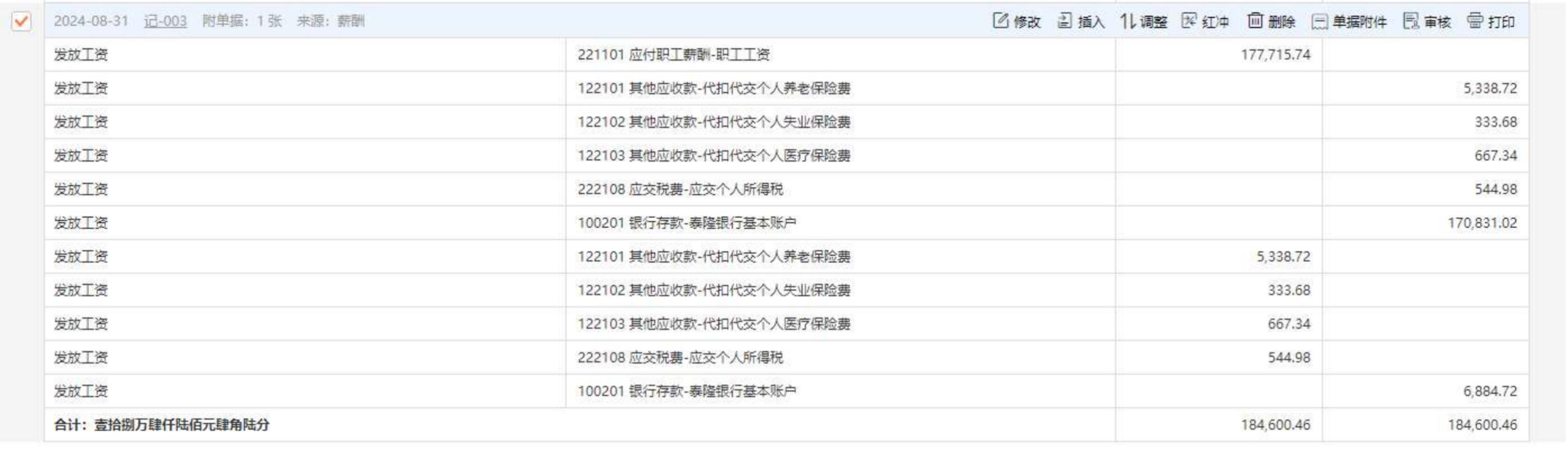}
\end{minipage}

\begin{minipage}{\linewidth}
\flushleft
\hspace*{\fill}
\hspace*{\fill}
\newline \textcolor{dodgerblue}{\textbf{Domain Knowledge:}}\newline
\begin{CJK*}{UTF8}{gbsn}
\small
银行存款属于资产类科目，借方表示增加，贷方表示减少，\dots。
\end{CJK*}
\newline Bank deposits are asset accounts, with debits indicating increases and credits indicating decreases.
\newline
\newline \textcolor{amber(sae/ece)}
{\textbf{Question:}} 
\newline 
\begin{CJK*}{UTF8}{gbsn}
\small
根据表格图像中发放的工资会计凭证，企业银行存款减少了多少钱？
\end{CJK*}
\newline 
Based on the payroll accounting voucher issued in the table image, how much did the company's bank deposits decrease?
\newline 
\newline \textcolor{forestgreen(web)}{\textbf{Ground Truth:}} \newline
\begin{CJK*}{UTF8}{gbsn}
\small
银行存款减少了177715.74元
\end{CJK*}
\newline 
Bank deposits decreased by 177,715.74 yuan
\newline
\newline
\newline\textcolor{blue}{\textbf{Qwen2-VL-72B:}} \newline 
\begin{CJK*}{UTF8}{gbsn}
\small 
根据表格中的信息,企业银行存款减少了\textcolor{red}{170,831.02元}。这是因为在会计凭证中,银行存款的减少是通过贷方记录的。
\textcolor{red}{\usym{2718}}
\end{CJK*}
\newline According to the information in the table, the company's bank deposits decreased by \textcolor{red}{170,831.02 yuan}. This is because in the accounting voucher, the decrease in bank deposits is recorded through the credit side. \textcolor{red}{\usym{2718}}
\newline
\newline
\textcolor{blue}{\textbf{GPT-4 Turbo with Vision:}} \newline 
\begin{CJK*}{UTF8}{gbsn}
\small
根据图片中的会计凭证，企业银行存款减少的总金额为\textcolor{red}{184,600.46元}。这是通过计算所有贷方金额的总和得出的，因为在资产类科目中，贷方表示金额的减少。\textcolor{red}{\usym{2718}}\end{CJK*}
\newline 
According to the accounting voucher in the image, the total amount of reduction in the company's bank deposits is \textcolor{red}{184,600.46 yuan}.  
This is arrived at by calculating the sum of all credit amounts, since in asset accounts, credits represent a decrease in amounts. \textcolor{red}{\usym{2718}}
\newline
\newline
\textcolor{blue}{\textbf{GPT-4o:}} \newline 
\begin{CJK*}{UTF8}{gbsn}
\small
根据图中信息，企业银行存款减少的金额是\textcolor{red}{184,600.46元}。\textcolor{red}{\usym{2718}}
\end{CJK*}
\newline According to the information in the image, the amount of reduction in the company's bank deposits is \textcolor{red}{184,600.46 yuan}. \textcolor{red}{\usym{2718}}
\newline
\newline
\textcolor{blue}{\textbf{TableMoE \textbf{(Ours):}}} 
\newline\begin{CJK*}{UTF8}{gbsn}
\small
让我们一步一步来思考: \newline 从表格图片中得知，存在两笔银行存款，分别为贷方余额170831.02，贷方余额6884.72。\newline 根据领域知识判断，贷方表示减少。\newline 银行存款减少金=170831.02+6884.72=177715.74。\newline 答案是：银行存款减少了177715.74元。\textcolor{green}{\usym{2714}}
\end{CJK*}
\newline Let's think step by step:
\newline From the table image, we know that there are two bank deposits, with a credit balance of 170831.02 and a credit balance of 6884.72, respectively.
\newline Based on domain knowledge, credits indicate a reduction .
\newline Bank deposit reduction = 170831.02 + 6884.72 = 177715.74.
\newline The answer is: bank deposits decreased by 177,715.74 yuan. \textcolor{green}{\usym{2714}} 
\end{minipage}
\end{tcolorbox}
\end{table}

\begin{table}[!htbp]
\small
\centering
\setlength{\abovecaptionskip}{1pt}%
\setlength{\belowcaptionskip}{0pt}%
\caption{
Visualization results of  multimodal table-centric reasoning based on table image with incomplete column header fonts on WMMTabDialog are provided to illustrate the performance differences across various models. 
} 
\label{tab:wmmtab_inf_rsn_02}
\begin{tcolorbox}[colframe=black,colback=whitesmoke]
\begin{minipage}{\linewidth}
\centering
\includegraphics[scale=0.27]{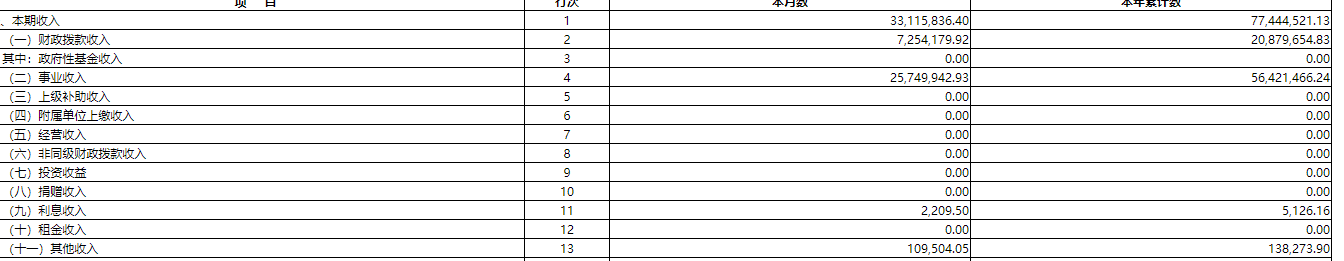}
\end{minipage}

\begin{minipage}{\linewidth}
\flushleft
\hspace*{\fill}
\hspace*{\fill}
\newline \textcolor{amber(sae/ece)}
{\textbf{Question:}} 
\newline 
\begin{CJK*}{UTF8}{gbsn}
\small
表中，截止上个月财政拨款收入本年累计数为多少？
\end{CJK*}
\newline 
In the table, what is the cumulative amount of fiscal appropriation revenue for this year as of last month?
\newline 
\newline \textcolor{forestgreen(web)}{\textbf{Ground Truth:}} \newline
13625474.91
\newline
\newline
\newline\textcolor{blue}{\textbf{Qwen2-VL-72B:}} \newline 
\begin{CJK*}{UTF8}{gbsn}
\small 
\textcolor{red}{20,879,654.83}.
\textcolor{red}{\usym{2718}}
\end{CJK*}
\newline
\newline
\textcolor{blue}{\textbf{GPT-4 Turbo with Vision:}} \newline 
\begin{CJK*}{UTF8}{gbsn}
\small
在表中，财政拨款收入的“截止上个月本年累计数”为\textcolor{red}{33,115,836.40}。\textcolor{red}{\usym{2718}}\end{CJK*}
\newline 
In the table, the "accumulated amount of fiscal appropriation revenue for the year as of last month" is \textcolor{red}{33,115,836.40}. \textcolor{red}{\usym{2718}}
\newline
\newline
\textcolor{blue}{\textbf{GPT-4o:}} \newline 
\begin{CJK*}{UTF8}{gbsn}
\small
表中，财政拨款收入截止上个月本年累计数为\textcolor{red}{20,879,654.83}。\textcolor{red}{\usym{2718}}
\end{CJK*}
\newline In the table, the cumulative amount of fiscal appropriation revenue for this year as of last month is \textcolor{red}{20,879,654.83}. \textcolor{red}{\usym{2718}}
\newline
\newline
\textcolor{blue}{\textbf{TableMoE \textbf{(Ours):}}} 
\newline\begin{CJK*}{UTF8}{gbsn}
\small
让我们一步一步来思考：\newline 从表格图片中得知，财政拨款收入本月数为7254179.92，本年累计数为20879654.83。\newline 上个月本年累计数=本年累计数-本月数=20879654.83-7254179.92=13625474.91。\newline 答案是：财政拨款收入上个月本年累计数为13625474.91。 \textcolor{green}{\usym{2714}}
\end{CJK*}
\newline Let's think it step by step:
\newline It can be seen from the table image that the fiscal allocation revenue this month is 7254179.92, and the cumulative number this year is 20879654.83.
\newline Last month’s cumulative number for this year = This year’s cumulative number – This month’s number = 20879654.83 - 7254179.92 = 13625474.91.
\newline The answer is: The cumulative amount of fiscal appropriation revenue last month this year is 13625474.91 \textcolor{green}{\usym{2714}} 
\end{minipage}
\end{tcolorbox}
\end{table}
\begin{table}[!htbp]
\small
\centering
\setlength{\abovecaptionskip}{1pt}%
\setlength{\belowcaptionskip}{0pt}%
\caption{
Visualization results of multimodal table-centric reasoning based on incomplete table image on WMMTabDialog are provided to illustrate the performance differences across various models. For simplicity, we omit the intermediate reasoning steps of other models and directly present the final results.} 
\label{tab:wmmtab_inf_rsn_03}
\begin{tcolorbox}[colframe=black,colback=whitesmoke]
\begin{minipage}{\linewidth}
\centering
\includegraphics[scale=0.175]{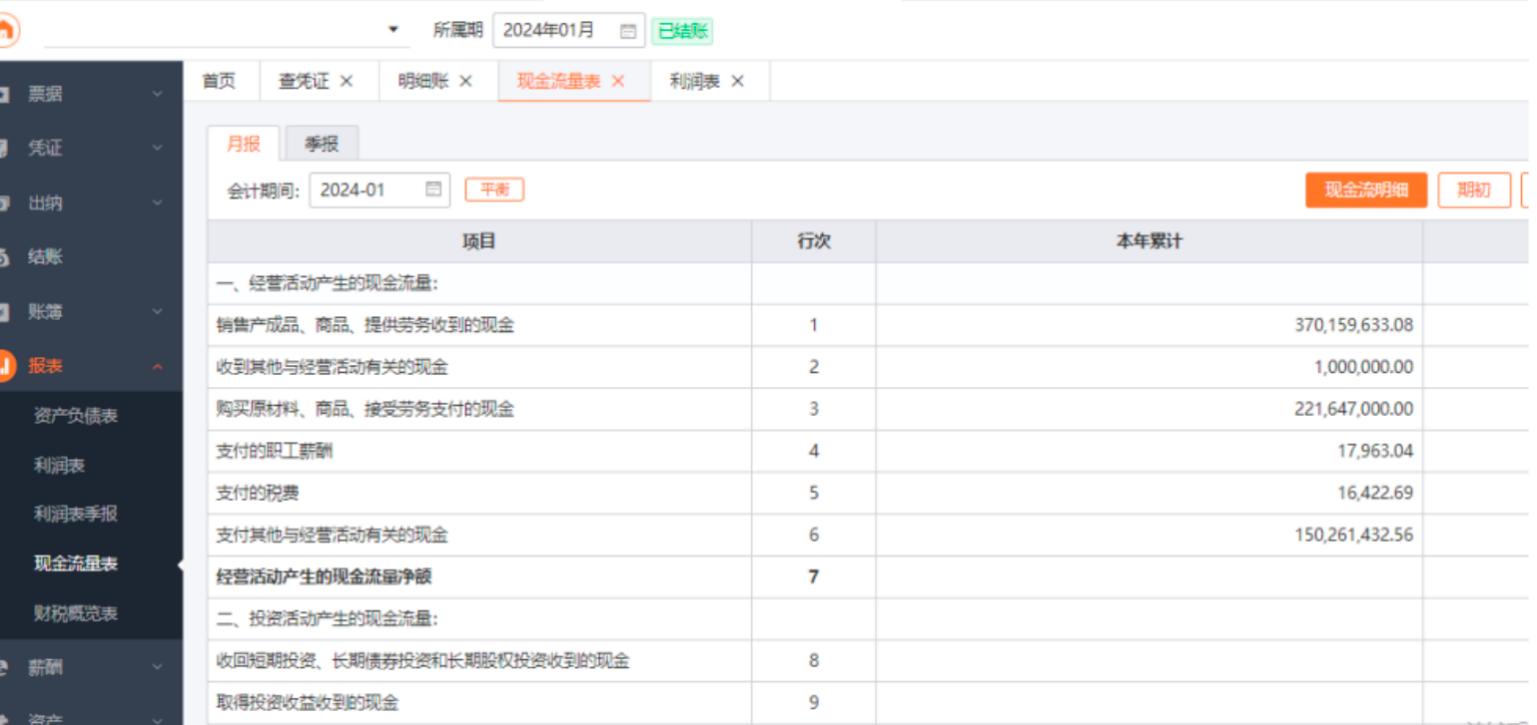}
\end{minipage}

\begin{minipage}{\linewidth}
\flushleft
\hspace*{\fill}
\hspace*{\fill}
\newline \textcolor{dodgerblue}{\textbf{Domain Knowledge:}}\newline
\begin{CJK*}{UTF8}{gbsn}
\small
现金流净额 = 现金流入量 - 现金流出量。
\end{CJK*}
\newline Net Cash Flow = Cash Inflow - Cash Outflow.
\newline
\newline \textcolor{amber(sae/ece)}
{\textbf{Question:}} 
\newline 
\begin{CJK*}{UTF8}{gbsn}
\small
现金流量表中，第7行经营活动产生的现金流量净额是多少？
\end{CJK*}
\newline 
On the statement of cash flows, what is the net cash flow from operating activities in 7th row?
\newline 
\newline \textcolor{forestgreen(web)}{\textbf{Ground Truth:}} \newline
-783185.21
\newline
\newline\textcolor{blue}{\textbf{Qwen2-VL-72B:}} \newline 
\begin{CJK*}{UTF8}{gbsn}
\small 
根据提供的现金流量表截图,经营活动产生的现金流量净额(第7行)是\textcolor{red}{150,261,432.56}。
\textcolor{red}{\usym{2718}}
\end{CJK*}
\newline
According to the cash flow statement provided, the net cash flow from operating activities (7th row) is \textcolor{red}{150,261,432.56}. \textcolor{red}{\usym{2718}}
\newline
\newline
\textcolor{blue}{\textbf{GPT-4 Turbo with Vision:}} \newline 
\begin{CJK*}{UTF8}{gbsn}
\small
在表中，财政拨款收入的“截止上个月本年累计数”为\textcolor{red}{33,115,836.40}。\textcolor{red}{\usym{2718}}\end{CJK*}
\newline 
In the table, the "accumulated amount of fiscal appropriation revenue for the year as of last month" is \textcolor{red}{33,115,836.40}. \textcolor{red}{\usym{2718}}
\newline
\newline
\textcolor{blue}{\textbf{GPT-4o:}} \newline 
\begin{CJK*}{UTF8}{gbsn}
\small
  因此，图中显示的经营活动产生的现金流量净额为\textcolor{red}{442,510,814.79}。\textcolor{red}{\usym{2718}}
\end{CJK*}
\newline Therefore, the net cash flow from operating activities shown in the image is \textcolor{red}{442,510,814.79}. \textcolor{red}{\usym{2718}}
\newline
\newline
\textcolor{blue}{\textbf{TableMoE \textbf{(Ours):}}} 
\newline\begin{CJK*}{UTF8}{gbsn}
\small
让我们一步一步来思考：\newline从领域知识中得知，现金流量表中，现金流净额=现金流入量-现金流出量。\newline 从表格图片中得知，图中第1、2行次为现金流入，第3-6行次为现金流出。\newline现金流入量=第1行次销售产成品、商品、提供劳务收到的现金+收到与其他经营活动有关的现金=370159633.08+1000000=470159633.08; 现金流出量=购买原材料、商品、接受劳务支付的现金+支付的职工薪酬+支付的税费+支付其他与经营活动有关的现金=221647000.00+17963.04+16422.69+150261432.56=371942828.3。\newline将已知的数值代入上述计算公式，现金流净额=470159633.08-371942828.3=-783185.21。
\newline答案是：第7行经营活动产生的现金流量净额为-783185.21。 \textcolor{green}{\usym{2714}}
\end{CJK*}
\newline Let's think it step by step:
\newline From the domain knowledge, we know that in the cash flow statement, net cash flow = cash inflow - cash outflow. \newline From the table image, we know that the 1st and 2nd rows in the figure are cash inflows, and the 3rd to 6th rows are cash outflows. \newline Cash inflow = cash received from the sale of finished products, goods, and services in the first row + cash received related to other operating activities = 370,159,633.08 + 1,000,000 = 470,159,633.08; Cash outflow = cash paid for the purchase of raw materials, goods, and services + employee salaries paid + taxes paid + other cash paid related to operating activities = 221,647,000.00 + 17,963.04 + 16,422.69 + 150,261,432.56 = 371,942,828.3. \newline Substitute the known values into the above calculation formula, net cash flow = 470159633.08-371942828.3=-783185.21.
\newline The answer is: The net cash flow from operating activities in 7th row  is -783185.21. \textcolor{green}{\usym{2714}}
\end{minipage}
\end{tcolorbox}
\end{table}
\begin{table}[!htbp]
\small
\centering
\setlength{\abovecaptionskip}{1pt}%
\setlength{\belowcaptionskip}{0pt}%
\caption{Visualization results for multimodal table reasoning on hard sample in WMMTabDialog is presented to highlight the performance distinctions among various models.
\textit{Note that the invoice tax amount details are displayed incompletely and the total tax amount has overlapping numbers}. We have erased the private sensitive information on the picture to prevent the leakage of personal or corporate information.} 
\label{tab:wmmtab_hard_invoice}
\begin{tcolorbox}[colframe=black,colback=whitesmoke]
\begin{minipage}{\linewidth}
\centering
\includegraphics[scale=0.28]{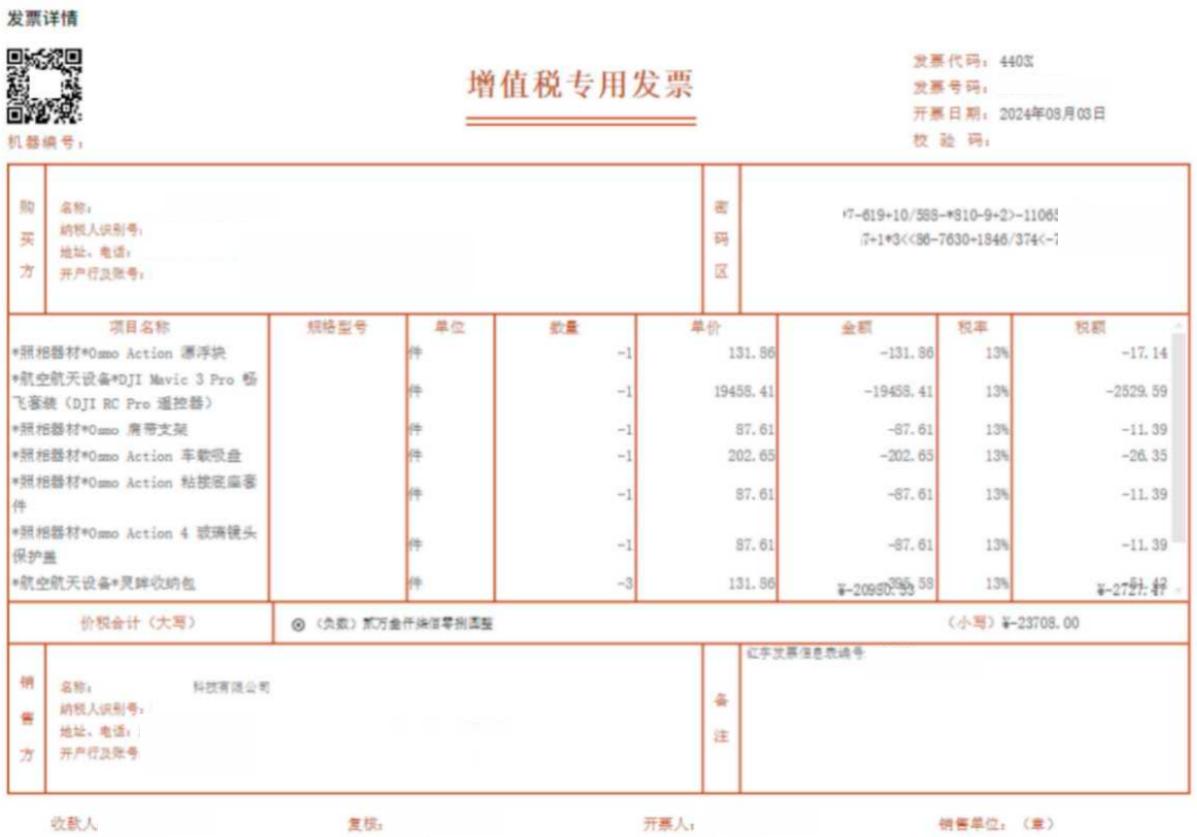}
\end{minipage}

\begin{minipage}{\linewidth}
\flushleft
\hspace*{\fill}
\hspace*{\fill}
\newline \textcolor{amber(sae/ece)}{\textbf{Question:}} 
\newline 
\begin{CJK*}{UTF8}{gbsn}
\small
表格图像中发票税额合计是多少？
\end{CJK*}
\newline 
What is the total invoice tax amount in the table image?
\newline 
\newline \textcolor{forestgreen(web)}{\textbf{Ground Truth:}} \newline
\begin{CJK*}{UTF8}{gbsn}
\small
-2727.47元
\end{CJK*}
\newline -2727.47 yuan
\newline
\newline\textcolor{blue}{\textbf{Qwen2.5-VL-72B-Instruct:}} \newline 
\begin{CJK*}{UTF8}{gbsn}
\small 
发票税额合计为 \textcolor{red}{-2721.47 元}。 \textcolor{red}{\usym{2718}}
\end{CJK*}
\newline
The total invoice tax amount is \textcolor{red}{-2721.47 yuan}.
\newline
\newline \textcolor{blue}{\textbf{GPT-4 Turbo with Vision:}} \newline 
\begin{CJK*}{UTF8}{gbsn}
\small
表中显示的发票税额合计是\textcolor{red}{23708.00元}。\textcolor{red}{\usym{2718}}\end{CJK*}
\newline 
The total invoice tax shown in the table is \textcolor{red}{23,708.00 yuan}. \textcolor{red}{\usym{2718}}
\newline
\newline \textcolor{blue}{\textbf{GPT-4o:}} \newline 
\begin{CJK*}{UTF8}{gbsn}
\small
发票税额合计是\textcolor{red}{2708.00 元}。\textcolor{red}{\usym{2718}}
\end{CJK*}
\newline The total amount of invoice tax is \textcolor{red}{2708.00 yuan}. \textcolor{red}{\usym{2718}}
\newline
\newline \textcolor{blue}{\textbf{TableMoE \textbf{(Ours):}}} 
\newline\begin{CJK*}{UTF8}{gbsn}
\small
让我们一步一步来思考：
\newline 表格图像中发票税额合计项等于第一列各个项目名称对应的最后一列的税额相加。
\newline 即 (-17.14) + (-2529.59) + (-11.39) + (-26.35) + (-11.39) + (-11.39) +  \textcolor{red}{61.42} = \textcolor{red}{-2545.83}。
\newline 因此，表格图像中发票税额合计是 \textcolor{red}{-2545.83}。
\textcolor{red}{\usym{2718}}
\end{CJK*}
\newline Let's think step by step:
\newline The total tax amount of the invoice in the table image is equal to the sum of the tax amounts in the last column corresponding to the item names in the first column.
\newline That is, (-17.14) + (-2529.59) + (-11.39) + (-26.35) + (-11.39) + (-11.39) + \textcolor{red}{61.42} = \textcolor{red}{-2545.83}. 
\newline Therefore, the total invoice tax amount in the table image is \textcolor{red}{-2545.83}. \textcolor{red}{\usym{2718}}
\end{minipage}
\end{tcolorbox}
\end{table}
\begin{table}[!htbp]
\small
\centering
\setlength{\abovecaptionskip}{1pt}%
\setlength{\belowcaptionskip}{0pt}%
\caption{
Visualization results of multimodal table-centric reasoning on WMMFinQA to illustrate the performance differences across various models. This sample's question-answering (QA) type is table image based QA. The version of GPT-4o used in this study is gpt-4o-2024-11-20, while other experiments utilized gpt-4o-2024-08-06 unless specified otherwise.} 
\label{tab:WMMFinQA_inf_rsn_01}
\begin{tcolorbox}[colframe=black,colback=whitesmoke]
\begin{minipage}{\linewidth}
\centering
\includegraphics[scale=0.43]{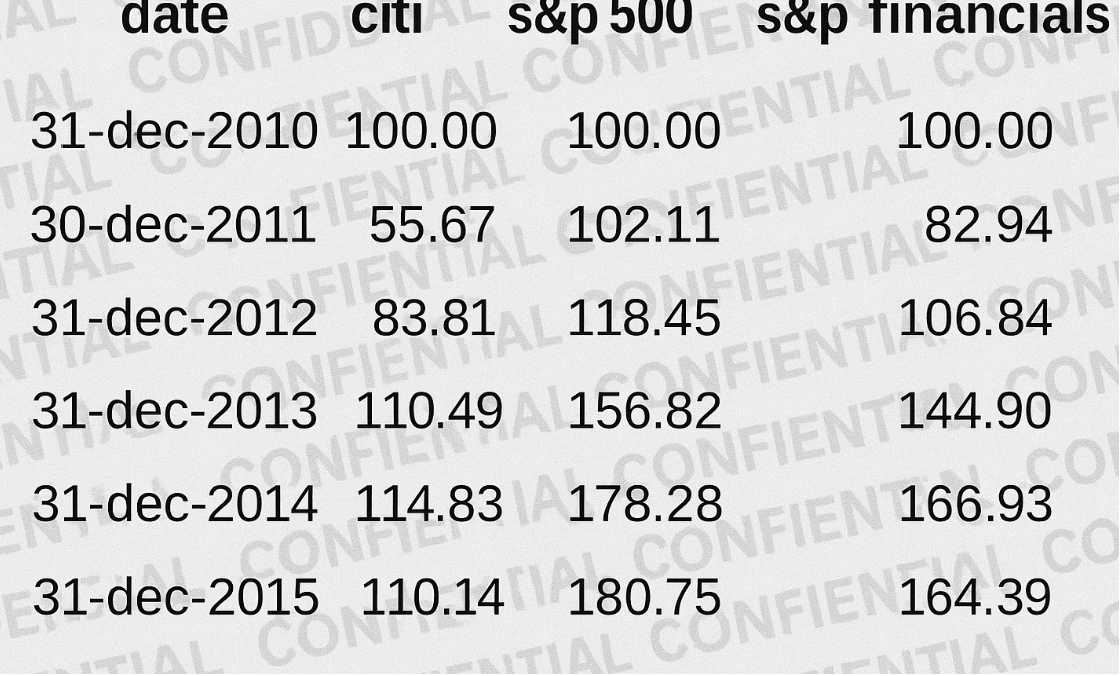}
\end{minipage}

\begin{minipage}{\linewidth}
\flushleft
\hspace*{\fill}
\hspace*{\fill}
\newline 
\newline \textcolor{amber(sae/ece)}{\textbf{Question:}} 
\newline 
What was the ratio of the growth of the cumulative total return for citi compared to s\&p 500 in 2013?
\newline 
\newline \textcolor{forestgreen(web)}{\textbf{Ground Truth:}} 
\newline 0.185
\newline
\newline\textcolor{blue}{\textbf{Qwen2.5-VL-72B-Instruct:}}  
\newline To find the ratio of the growth of the cumulative total return for Citi compared to the S\&P 500 in 2013, we need to follow these steps:
\newline Identify the values for Citi and S\&P 500 at the end of 2012 and 2013:
Citi: End of 2012: \$83.81, End of 2013: \$110.49
S\&P 500: End of 2012: \$118.45, End of 2013: \$156.82
\newline Calculate the increase in cumulative total return for Citi and S\&P 500 from 2012 to 2013:Increase for Citi:[110.49 - 83.81 = 26.68] Increase for S\&P 500:[156.82 - 118.45 = 38.37]
\newline Determine the ratio of the growth of the cumulative total return for Citi compared to S\&P 500:$[\text{Ratio} = \frac{\text{Increase for Citi}}{\text{Increase for S\&P 500}} = \frac{26.68}{38.37}]$
\newline Simplify the ratio:$[\frac{26.68}{38.37} \approx 0.695]$
\newline Therefore, the ratio of the growth of the cumulative total return for Citi compared to the S\&P 500 in 2013 is approximately (\textcolor{red}{0.695}). \textcolor{red}{\usym{2718}}
\newline
\newline 
\textcolor{blue}{\textbf{GPT-4o:}} \newline 
To find the ratio of the growth of the cumulative total return for Citi compared to the S\&P 500 in 2013:
\newline 1. **Calculate the growth for Citi in 2013:**
$\text{Growth for Citi} = \frac{\text{Value at 31-Dec-2013} - \text{Value at 31-Dec-2012}}{\text{Value at 31-Dec-2012}} \times 100 \frac{110.49 - 83.81}{83.81} \times 100 = \frac{26.68}{83.81} \times 100 \approx 31.84\%$.
\newline 2. **Calculate the growth for the S\&P 500 in 2013:** $\text{Growth for S\&P 500} = \frac{\text{Value at 31-Dec-2013} - \text{Value at 31-Dec-2012}}{\text{Value at 31-Dec-2012}}\times 100 \frac{148.65 - 118.45}{118.45} \times 100 = \frac{30.2}{118.45} \times 100 \approx 25.5\%$
\newline 3. **Calculate the ratio:**
$\text{Ratio of growth} = \frac{\text{Growth for Citi}}{\text{Growth for S\&P 500}}$
\newline Substitute the values: $\text{Ratio of growth} = \frac{31.84}{25.5} \approx 1.25$. 
\newline Final Answer: The ratio of the growth of Citi to the S\&P 500 in 2013 is approximately \textcolor{red}{1.25}. \textcolor{red}{\usym{2718}}
\newline
\newline
\textcolor{blue}{\textbf{TableMoE \textbf{(Ours):}}}
\newline Let's think step by step:
\newline From the table image, the growth of the cumulative total return for Citi in 2013 is calculated as: Growth in 2013 - Growth in 2010 = 110.49 - 100 = 10.49. 
\newline Also from the table image, the growth of the cumulative total return for s\&p 500 in 2013 is calculated as: Growth in 2013 - Growth in 2010 = 156.82 - 100 = 56.82.
\newline To find the ratio of the growth of the cumulative total return for Citi compared to S\&P 500, use the formula: (Growth of Citi / Growth of S\&P 500).
Substituting the values, we get (\$10.49 / \$56.82) = 0.1846.
\newline Rounding to three decimal places gives 0.185.
\newline \{The answer is: 0.185\}  \textcolor{green}{\usym{2714}} 
\end{minipage}
\end{tcolorbox}
\end{table}


\begin{table}[!htbp]
\small
\centering
\setlength{\abovecaptionskip}{1pt}%
\setlength{\belowcaptionskip}{0pt}%
\caption{
Visualization results of multimodal table-centric reasoning on WMMFinQA to illustrate the performance differences across various models. The sample's question-answering (qa) type is text (question and domain knowledge) and table (table image) based qa, which means that assistant (MLLM) needs acquire information from both text and table to infer the results.} 
\label{tab:WMMFinQA_inf_rsn_02}
\begin{tcolorbox}[colframe=black,colback=whitesmoke]
\begin{minipage}{\linewidth}
\centering
\includegraphics[scale=0.43]{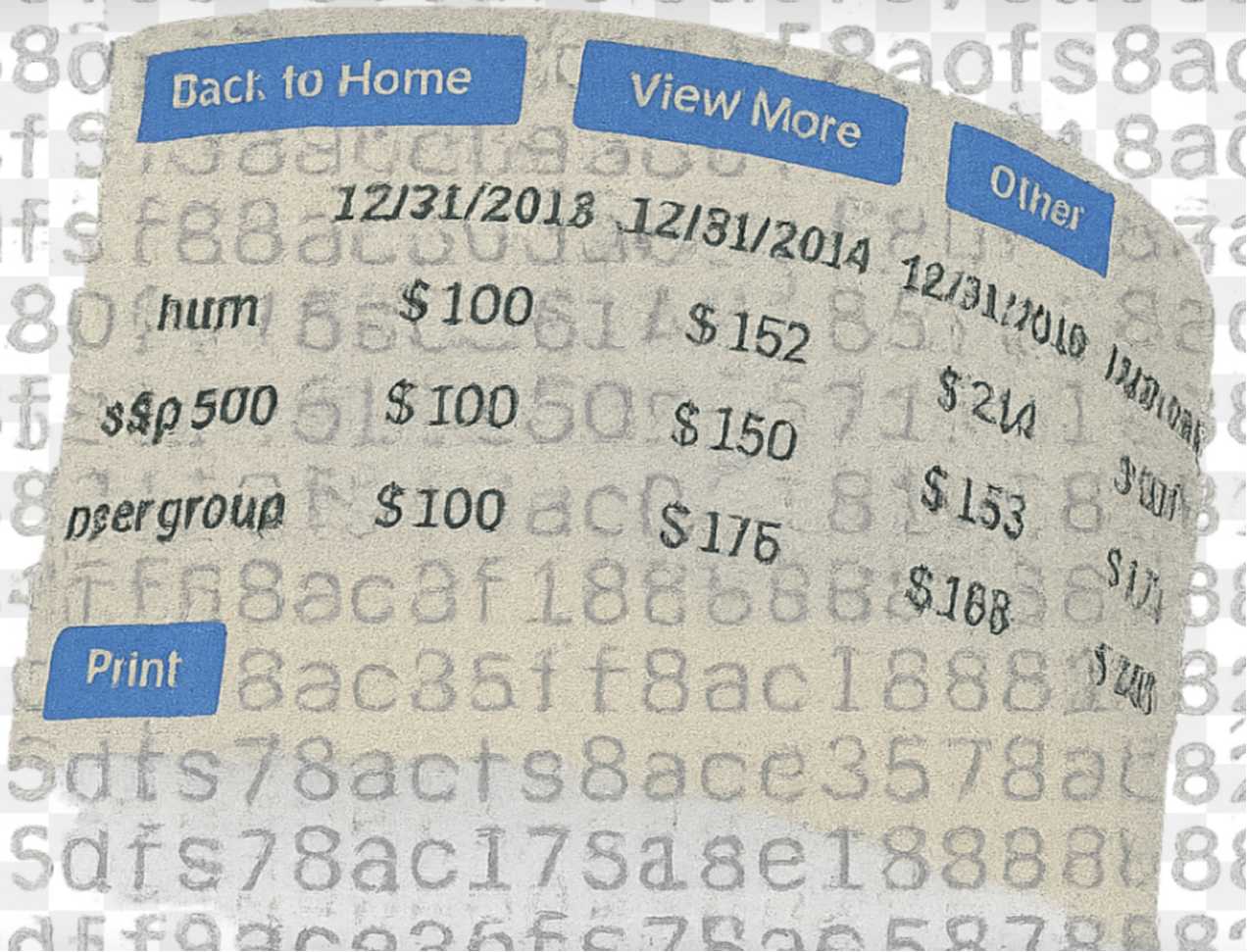}
\end{minipage}

\begin{minipage}{\linewidth}
\flushleft
\hspace*{\fill}
\hspace*{\fill}
\newline \textcolor{dodgerblue}{\textbf{Domain Knowledge:}}\newline
\dots the graph assumes an investment of \$ 100 in each of our common stock , the s\&p 500 , and the peer group on december 31 , 2012 , and that dividends were reinvested when paid \dots.
\hspace*{\fill}
\newline 
\newline \textcolor{amber(sae/ece)}{\textbf{Question:}} 
\newline 
What is the highest return for the second year of the investment?
\newline 
\newline \textcolor{forestgreen(web)}{\textbf{Ground Truth:}} 
\newline 114
\newline
\newline\textcolor{blue}{\textbf{Qwen2.5-VL-72B-Instruct:}}  
\newline To determine the highest return for the second year of the investment, we need to look at the values for the year 2013 (which is the second year after the initial investment in 2012).
\newline The value for "hum" in 2013 is \$152, the value for "s\&p 500" in 2013 is \$132, and the value for "peer group" in 2013 is \$137. The highest value among these is \$152 for "hum".
\newline $[ \text{Percentage Return} = \left(\frac{\text{Final Value} - \text{Initial Value}}{\text{Initial Value}} \right) \times 100 ]$
\newline Here, the initial value is \$100 and the final value is \$152.
$[ \text{Percentage Return} = \left( \frac{152 - 100}{100} \right) \times 100 = \left( \frac{52}{100} \right) \times 100 = 52\% ]$
\newline So, the highest return for the second year of the investment is \textcolor{red}{52\%}. \textcolor{red}{\usym{2718}}
\newline
\newline 
\textcolor{blue}{\textbf{GPT-4 Turbo with Vision:}} \newline 
The investment values for the second year (2013) are given for different entities.
\newline For HUM, the value in the second year is \$152.
\newline For S\&P 500, the value in the second year is \$132.
\newline For the peer group, the value in the second year is \$137.
\newline To find the highest return for the second year, we compare these values.
\newline The highest value among these is \$152 for HUM.
\newline To calculate the return, use the formula: ((Value in Second Year / Initial Value) - 1) * 100.
\newline Substituting the values for HUM, we get ((152 / 100) - 1) * 100 = (1.52 - 1) * 100 = 0.52 * 100 = 52\%.
\newline The highest return for the second year of the investment is \textcolor{red}{52\%}.
\textcolor{red}{\usym{2718}}
\newline
\newline
\textcolor{blue}{\textbf{TableMoE \textbf{(Ours):}}}
\newline Let's think step by step:
\newline Based on the table image, we can determine that the first year of the investment is from 12/31/2012 to 12/31/2013, while the second year of the investment is from 12/31/2013 to 12/31/2014.
\newline The investment values for the second year (12/31/2014) are given for different entities.
\newline For HUM, the value in the second year is \$214.
\newline For S\&P 500, the value in the second year is \$150.
\newline For the peer group, the value in the second year is \$175.
\newline To find the highest return for the second year, we compare these values.
\newline The highest value among these is \$214 for HUM.
\newline To calculate the return, use the formula: $(\frac{\text{Value in Second Year} - \text{Initial Value}}{\text{Initial Value}})\times 100 = \frac{\$214 - \$100 }{\$100} \times 100 = \frac{\$114}{\$100} \times 100 = 114\%$.
\newline \{The answer is: 114\%.\}  \textcolor{green}{\usym{2714}} 
\end{minipage}
\end{tcolorbox}
\end{table}


\begin{table}[!htbp]
\small
\centering
\setlength{\abovecaptionskip}{1pt}%
\setlength{\belowcaptionskip}{0pt}%
\caption{
Visualization results of multimodal table-centric reasoning on WMMFinQA to illustrate the performance differences across various models. The sample's QA type is table image based QA. The version of GPT-4o used in this study is gpt-4o-2024-11-20, while other experiments utilized gpt-4o-2024-08-06 unless specified otherwise.
} 
\label{tab:WMMFinQA_inf_rsn_03}
\begin{tcolorbox}[colframe=black,colback=whitesmoke]
\begin{minipage}{\linewidth}
\centering
\includegraphics[scale=0.45]{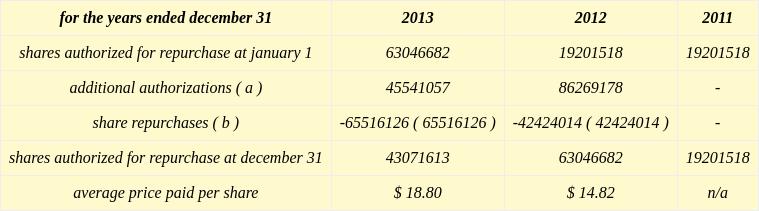}
\end{minipage}

\begin{minipage}{\linewidth}
\flushleft
\hspace*{\fill}
\newline \textcolor{amber(sae/ece)}{\textbf{Question:}} 
\newline 
What is the total cash outflow spent for shares repurchased during 2013 , in millions?
\newline 
\newline \textcolor{forestgreen(web)}{\textbf{Ground Truth:}} 
\newline 809.7
\newline
\textcolor{blue}{\textbf{Qwen2.5-VL-72B-Instruct:}} \newline
\newline To calculate the total cash outflow spent for shares repurchased during 2013, we need to multiply the number of shares repurchased by the average price paid per share.
\newline From the table: Shares repurchased in 2013: 65,516,126 and Average price paid per share in 2013: \$18.80
Now, let's calculate the total cash outflow:
$[ \text{Total cash outflow} = \text{Shares repurchased} \times \text{Average price per share} = 65,516,126 \times 18.80 = 1,232,479,996.80]$
\newline To convert this to millions:
$[ \text{Total cash outflow in millions} = \frac{1,232,479,996.80}{1,000,000} = 1,232.48]$
\newline So, the total cash outflow spent for shares repurchased during 2013 is approximately \textcolor{red}{\$1,232.48 million}. \textcolor{red}{\usym{2718}}
\newline
\newline
\textcolor{blue}{\textbf{GPT-4o:}} \newline 
To calculate the total cash outflow spent for shares repurchased during 2013:
\newline 1. **Determine the number of shares repurchased in 2013:**
$\text{Shares repurchased} = \text{-65516126 (as given in the table, which is "b")}$
This means 65,516,126 shares were repurchased.
\newline 2. **Identify the average price paid per share in 2013:**
$ \text{Average price paid per share} = 18.80 \, \text{(as given in the table)}$
\newline 3. **Calculate the total cash outflow:**
$\text{Total cash outflow} = \text{Shares repurchased} \times \text{Average price paid per share}$
 Substituting the values:
$\text{Total cash outflow} = 65,516,126 \times 18.80 = 1,231,705,177 \, \text{(in dollars, or 1,231.71 million dollars)}$
\newline Final Answer:
The total cash outflow spent for shares repurchased during 2013 is approximately ** \textcolor{red}{1,231.71 million} dollars**.
\textcolor{red}{\usym{2718}}
\newline
\newline
\textcolor{blue}{\textbf{TableMoE \textbf{(Ours):}}}
\newline Let's think step by step:
\newline The table image provides the shares authorized for repurchase at december 31 of 2013 is 43,071,613.
\newline It also provides the average price paid per share during 2013 as \$18.80.
\newline To find the total cash outflow for shares repurchased, we use the formula: Number of Shares Repurchased * Average Price Paid per Share.
\newline Substituting the values, we get 43,071,613 * \$18.80 = \$809,746,324.4.
\newline To convert this amount into millions, we divide by 1,000,000.
Thus, \$809,746,324.4 / 1,000,000 = \$809.7 million.
\newline The total cash outflow for shares repurchased during 2013 is 809.7 million.
\newline\{The answer is: 809.7\} \textcolor{green}{\usym{2714}} 
\end{minipage}
\end{tcolorbox}
\end{table}
\begin{table}[!htbp]
\small
\centering
\setlength{\abovecaptionskip}{1pt}%
\setlength{\belowcaptionskip}{0pt}%
\caption{
Visualization results of multimodal table-centric reasoning on WMMTatQA to illustrate the performance differences across various models. The sample's QA type is table image based QA. The version of GPT-4o used in this study is gpt-4o-2024-11-20, while other experiments utilized gpt-4o-2024-08-06 unless specified otherwise.
} 
\label{tab:WMMTatQA_inf_rsn_01}
\begin{tcolorbox}[colframe=black,colback=whitesmoke]
\begin{minipage}{\linewidth}
\centering
\includegraphics[scale=0.30]{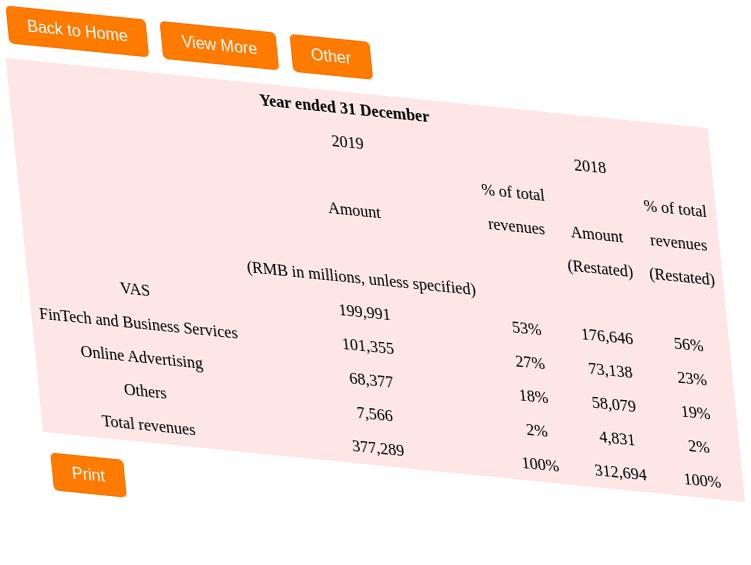}
\end{minipage}

\begin{minipage}{\linewidth}
\flushleft
\hspace*{\fill}
\newline \textcolor{amber(sae/ece)}{\textbf{Question:}} 
\newline 
What is the year-on-year change in VAS Business revenue?
\newline 
\newline \textcolor{forestgreen(web)}{\textbf{Ground Truth:}} 
\newline 23345
\newline
\newline
\textcolor{blue}{\textbf{Qwen2.5-VL-72B-Instruct:}} 
\newline The VAS Business revenue in 2019 was 199,991 million RMB, and in 2018 it was 176,646 million RMB.
\newline To calculate the year-on-year change:
(199,991 - 176,646) / 176,646 * 100 = 13.2\%
\newline So, the year-on-year change in VAS Business revenue is an increase of \textcolor{red}{13.2\%}. \textcolor{red}{\usym{2718}}
\newline
\newline
\textcolor{blue}{\textbf{GPT-4o:}} \newline 
To find the year-on-year (YoY) change in VAS (Value-Added Services) Business revenue, follow these steps:
\newline 1. **Extract the VAS revenues for 2019 and 2018:**
\newline \quad - VAS revenue in 2019 = **\textcolor{red}{199,391}** (in RMB millions)
   \newline \quad - VAS revenue in 2018 = **176,646** (in RMB millions)
\newline 2. **Calculate the change in revenue:**
\newline $\text{Change in revenue} = \text{Revenue in 2019} - \text{Revenue in 2018} = \text{199,391 - 176,646 = 22,745} \text{(in RMB millions)}$
\newline 3. **Calculate the year-on-year percentage change:**
\newline  $ \text{YoY percentage change} = \left( \frac{\text{Change in revenue}}{\text{Revenue in 2018}} \right) \times 100$
\newline $\text{YoY percentage change} = \left( \frac{22,745}{176,646} \right) \times 100 \approx 12.88\%$
\newline Final Answer: The year-on-year change in VAS Business revenue is **RMB \textcolor{red}{22,745} million** (a growth of approximately **12.88\%**).
\textcolor{red}{\usym{2718}}
\newline
\newline
\textcolor{blue}{\textbf{TableMoE \textbf{(Ours):}}}
\newline Let's think step by step:
\newline The year-on-year change in VAS (Value-Added Services) Business revenue from 2018 to 2019 is an increase from RMB 176,646 million to RMB 199,991 million. 
\newline To calculate the change: \text{Change in VAS from 2018 to 2019} = \text{New Value} - \text{Old Value} =  199,991 - 176,646 = 23345.
\newline Thus, the year-on-year change in VAS Business revenue is 23345 million.
\newline\{The answer is: 23345\} \textcolor{green}{\usym{2714}} 
\end{minipage}
\end{tcolorbox}
\end{table}
\begin{table}[htbp]
\small
\centering
\setlength{\abovecaptionskip}{1pt}%
\setlength{\belowcaptionskip}{0pt}%
\caption{
Visualization results of multimodal table-centric reasoning on WMMTatQA to illustrate the performance differences across various models. The sample's QA type is table image based QA. The version of GPT-4o used in this study is gpt-4o-2024-11-20, while other experiments utilized gpt-4o-2024-08-06 unless specified otherwise.
} 
\label{tab:WMMTatQA_inf_rsn_02}
\begin{tcolorbox}[colframe=black,colback=whitesmoke]
\begin{minipage}{\linewidth}
\centering
\includegraphics[scale=0.50]{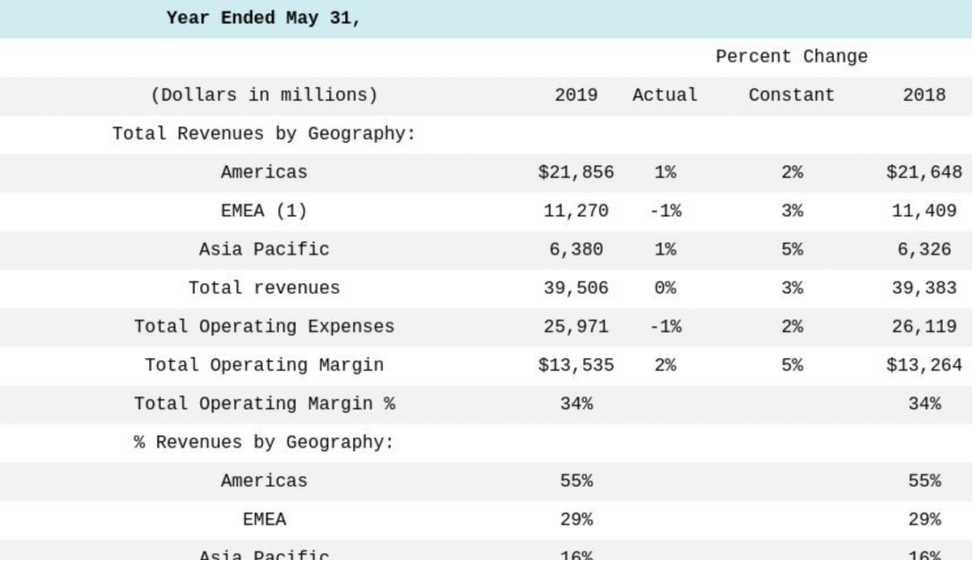}
\end{minipage}

\begin{minipage}{\linewidth}
\flushleft
\hspace*{\fill}
\newline \textcolor{amber(sae/ece)}{\textbf{Question:}} 
\newline 
What was the percentage increase in revenue from the EMEA and Americas regions in 2019 as compared to 2018?
\newline 
\newline \textcolor{forestgreen(web)}{\textbf{Ground Truth:}} 
\newline 0.21
\newline
\newline
\textcolor{blue}{\textbf{Qwen2.5-VL-72B-Instruct:}} 
\newline The percentage increase in revenue from the EMEA and Americas regions in 2019 as compared to 2018 can be calculated using the "Actual" percent change provided in the table.
\newline For the Americas region:
The actual percent change is 1\%.
\newline For the EMEA region:
The actual percent change is -1\%.
\newline So, the Americas region saw a \textcolor{red}{1\%} increase in revenue, while the EMEA region saw a \textcolor{red}{1\%} decrease in revenue. \textcolor{red}{\usym{2718}}
\newline
\newline
\textcolor{blue}{\textbf{GPT-4o:}} \newline 
To find the percentage increase in revenue for the **EMEA** and **Americas** regions in 2019 compared to 2018, use the formula:
$\text{Percentage Increase} = \frac{\text{Revenue in 2019} - \text{Revenue in 2018}}{\text{Revenue in 2018}} \times 100$
**Americas** Region:
- 2019 Revenue: \$21,856 million
- 2018 Revenue: \$21,648 million
$\text{Percentage Increase} = \frac{21,856 - 21,648}{21,648} \times 100 = \frac{208}{21,648} \times 100 \approx 0.96\%$
\newline **EMEA** Region:
- 2019 Revenue: \$11,378 million
- 2018 Revenue: \$10,927 million
$\text{Percentage Increase} = \frac{11,378 - 10,927}{10,927} \times 100 = \frac{451}{10,927} \times 100 \approx 4.13\%$
\newline Final Answer:
- **Americas**: Approximately ** \textcolor{red}{0.96\%}** increase.
- **EMEA**: Approximately ** \textcolor{red}{4.13\%}** increase.
\textcolor{red}{\usym{2718}}
\newline
\newline
\textcolor{blue}{\textbf{TableMoE \textbf{(Ours):}}}
\newline Let's think step by step:
\newline The total revenue from the Americas and  EMEA in 2019 is: Revenue of Americas and EMEA in 2019 = Revenue of Americas in 2019 + Revenue of EMEA in 2019 = \$21,856 + \$11,270 = \$33,126.
\newline The total revenue from the Americas and  EMEA in 2018 is: Revenue of Americas and EMEA in 2018 = Revenue of Americas in 2018 + Revenue of EMEA in 2018 = \$21,648 + \$11,409 = \$33057.
\newline  The percentage increase in revenue from the EMEA and Americas regions in 2019 as
compared to 2018 is: (Revenue of Americas and EMEA in 2019 - Revenue of Americas and EMEA in 2018) / Revenue of Americas and EMEA in 2018 $\times$ 100 = (33126 - 33057) / 33057 $\times$ 100 $\approx$ 0.21\%.
\newline\{The answer is: 0.21\} \textcolor{green}{\usym{2714}} 
\end{minipage}
\end{tcolorbox}
\end{table}
\begin{table}[htbp]
\small
\centering
\setlength{\abovecaptionskip}{1pt}%
\setlength{\belowcaptionskip}{0pt}%
\caption{
Visualization \textbf{hallucination} evaluation results on WMMTatQA to illustrate the performance differences across various models. We observed that, except for TableMoE and GPT-4o, Qwen2.5 model tended to hallucinate during reasoning when confronted with very blurred image. The version of GPT-4o used in this study is gpt-4o-2024-11-20, while other experiments utilized gpt-4o-2024-08-06 unless specified otherwise.
} 
\label{tab:WMMTatQA_inf_rsn_03}
\begin{tcolorbox}[colframe=black,colback=whitesmoke]
\begin{minipage}{\linewidth}
\centering
\caption*{Original Image}
\includegraphics[scale=0.30]{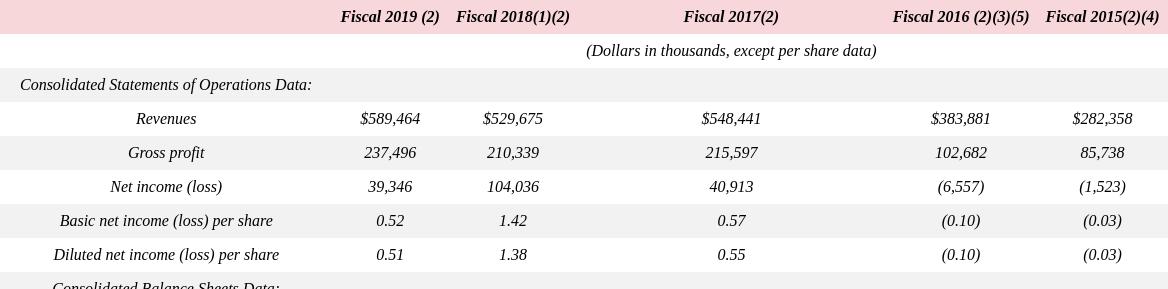}
\end{minipage}
\begin{minipage}{\linewidth}
\centering
\caption*{Blurred Image}
\includegraphics[scale=0.37]{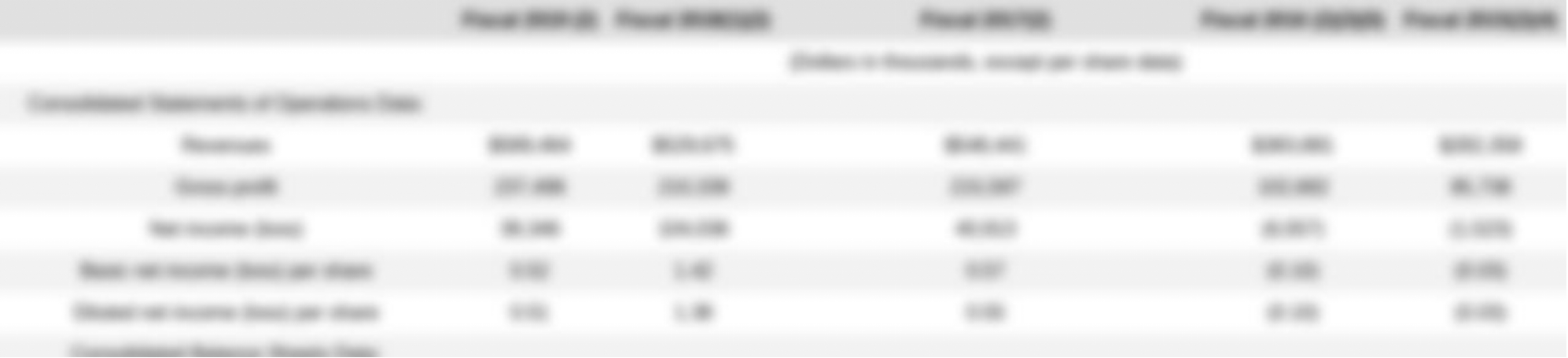}
\end{minipage}

\begin{minipage}{\linewidth}
\flushleft
\hspace*{\fill}
\newline \textcolor{amber(sae/ece)}{\textbf{Question:}} 
\newline 
In which years was revenue under 500,000 thousands?
\newline 
\newline \textcolor{forestgreen(web)}{\textbf{Ground Truth:}} 
\newline  2016, 2015
\newline
\newline
\textcolor{blue}{\textbf{Qwen2.5-VL-72B-Instruct:}} 
\newline 
Based on the provided image, revenue was under 500,000 thousands in the following years:
Fiscal Year 2019: Revenue is \$483,600.
Fiscal Year 2020: Revenue is \$467,475.
\textcolor{red}{\usym{2718}}
\newline
\newline
\textcolor{blue}{\textbf{GPT-4o:}} \newline 
The table is blurred, making it difficult to directly identify the years in which the revenue was under 500,000 (in thousands). Please provide a clearer image or describe the data, and I'll help analyze it!
\textcolor{green}{\usym{2714}}
\newline
\newline
\textcolor{blue}{\textbf{TableMoE \textbf{(Ours):}}}
\newline Due to the blurriness of the table, I can't pinpoint the specific years when revenue dropped below 500,000 (in thousands). please share a clearer image or detail the data, so I can analyze it properly.
 \textcolor{green}{\usym{2714}}
\end{minipage}
\end{tcolorbox}
\end{table}

\begin{table}[htbp]
\small
\centering
\setlength{\abovecaptionskip}{1pt}%
\setlength{\belowcaptionskip}{0pt}%
\caption{
Visualization of question answering (QA) type for multimodal table-centric reasoning on WMMTatQA. This sample is text (domain knowledge) and table image based QA, which means that assistant (MLLM) needs acquire information from both text and image to infer the results. Domain knowledge is a bit lengthy, most of the context has been omitted due to space constraints.} 
\label{tab:WMMTatQA_inf_rsn_04}
\begin{tcolorbox}[colframe=black,colback=whitesmoke]
\begin{minipage}{\linewidth}
\centering
\includegraphics[scale=0.37]{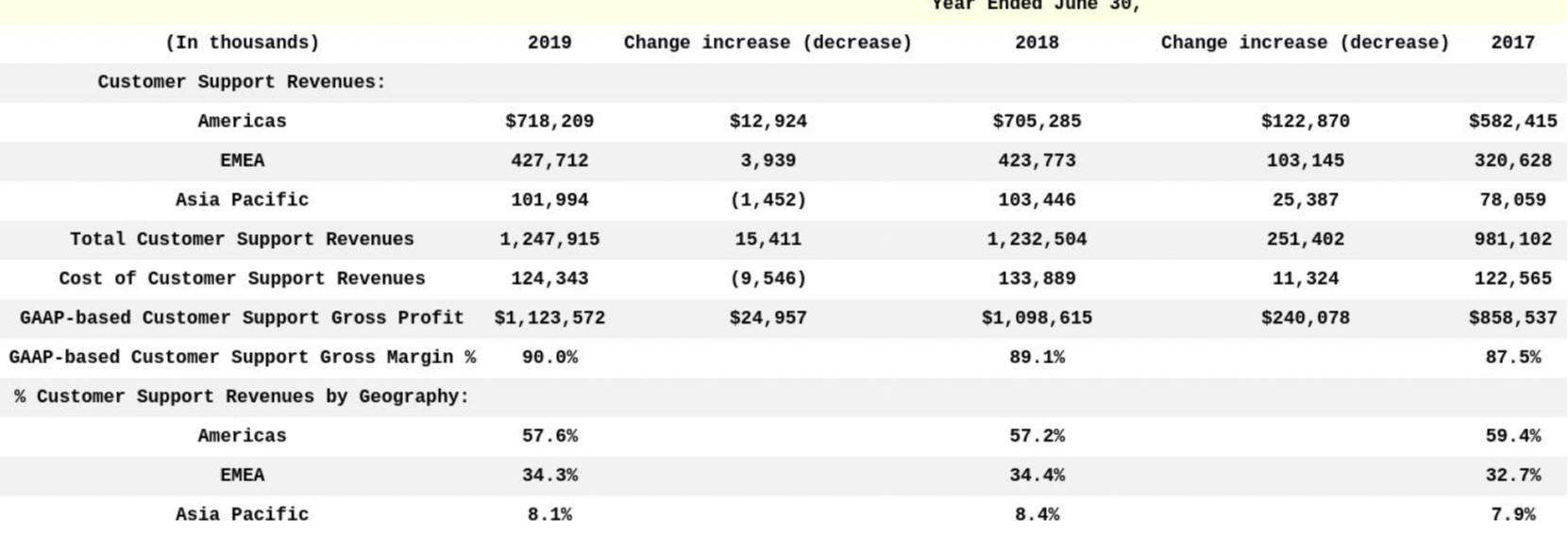}
\end{minipage}
\begin{minipage}{\linewidth}
\flushleft
\hspace*{\fill}
\hspace*{\fill}
\newline \textcolor{dodgerblue}{\textbf{Domain Knowledge:}}\newline
\dots. Cost of Customer support revenues decreased by \$9.5 million during the year ended June 30, 2019 as compared to the prior fiscal year, due to a decrease in labour-related costs of approximately \$9.9 million, partially offset by an increase in other miscellaneous costs of \$0.4 million, \dots.
\hspace*{\fill}
\newline \textcolor{amber(sae/ece)}{\textbf{Question:}} 
\newline 
For Fiscal year 2019, excluding the increase in other miscellaneous costs of \$0.4 million, what is the Cost of Customer Support Revenues for fiscal year 2019 in thousands(1,000s)?
\newline 
\newline \textcolor{forestgreen(web)}{\textbf{Ground Truth:}} 
\newline 123943
\newline
\textcolor{blue}{\textbf{TableMoE \textbf{(Ours):}}}
\newline Let's think step by step.
\newline The domain knowledge states that Cost of Customer support revenues was increased by other miscellaneous costs of \$0.4 million.
\newline Next, the table image shows that Cost of Customer support revenues for FY year 2018 is 133,889.
\newline Furthermore, the table image also indicates that Cost of Customer support revenues decreased by 9,546 from FY year 2018 to 2019.
\newline Thus, excluding the increase in other miscellaneous costs of \$0.4 million, we calculate Cost of Customer support revenues with the formula: (Cost of Customer support revenues for FY year 2018 - The Changed Value) - 0.4 * 1000. 
\newline Plugging in the values: (133,889 - 9,546) - 0.4 * 1000 = 123943
\newline
\{The answer is : 123943\}.
\end{minipage}
\end{tcolorbox}
\end{table}

\end{document}